\documentclass[10pt,twocolumn,letterpaper]{article}
\usepackage{cvpr}             
\usepackage{epsfig}
\usepackage{graphicx}
\usepackage{amsmath}
\usepackage{amssymb,mathtools}

\usepackage{bm}
\usepackage{booktabs}
\usepackage{dblfloatfix}
\usepackage[accsupp]{axessibility}  
\usepackage{xspace}
\usepackage[dvipsnames]{xcolor}
\usepackage{comment}
\usepackage{array}
\usepackage{makecell}
\usepackage{pifont}
\usepackage{multirow}
\usepackage{flushend}
\usepackage{colortbl}
\usepackage{enumitem}
\usepackage{comment} 
\usepackage{sidecap} 
\usepackage{nicefrac} 
\newcolumntype{R}[2]{%
    >{\adjustbox{angle=#1,lap=1.3\width-(#2)}\bgroup}%
    l%
    <{\egroup}%
} 
\usepackage{pgfkeys}
\usepackage[]{pgfplots,pgfplotstable}
\usepackage{algorithm}
\usepackage{algpseudocode}

\usepackage[export]{adjustbox} 

\usepackage[]{tikz}
\usetikzlibrary{arrows, positioning, math, fit, backgrounds}
\usepackage{tkz-kiviatt,numprint,fullpage}  
\usepackage[nomessages]{fp}

\newcommand{\norm}[1]{||#1||}

\pgfplotsset{compat=1.11,
    /pgfplots/ybar legend/.style={
    /pgfplots/legend image code/.code={%
       \draw[##1,/tikz/.cd,yshift=-0.25em]
        (0cm,0cm) rectangle (3pt,0.8em);},
   },
}

\makeatletter
\renewcommand\paragraph{\@startsection{paragraph}{4}{\z@}%
    {0.8ex \@plus 1ex \@minus .1ex}%
    {-1em}%
    {\normalfont \normalsize \bfseries}}
\makeatother

\def\addlegendimage{\csname pgfplots@addlegendimage\endcsname}

\newcommand\ours{\logen} 
\newcommand\logen{LOGen\xspace}
\newcommand\pixart{PixArt\xspace}
\newcommand\pixartal{PixArt\nobreakdash-$\alpha$\xspace}
\newcommand\pixartl{PixArt\nobreakdash-L\xspace}
\newcommand\dittd{DiT-3D\xspace}
\newcommand\dittdl{DiT-3DL\xspace}
\newcommand\ditpe{\dittdl\PE\xspace}
\newcommand\ditve{\dittdl\VE\xspace}
\newcommand\minkunet{MinkUNet\xspace}

\newcommand\PE{PE\fch}
\newcommand\VE{VE\tch}
\newcommand\tch{\textsubscript{\,3\,ch}\xspace}
\newcommand\fch{\textsubscript{\,4\,ch}\xspace}

\newcommand\NA{\footnotesize N/A}

\setlist[itemize]{leftmargin=*,topsep=2pt,itemsep=2pt}

\definecolor{cvprblue}{rgb}{0.21,0.49,0.74}
\usepackage[pagebackref,breaklinks,colorlinks,allcolors=cvprblue]{hyperref}

\title{LOGen: Toward Lidar Object Generation by Point Diffusion}
\author{\hspace{-3mm}Ellington Kirby$^1$,
Mickael Chen$^1$,
Renaud Marlet$^{1,2}$,
Nermin Samet$^1$ \\[2mm]
\hspace{-5mm}\textsuperscript{1}Valeo.ai, Paris, France
\hspace{2.5mm}\textsuperscript{2}LIGM, Ecole des Ponts, Univ Gustave Eiffel, CNRS, Marne-la-Vall\'ee, France \\
{\tt\small \url{https://nerminsamet.github.io/logen/}}
}

\begin{document}

\twocolumn[{%
\renewcommand\twocolumn[1][]{#1}%
\maketitle
\centering
\includegraphics[width=0.78\linewidth]{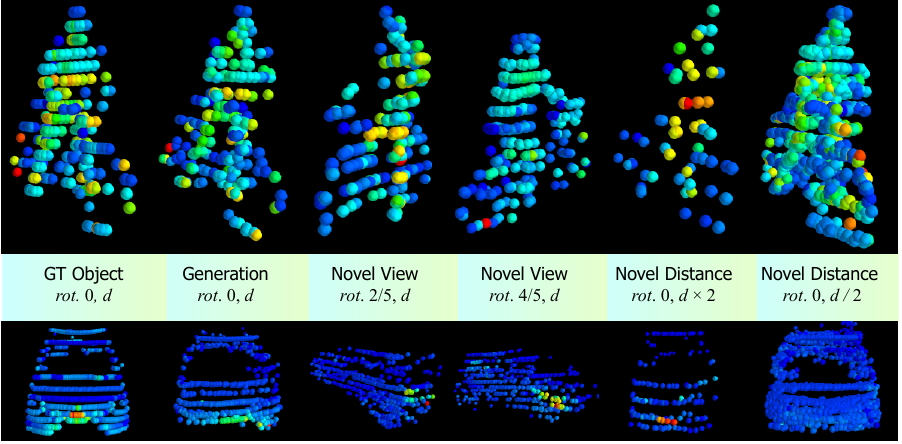}
\captionof{figure}{\textbf{Novel LiDAR objects generated by \ours} trained on the nuScenes train set, produced using the conditioning information of real objects from the nuScenes validation set. 
Relative to the sensor, $rot$ is the object's rotation while $d$ is the distance. Point color is according to LiDAR intensity. More examples are in the appendix.}
\label{fig:teaser}
\vspace{3mm}
}]

\begin{abstract}
   The generation of LiDAR scans is a growing topic with diverse applications to autonomous driving. However, scan generation remains challenging, especially when compared to the rapid advancement of image and 3D object generation. We consider the task of LiDAR object generation, requiring models to produce 3D objects as viewed by a LiDAR scan. It focuses LiDAR scan generation on a key aspect of scenes, the objects, while also benefiting from advancements in 3D object generative methods. We introduce a novel diffusion-based model to produce LiDAR point clouds of dataset objects, including intensity, and with an extensive control of the generation via conditioning information. Our experiments on nuScenes and KITTI-360 show the  quality of our generations measured with new 3D metrics developed to suit LiDAR objects. The code is available at \url{https://github.com/valeoai/LOGen}.
\end{abstract}

\section{Introduction}
\label{sec:introduction}

As 2D generative methods have exploded in their capacity to generate diverse and high quality images \cite{peebles2023scalable}, one area of focus for 3D object generation has been understudied: objects in the context of LiDAR sensors. LiDAR is an essential modality for autonomous agents in driving scenarios. It provides accurate 3D information about the environment,
making it crucial for safe navigation. The utility and corresponding shortcomings of LiDAR datasets \cite{liu2024survey,muhammad2022vision} makes the generation of LiDAR objects interesting from a variety of perspectives: sensor level generation for simulation \cite{manivasagam2020LiDARsim}, rare class augmentation \cite{zhang2022pointcutmix}, out-of-distribution generation \cite{loiseau2024reliability}, corner case generation \cite{li2022coda}, object disentanglement \cite{niemeyer2021giraffe}, etc.

Objects perceived by LiDAR sensors are distinct from classically rendered 3D objects in the impact that the relative angle and distance of objects to the sensor has on the final appearance of an object \cite{zyrianov2022lidargen}. Further, LiDAR objects contain an additional channel of information when compared to classic 3D objects: the intensity of the laser beam reflected back to the sensor. Models which generate LiDAR objects must be capable of using this relative positioning and creating the intensity channel in order to generate realistic objects \cite{viswanath2024reflectivity}.

The generation of LiDAR data has been explored most extensively at the scene level, where many works have attempted to generate entire novel scans of autonomous vehicles in new surroundings \cite{zyrianov2022lidargen,zyrianov2024LiDARdm,xiong2023ultraLiDAR}. However LiDAR scene generation remains a difficult task, with many works turning towards simplifying assumptions to focus the generative task onto more tractable footing \cite{ran2024lidm,nunes2024lidiff,nakashima2023r2dm,wu2024text2LiDAR,yue2018LiDARpcgen}. We propose to bring recent advancements in generative methods \cite{chen2023pixart,mo2023dit} to the domain of LiDAR data, focusing on the generation of high quality and realistic LiDAR objects.

We introduce a transformer-based diffusion model, dubbed \ours (for LiDAR Object Generator), which is designed to generate LiDAR scans of new objects. The generation is parameterized by the object class, angle, and distance from which objects are seen by the LiDAR. This conditioning allows fine grained control of the appearance of generated objects.

We compare \ours against suitable alternatives that we design, inspired from other tasks: a sparse 3D CNN (\minkunet \cite{choy20194d}); a 2D transformer-based generator (\pixartal \cite{chen2023pixart}) adapted to 3D; and a 3D transformer-based generator (\dittd \cite{mo2023dit}), originally designed for traditional 3D objects, which we adapt to LiDAR objects. We also compare to concurrent work \cite{yan2025olidm}. 
We show the quality of our generated objects compared to these baselines. We also evaluate at scene level on the semantic segmentation task to assess the realism of objects we generate. Examples of generation are shown in Fig.\,\ref{fig:teaser}.
Our contributions are as follows:
\begin{itemize}[topsep=1pt,itemsep=-3pt]
    \item We propose a diffusion-based method specifically designed to produce LiDAR point clouds of objects, including intensity and with an extensive control of generation (conditioning). 
    
    \item We propose a new transformer-based LiDAR object diffusion model, operating on 3D points, as well as three other diffusion-based baselines.
    
    \item Our experiments show that despite the compact size of our model (7.5M parameters), we achieve superior quality, which we also measure with new metrics suited to LiDAR objects. 
\end{itemize}
We believe that the LiDAR object generation task is ripe for future exploration, and that the promising results we present here will serve as a basis for this task.

\section{Related Work}
\label{sec:relatedwork}
\noindent{\textbf{3D object generation.}} Various techniques have been proposed for 3D object generation: GANs \cite{achlioptas2018learningrepres, shu2019treegan}, VAEs \cite{li2022editvae, vahdat2022lion}, normalizing flows \cite{yang2019pointflow}, diffusion at point level \cite{luo2021diffusion, tyszkiewicz2023gecco}, and diffusion on features at voxel level with a transformer architecture \cite{mo2023dit}. However, these methods are only typically evaluated on datasets such as ShapeNet \cite{shapenet2015}, ModelNet \cite{wu2015modelnet}, or CO3D \cite{reizenstein2021common} after sampling points uniformly on shape surfaces, rather than following LiDAR sampling patterns. Their behavior with uneven point distributions of (sensor-specific) LiDAR scans is unexplored. 
Besides, these methods do not use geometric conditioning related to viewpoint and do not generate point intensities. Further, most of these methods are trained and evaluated with a high and fixed number of points: typically 2048. It is not clear how they would behave with varying and sparse point clouds, as is the case of LiDAR objects.

\medskip

\noindent{\textbf{LiDAR data generation.}}
Methods have been proposed to generate full LiDAR scans, by LiDAR simulation \cite{gschwandtner2011blensor, fang2020augmentedLiDAR, fang2021LiDARaug} in synthetic environments \cite{yue2018LiDARpcgen}, or by learning to generate LiDAR data using energy-based models \cite{zyrianov2022lidargen}, VQ-VAEs \cite{xiong2023ultraLiDAR}, diffusion in range images \cite{nakashima2023r2dm, hu2024rangeldm, ran2024lidm, buburuzan2025mobi} or diffusion on 3D points \cite{nunes2024lidiff}, possibly with images \cite{zhang2023nerfLiDAR}. Further, simulation and learning can be combined to improve realism \cite{manivasagam2020LiDARsim, zyrianov2024LiDARdm}. Some scan generation methods incorporate objects via: conditioning on a semantic map \cite{ran2024lidm}; placing learned codes of object classes into a canvas representing the scene before decoding \cite{xiong2023ultraLiDAR}; in-painting objects into range images \cite{buburuzan2025mobi};  and using text-based conditioning \cite{wu2024text2LiDAR}. However these methods do not focus on the visual fidelity of generated objects, do not directly learn object appearance and intensity from the data at an object level, and do not evaluate the object perceptive quality.

OLiDM \cite{yan2025olidm} is a concurrent work on LiDAR object generation that highlights the importance of distinguishing between foreground and background objects, employing a diffusion transformer model to generate LiDAR objects. They allow conditioning on object viewpoint parameters to produce localized objects, and perform scene level generation using the generated objects as conditions. \ours in contrast focuses on architecture level decisions and uses novel methods of incorporating conditioning information and encoding points. \ours features more robust evaluation of generated objects, with classwise metrics and perceptual evaluations including generated intensities.

\section{Method}
\label{sec:method}
The goal of the \emph{LiDAR object generation task} is to learn, for known classes in a labeled LiDAR dataset, to generate 3D point clouds of novel objects, as they would be captured by the same LiDAR sensor from arbitrary viewpoints. As opposed to simulation, the training distribution is given by real-world objects extracted from 3D scans. 

We train a diffusion model to generate point clouds of novel objects. The diffusion process is applied directly on points given as input, and is conditioned by geometric information concerning the object to generate.

\subsection{Object parameterization}

An object in a LiDAR dataset is given by its class $c$, a set of 3D points with intensities, as well as information on the object bounding box, namely, the box center $(x,y,z)$, the box length, width and height $(l, w, h)$ and the box yaw~$\psi$, i.e., the orientation around the vertical z axis (azimuth) of the heading of the object.

\paragraph{Parameters.} To reduce the number and variability of input and conditioning parameters when training a model, we normalize the representation of objects with respect to the sensor location. 
For this, the Cartesian coordinates of object points are made relative to the oriented box center, i.e., relative to the box center and rotated around the z axis to align with the object box heading. The intensities are scaled according to the log-max.

We also decrease the number of parameters regarding the box center location and orientation as the object appearance, seen from the sensor, only depends on the distance $d$ to the object and on the observation angle $\phi$, i.e., the angle between the object heading $\psi$ and the ray from the object bounding box center to the sensor. (As an illustration, $\phi\,{=}\,0$ means the object is facing the sensor, regardless of the object location around the ego vehicle.) In this conversion from Cartesian to cylindrical coordinates, the altitude $z$ of the box center is kept.

In the end, the data distribution to learn consists of objects represented by a set of points $s = \{P_i\}_{i=1}^N$ in $\mathbb{R}^{N \times 4}$ with (centered) Cartesian coordinates and intensities.  We do not normalize the point coordinates to preserve 
the absolute dimensions of objects and the specificity of the LiDAR scanning patterns, depending in particular on the distance to the sensor.

Last, we condition the generation on the following box information: $\kappa = (\phi, d, z, l, w, h)$. While our generation is classwise, class $c$ is not directly included here as we train one model per class. This choice was made 
for compute and efficiency reasons: the per sample compute of single vs multi-class models is equivalent, thus allowing us to parallelize training. We nevertheless also experimented with a multi-class model, which was lagging behind the single-model-per-class approach in realism metrics (see the appendix).

\paragraph{Point embedding.} The way to embed points for diffusion has strong implications on the memory consumption and realism of generated outputs. 

DiT-3D \cite{mo2023dit} introduces voxel embeddings that efficiently structure and aggregate points into patches. This technique, alongside window attention, allows the model to compute attention across shapes of thousands of points. The de-voxelization procedure places points in a final position inside their voxel, based on a weighted average of their activations. It has the effect of creating uniform and smooth point clouds across the underlying surface. Such uniform distributions are ideal for the goal of producing a continuous 3D mesh, but do not match the banded pattern observed when a LiDAR captures objects in the real world. 

To address this issue, we introduce a simple embeddings based on PointNet \cite{qi2017pointnet}. These features are just the output of a single 1D convolution, with which we project from our input dimension (3+1) into the embedding dimension. Because the point clouds are in general magnitudes smaller 
(e.g., averaging from 40 points for bikes to 463 for bus in nuScenes)
than the point clouds generated in the original DiT-3D paper (thousands of points), we can use standard memory-efficient attention layers and avoid the need for voxelization. In the experiment section, we examine the impact of both types of embeddings.

\paragraph{Conditioning encoding.} A key stage when using diffusion is the encoding of the conditioning and time step. Classically, we rely on Fourier features \cite{tancik2020fourier}, which involves a higher dimensional projection followed by a sine/cosine encoding of the corresponding features. This applies to all scalar values of the condition but angle~$\phi$. To account for the periodic discontinuity of representation for an angle, we encode $\phi$ using the cyclical embeddings from \cite{lee2023spatio}, which preserves the cyclical nature of angular data. In our experiments, we train one network per class; we thus do not need to encode the class.

\begin{figure*}[t!]
    \centering
    \includegraphics[width=\linewidth]{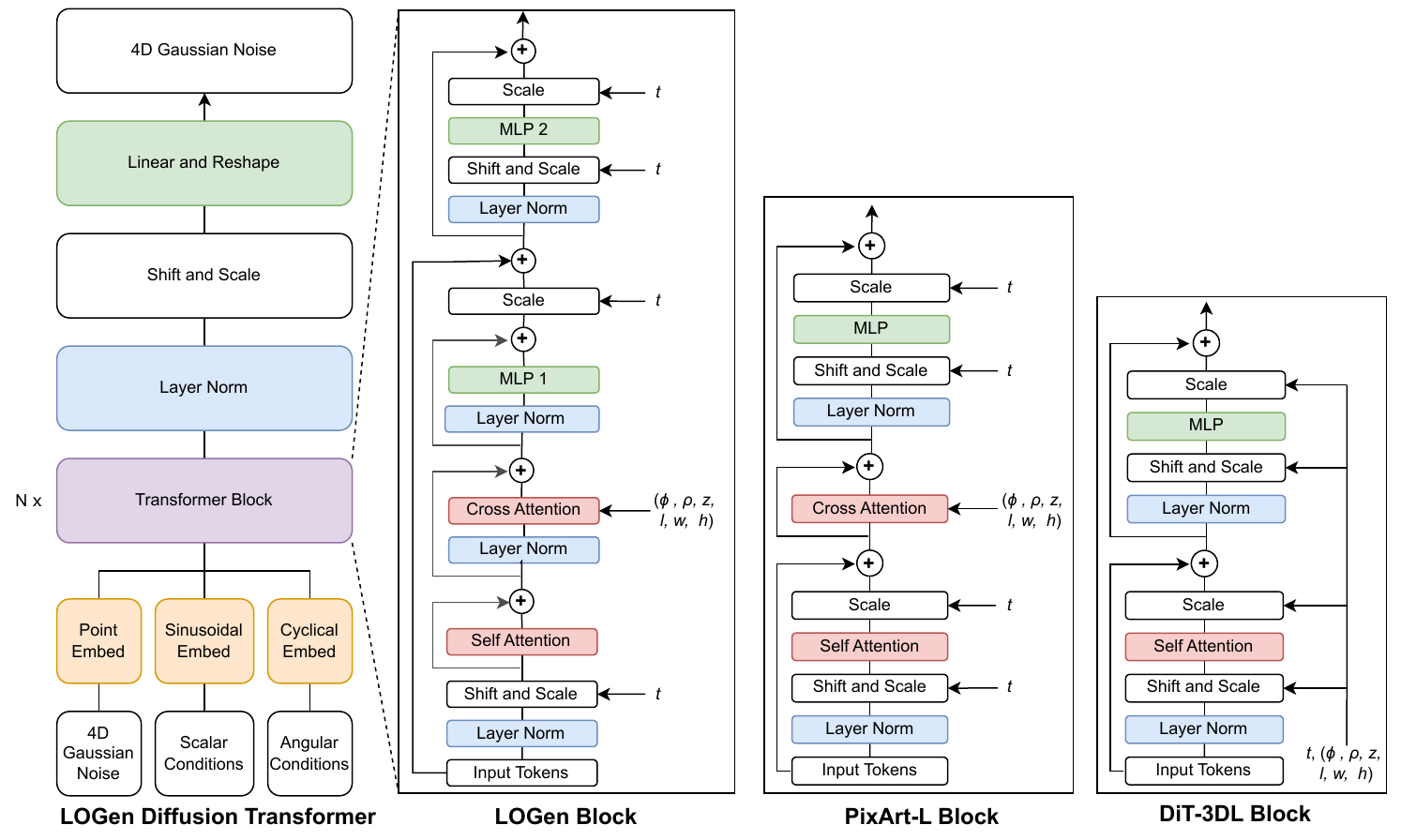}
    \caption{\textbf{Architecture of \ours and baselines \pixartl and \dittdl.}}
    \vspace*{-10pt}
    \label{fig:logen}
\end{figure*}

\paragraph{LiDAR object diffusion.} Given a dataset with a number of object point clouds $S \,{=}\, \{s_j\}_{j=1}^M$ alongside conditions $K \,{=}\, \{\kappa_j\}_{j=1}^M$,
the objective is to learn the conditional distribution $p(S|K)$. This is done via a Denoising Diffusion process: we learn to denoise input 4D Gaussian noise ($N$ noisy points) into a target shape ($N$ points representing a scanned object, with intensities).

We rely on the time-discrete formulation of the diffusion process known as Denoising Diffusion Probabilistic Models (DDPM) \cite{ho2020DDPM}, which is also used in DIT-3D \cite{mo2023dit} and in LiDAR scene generation methods \cite{nunes2024lidiff}. 
During training, we sample a random time step $t$ and produce, via a forward noising process, a noisy sample $s_t$ of an input object point cloud $s_0 = s$.
Starting from a real LiDAR object $s_0$, the forward noising procedure produces a series of increasingly noisy 4D point clouds $s_1,\ldots,s_t$, finally arriving at $s_T\,{\sim}\,\mathcal{N}(\mathbf{0}, \mathbf{I})\,{\in}\,\mathbb{R}^4$. The training objective is to learn a distribution of the form $p_{\theta}(s_{t-1}|s_t)=\mathcal{N}(s_{t-1}|\bm\mu_{\theta}(s_t, t), \sigma^{2}_t\mathbf{I})$, parameterized by a denoising neural network. 

The training loss optimizes the evidence variational lower bound, minimizing the Kullback-Leibler divergence between the forward and reverse processes. After simplification, it boils down to minimizing a mean-squared loss:
$ \arg \min_{\theta} ||\epsilon_0 - \hat{\epsilon}_{\theta}(x_t, t)||^2_2 $ between the model output $\hat{\epsilon}_{\theta}(s_t, t)$ and the ground-truth Gaussian noise~$\epsilon_0$. Given $\hat{\epsilon}_{\theta}(s_t, t)$, we can then compute $\bm\mu_{\theta}(s_t, t)$ and get a denoising step recurrence equation defining $s_{t-1}$ from $s_{t}$.

To exploit conditioning, the above formulation is extended with the extra input of a condition $\kappa$, leading to a denoising network $\hat{\epsilon}_{\theta}(s_t, t, \kappa)$.
To apply the conditional guidance, we follow the Classifer Free Guidance formulation from \cite{ho2022classifier}. This implies the dropout of conditioning information during the training procedure, where all conditions are replaced by a pre-defined null embedding, which results in an unconditioned pass of the model. 

\subsection{Architecture}
\label{sec:archi}

We now describe the architecture of \ours and of our baselines, that we illustrate on Fig.\,\ref{fig:logen}.
\dittd \cite{mo2023dit} is straightforwardly derived from DiT \cite{peebles2023scalable}, and thus shares a similar architecture. \pixartal \cite{chen2023pixart} adds cross attention to DiT to inject conditions. Our baselines \dittdl and \pixartl are directly adapted from their original counterparts to handle: (i)~LiDAR data, i.e., point embedding, and (ii)~LiDAR object conditioning.  \ours is then a further modification of \pixartl, as described below.
The conditioning $\kappa = (\phi, d, z, l, w, h)$ is applied in the denoising network to direct the generative process. Given the importance of the conditioning in our task, we explore three variants in the architecture design space.
\begin{itemize}

    \item \emph{AdaLN-Zero} is the best-performing guidance method proposed in DiT \cite{peebles2023scalable}, and reused in \dittd. In our case, for \dittdl, the six conditioning values are concatenated to the time step and used to learn the adaptive layer-norm shift and scale parameters.

    \item \emph{AdaLN-single} is the guidance proposed and used in \pixartal \cite{chen2023pixart}. While in the adaptive layer norm the condition is used to learn global shift and scale parameters, cross-attention layers allow the model to learn the guidance on a local scale. We do the same in \pixartl.

    \item We observe that the separation of the self-attention layers from the cross-attention layers by a scale operation in the \pixartl architecture degrades the generation results (cf.\ Tab.\,\ref{tab:combined-metrics}). Thus we introduce a third form of conditioning in \ours blocks, where we apply the cross attention immediately after the self-attention layer, and scale after these operations. 
\end{itemize}

\section{Experiments}
\label{sec:experiments}
\subsection{Experimental Setup}

\paragraph{Datasets.} We evaluate our approach on \textbf{nuScenes} \cite{caesar2020nuscenes}, which was acquired with a 32-beam LiDAR, thus creating notably sparse sweep patterns. To construct object samples to learn from, we use the semantic point labels and the 3D box annotations of objects. We take as object instance all points falling into a box, filtering out possible points of a different class. We also discard objects with less than 20 points. Point intensity, which ranges in 0--255, has a highly irregular distribution, with most examples clustered from 0--30, and with outliers. To create a more uniform intensity distribution, we use log max scaling.

In total, we extract 268,684 objects with at least 20 points, which covers more than 95\% of all object points.\,(Similarly, the official instance segmentation evaluation of nuScenes only considers instances with at least 15 points~\cite{caesar2020nuscenes}.)
In order to unify the number of points per cloud in a batch, we pad each point cloud to the max length in the batch. 

We further evaluate on objects extracted from \textbf{KITTI-360} \cite{kitti360}. We use semantic and instance labels from accumulated scans to extract roughly 40k objects in 3 classes as in~\cite{yan2025olidm}: \textit{car}, \textit{person}, \textit{bicycle}. We also perform cross-dataset evaluation on \textbf{SemanticKITTI} \cite{semantickitti}.

For all experiments, we use the conditioning information from objects in the validation set, generating similar views of new objects of the same category. 

\paragraph{Object-level metrics.}

To evaluate the ability to generate a sensible variant of a real object given only the conditioning information corresponding to the real object, we use two pointset metrics: Chamfer Distance (\textbf{CD}) and Earth Mover's Distance (\textbf{EMD}).

To evaluate perceptual realism, we use the Fréchet PointNet Distance (\textbf{FPD}) \cite{shu2019treegan}, an adaptation of the classic Fréchet Inception Distance (FID) \cite{heusel2017fid}, by replacing 2D Inception features with 3D PointNet features \cite{qi2017pointnet}. However, the original FPD used as reference relies on the original PointNet classifier, trained on Model\-Net\-40 \cite{qi2017pointnet}, which is not suited for LiDAR objects. We thus use instead a new PointNet classifier trained on the real LiDAR objects of nuScenes. We additionally consider the Kernel Inception Distance (KID), which has advantages over FID, in particular for small datasets \cite{binkowski2018kid}. Like for FID, we adapt KID into a Kernel PointNet Distance (\textbf{KPD}), replacing the Inception network by our trained PointNet.

We also consider distribution comparisons at a pointset level. We use coverage (\textbf{COV}) \cite{achlioptas2018learningrepres}, which measures the proportion of real pointsets matched to at least
one generated pointset, and 1-nearest neighbor accuracy (\textbf{1-NNA}) \cite{yang2019pointflow}, derived from \cite{lopezpaz2018revisiting}, which merges real and synthetic objects and counts the proportion of closest neighbors of each sample that are of the same source. 1-NNA ranges from 100\% when distributions are separate, to 50\% when they are indistinguishable. COV and 1-NNA are measured with CD and EMD.
We extend COV and 1-NNA to measure the quality of the generated intensities. We extract 256-dimensional normalized intensity histograms from generated objects to use as features for COV and 1-NNA computations.
Last, in \cite{yan2025olidm}, generated objects are extracted from generated scenes after being found by a 3D detector. On KITTI-360, we compare our generations to such detected objects, using object level metrics given in \cite{yan2025olidm}.
The Appendix details all metrics.

\begin{table*}[!t]
\centering
\setlength{\tabcolsep}{4pt}
\begin{tabular}{l @{~} c | c c | c c c | c c c | c c | c c}
\toprule
\multirow{3}{*}{\textbf{Model}} & 
\multirow{3}{*}{\textbf{Emb.}} & 
\multicolumn{2}{c|}{\textbf{Point metrics}} & 
\multicolumn{3}{c|}{\textbf{1-NNA\,(\%)\,\,$\downarrow$}} & 
\multicolumn{3}{c|}{\textbf{COV\,(\%)\,\,$\uparrow$}} & 
\multicolumn{2}{c|}{\textbf{FPD\,$\downarrow$}} & 
\multicolumn{2}{c}{\textbf{KPD\,$\downarrow$}} \\
& & CD & EMD 
  & CD & EMD & Int. 
  & CD & EMD & Int. 
  & 3 ch. & 4 ch. 
  & 3 ch. & 4 ch. \\
\midrule
\minkunet & - 
  & 0.208 & 0.212 
  & 79.5 & 74.4 & \NA 
  & 33.7 & 41.7 & \NA 
  & 25.4 & \NA 
  & 3.18 & \NA \\

\dittdl & \VE 
  & 0.136 & 0.121 
  & 75.8 & 70.9 & \NA 
  & 34.3 & 42.0 & \NA 
  & 2.80 & \NA 
  & 0.25 & \NA \\

\dittdl & \PE 
  & 0.142 & 0.125 
  & \textbf{73.4} & 70.7 & 74.4 
  & 35.5 & 42.1 & 31.4 
  & 2.62 & 6.99 
  & 0.25 & 0.75 \\

\pixartl & \PE 
  & 0.196 & 0.204 
  & 76.5 & 72.5 & 71.6 
  & 33.5 & 41.8 & 32.9 
  & 2.95 & 5.16 
  & 0.24 & 0.41 \\

\ours & \PE 
  & \textbf{0.130} & \textbf{0.111} 
  & 74.1 & \textbf{70.6} & \textbf{68.2} 
  & \textbf{36.0} & \textbf{42.3} & \textbf{37.0} 
  & \textbf{1.34} & \textbf{2.18} 
  & \textbf{0.10} & \textbf{0.12} \\
\bottomrule
\end{tabular}
\captionof{table}{\textbf{Combined point-space and feature-space evaluation.} CD and EMD are geometric fidelity metrics. 1-NNA and COV (in CD/EMD/intensity space) assess realism and coverage. FPD and KPD evaluate feature distribution alignment under 3- and 4-channel settings (= 3 ch.\ + intensity). Point embedding (Emb.) is: none (-), voxel-based (\VE) or PointNet-based (\PE).}
\label{tab:combined-metrics}
\vspace*{-2pt}
\end{table*}

\begin{figure*}[t]
\vspace{8pt}
\captionsetup[subfigure]{labelformat=empty}
    \centering
    \resizebox{0.9\linewidth}{!}{
    \begin{tabular}{c@{~~}c@{~}c@{~~~~}c@{~}c@{~~}c}
        \rotatebox{90}{\hspace*{8mm}Truck}&
        \includegraphics[width=0.235\linewidth]{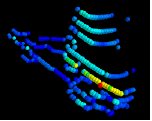} &
        \includegraphics[width=0.235\linewidth]{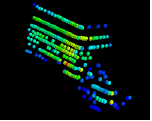} &
        \includegraphics[width=0.25\linewidth]{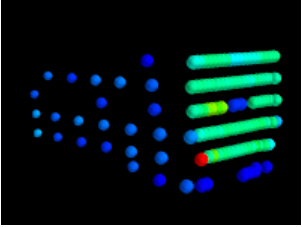} &
        \includegraphics[width=0.25\linewidth]{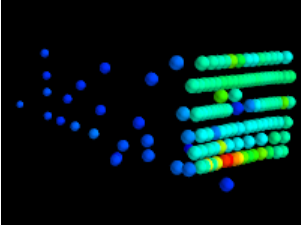} &
        \rotatebox{90}{\hspace*{10mm}Bus}
        \\
        \rotatebox{90}{\hspace*{3mm}Traffic cone} &
        \includegraphics[width=0.235\linewidth]{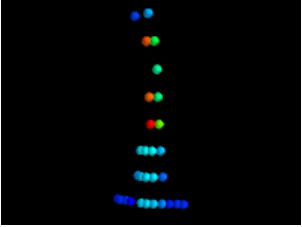} &
        \includegraphics[width=0.235\linewidth]{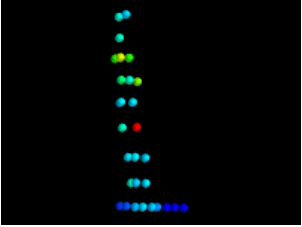} &
        \includegraphics[width=0.25\linewidth]{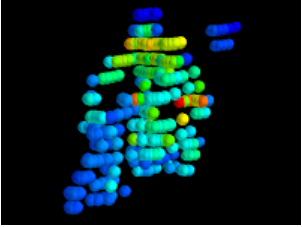} &
        \includegraphics[width=0.25\linewidth]{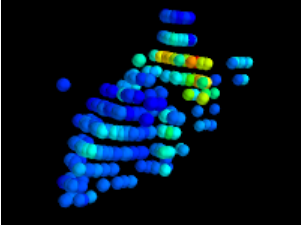} &
        \rotatebox{90}{\hspace*{6mm}Motorbike}
        \\
    \end{tabular}
    }
\caption{\textbf{Generations with intensities.} For each pair, the left object is from the nuScenes val set, the right object is generated using the same conditioning and number of points.}
\label{fig:nuscene-intensities}
\end{figure*}
\paragraph{Scene-level metrics.}

A second set of evaluations is conducted on the perception of generated objects in a scene context by a typical 3D segmentation model with the contention that \textbf{more realistic objects should be segmented at a similar rate to real objects}. We consider the replacement of ground-truth objects by objects generated with the same conditioning, then measure the change in classwise \textbf{mIoU} segmentation performance of the model on this modified data. We use a variant of nuScenes where each object with at least 20 points is replaced by a generated object of the same class and box size, seen under the same viewing angle, and with the same number of points. We highlight the correspondence between this metric and the \#Box metric from \cite{yan2025olidm}, where both evaluate the expected performance at identifying generated objects of a model trained to identify real objects.

As segmenter, we use the popular SPVCNN \cite{tang2020spvcnn}, which 
has been widely used in evaluation settings \cite{unal2023dial, samet2023seedal, xie2023annotator}.
We consider two different versions: \textbf{SPVCNN\tch} trained on points coordinates only, while \textbf{SPVCNN\fch} is also trained using point intensities.

\paragraph{Baselines.} Due to task novelty, we build baselines:
\textbf{\minkunet} is the sparse convolutional architecture used in LiDiff \cite{nunes2024lidiff}, which ignores intensities. \textbf{\pixartl} is our adaptation of \pixartal to LiDAR points and conditioning, using our PointNet embeddings (PE) to allow intensity generation. \textbf{\ditve} is our adaptation of \dittd to LiDAR conditioning, i.e., with its original voxel embeddings (VE) and no intensity handling. (The voxelization library used in \dittd \cite{mo2023dit} does not handle a 4th input channel; handling intensities would require adapting the voxelization layer like in \minkunet \cite{choy20194d}, as separate input features.) \textbf{\ditpe} replaces VE by our PE in \ditve\!\!\!, thus handling intensities.

\subsection{Object-level evaluation}
\begin{figure}[t]
\centering
\begin{minipage}[t]{0.7\linewidth}
    \centering
    \resizebox{\linewidth}{!}{
    \begin{tabular}{l | c c }
        \toprule
        \textbf{Method} & CD\,$\downarrow$ &  JSD\,$\downarrow$ \\
        \midrule
        LiDARGen \cite{zyrianov2022lidargen}   & 33.10 &  0.41  \\
        UltraLiDAR \cite{xiong2023ultraLiDAR} & 18.30 & 0.70\\
        R2DM \cite{nakashima2023r2dm}       & 17.60 & 0.86 \\
        OLiDM \cite{yan2025olidm}      &  1.88 & 0.91  \\
        \midrule
        \textbf{LOGen (ours)} & \textbf{0.24} & \textbf{0.19}  \\
        \bottomrule
    \end{tabular}
    }
\end{minipage}
\hfill
\begin{minipage}[t]{\linewidth}
\renewcommand{\arraystretch}{0.2}
    \centering
    \vspace{5pt}
    \begin{tabular}{@{\,\,}c@{~}c@{~}c@{~}c}
        \includegraphics[height=0.27\linewidth,trim={90 0 90 0},clip]{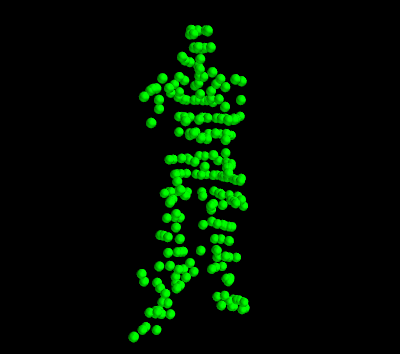} &
        \includegraphics[height=0.27\linewidth,trim={90 0 90 0},clip]{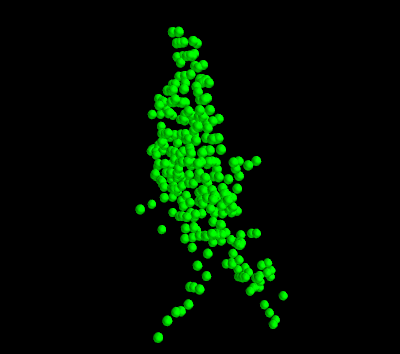} &
        \includegraphics[height=0.27\linewidth]{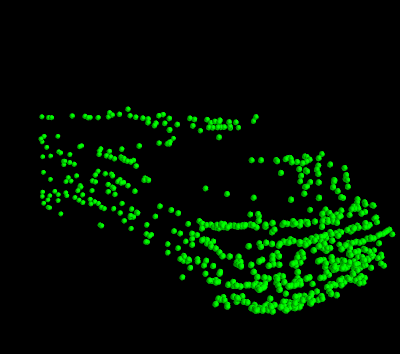} &
        \includegraphics[height=0.27\linewidth]{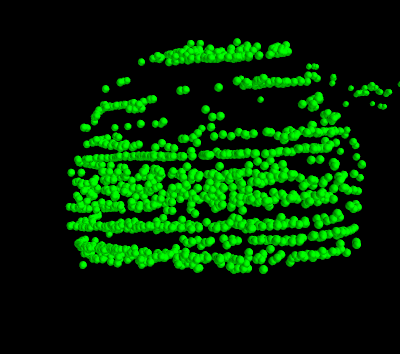} \\
    \end{tabular}
\end{minipage}
\caption{\textbf{Quantitative \& qualitative comparison of LiDAR objects on KITTI-360 \cite{kitti360}.} (Top) Evaluation on 1000 sampled objects (for \ours) or detected objects (for others) using Chamfer Distance (CD) and Jensen-Shannon Divergence (JSD). (Bottom) Examples of generated objects from KITTI-360 for \textit{person} and \textit{car}. Classwise results of \ours and additional metrics EMD, FPD, and KPD are in the appendix.
}
\label{tab:table_intensity_combined}
\end{figure}

Qualitative generation examples are shown in Figs.\,\ref{fig:teaser}, \ref{fig:nuscene-intensities}, \ref{tab:table_intensity_combined}.
Fig.~\ref{fig:visual-all-models} presents qualitative comparisons between \ours and the baseline methods for nuScene classes.
In Fig.~\ref{fig:visual-augs}, we present novel objects produced by LOGen.

\paragraph{Point embeddings.} Table~\ref{tab:combined-metrics} shows that the voxel embeddings of \dittd do not perform as well as our PointNet embeddings, which supports our proposal (Sect.\,\ref{sec:archi}). \minkunet succeeds in generating local structure (LiDAR pattern), but fails to generate objects meeting size conditions. The lack of point embeddings in \minkunet creates global size inconsistencies.

\paragraph{Architecture.} Table~\ref{tab:combined-metrics} also shows that \ours transformer blocks perform slightly better than the blocks proposed in \pixartal and \dittd. The superiority of \ours blocks is more pronounced as measured by the feature-based metrics of FPD and KPD.

\paragraph{Comparison on KITTI-360.} The table in Fig.\,\ref{tab:table_intensity_combined} shows the superior visual fidelity of objects generated by LOGen when compared to other scene and object generative methods.

\subsection{Scene-level evaluation}

\begin{table*}[!t]
\centering
\setlength\extrarowheight{-0.5pt}
\begin{tabular}{l l@{~}c |@{~~~}c c c c c c c c c c | c }
\toprule
& Model  & Emb. & \clap{barrier} & \clap{bicycle} & bus & car & \clap{\!con.veh.}  & \clap{moto.}  & \clap{pedes.} & \clap{tra.con.} & \clap{trailer} & \clap{truck} & mean\\
\midrule

\parbox[t]{2mm}{\multirow{6}{*}{\rotatebox[origin=c]{90}{3 channel}}} & \minkunet & - & 76.1 & 31.4  & 18.7 & \textbf{84.4} & 14.7 & 63.6 & 69.8 & 59.3  & 12.4 & 49.7  & 48.0  \\
& \dittdl   & \VE &  48.4 & 42.5 & 0.3 & 34.1 & 0.2 & 63.8 & 75.6 & 69.6 & 7.7 & 3.4 & 34.6 \\
& \dittdl   & \PE & 61.6 & 39.2 & 11.1 & 84.2 & 1.7 & 68.8 & 77.5 & 51.3 & 15.9 & 26.6 & 43.9 \\
& \pixartl  & \PE & 58.9 & 32.1 & 3.9 & 75.7 & 0.9 & 76.6 & 79.0 & 56.7 & 13.6 & 51.3 & 44.9 \\
& \ours     & \PE & \textbf{78.2} & \textbf{50.9 }& \textbf{60.1} & 79.1 & \textbf{28.1 }& \textbf{82.2 }& \textbf{84.2} & \textbf{73.5} & \textbf{37.5} & \textbf{63.2} & \textbf{63.7 }\\
\cmidrule(lr){2-14}
& GT        &  &  80.3 & 48.8 & 90.7 & 94.2 & 40.6 & 86.0 & 85.5 & 73.4 & 60.7 & 85.2 & 74.5\\
\midrule
\parbox[t]{2mm}{\multirow{4}{*}{\rotatebox[origin=c]{90}{4 channel}}} 
& \dittdl   & \PE & 62.5  & 29.0  &  18.3 & 85.9  & 6.3  &  62.6 & 68.4 & 47.2  & 15.9  & 37.7  & 43.4 \\
& \pixartl  & \PE & 56.6  & 33.7  &  14.5 & 82.4  & 8.1  &  78.5 & 79.3 & 62.8  & 26.7  & 52.6  & 49.5 \\
& \ours     & \PE & \textbf{78.0}  & \textbf{48.9 } &  \textbf{71.7} & \textbf{91.3}  & \textbf{37.1}  &  \textbf{82.0} & \textbf{84.8} & \textbf{79.1}  & \textbf{45.1}  & \textbf{72.9 } & \textbf{69.1} \\
\cmidrule(lr){2-14}
& GT        &  &  81.6   & 50.4 & 90.9 & 94.6 & 44.5 & 88.1 & 85.6 & 77.4 & 64.6 & 85.6 & 76.3\\
\bottomrule
\end{tabular}
\hspace*{6mm}
\captionof{table}{\textbf{Scene-level evaluation.}
    SPVCNN is trained with 3 or 4 channels (coordinates +\,intensity) on nuScenes (train set) with real objects. It is tested on nuScenes (val set) where each object (with $\ge$ 20 points) 
    is replaced with a generated object of the same class, box size and sensor viewing angle, and with the same number of points, and where objects with less than 20 points are removed from the dataset. Displayed results are class-wise IoU\%.}
    \label{tab:scenecompare}
    
\end{table*}

\begin{table*}[!t]
   \centering
    \setlength{\tabcolsep}{4pt}
    \hspace*{-1ex}\begin{tabular}{l @{~} |@{~~} c c | c c c c | c c c c | c c}
        \toprule
          & \multicolumn{2}{c|}{Point metrics} & \multicolumn{2}{c}{\!\!\relax{1-NNA}\,$\downarrow$}  &\multicolumn{2}{c|}{\relax{COV}\,$\uparrow$} &
         \multicolumn{2}{c}{FPD\,$\downarrow$}  & \multicolumn{2}{c|}{\!\!KPD\,\,$\downarrow$} & \multicolumn{2}{c}{APC\,\,$\uparrow$}
         \\
         \logen\PE & \!\relax{CD}\,$\downarrow$ & {EMD}\,$\downarrow$ 
         & \!\it CD & \it EMD & \it CD & \it EMD
         & 3\,ch. & 4\,ch. & 3\,ch. & 4\,ch. & 3\,ch. & 4\,ch. 
         \\
        \midrule
        Single-model-per-class & \textbf{0.130} & \textbf{0.111} & 74.1 &	\relax{70.6} &	\relax{36.0} &	\relax{42.3} &
        \textbf{1.34} & \textbf{2.18}	& \textbf{0.10} & \textbf{0.12} & \textbf{40.9} & \textbf{48.1}
        \\
        Multi-class model & \relax{0.149} & \relax{0.134} & \textbf{72.8} &	\textbf{68.9} &	\textbf{37.1} &	\textbf{43.3} &
        2.35 & 4.28 & 0.21 & 0.45 & 36.2 & 42.5
        \\
        \bottomrule
    \end{tabular}
\caption{\textbf{Single-model-per-class vs. multi-class model.} Comparison of a same size single-model-per-class and multi-class models on object generation metrics. The multi-class model lags behind the single-model-per-class approach in realism metrics.}
    \label{tab:multiclass}
    \vspace{-8pt}
\end{table*}
\paragraph{Semantic quality of generated data.}

Table~\ref{tab:scenecompare} shows that the \ours generations are more semantically identifiable from the point of view of an SPVCNN segmenter. Though trained on real objects, the segmenter is able to recognize the semantics of \ours objects at a much higher mIoU than other models, +19.6 pts vs the second best model. This is particularly pronounced for rare classes such as bicycle, bus, trailer, and construction vehicle. Besides, intensities generated by \ours are realistic enough to provide +5.4 mIoU\% at scene level.

\begin{figure*}
\captionsetup[subfigure]{labelformat=empty}
\centering
\setlength\tabcolsep{1.5pt} 
\resizebox{1.0\textwidth}{!}{
\begin{tabular}{rccccc}

GT~~~~~~~~~~~~~~  & 
\ours &
\dittdl \PE  &
\dittdl \VE &
\pixartl  &
\minkunet  \\

{\rotatebox[origin=t]{90}{\!\!\!barrier\vphantom{p}}} 
\includegraphics[width=0.20\textwidth,  valign=m, keepaspectratio,] {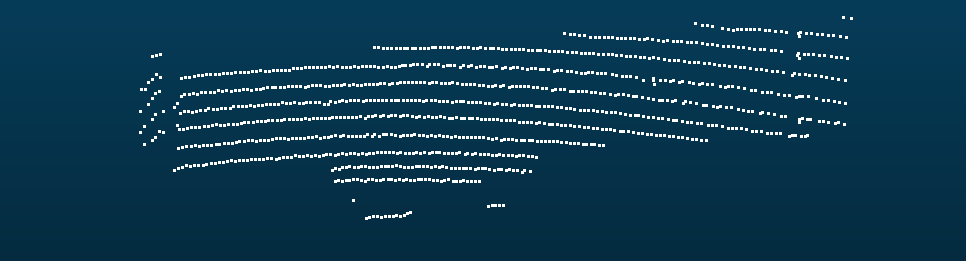} &
\includegraphics[width=0.20\textwidth, valign=m,  keepaspectratio,] {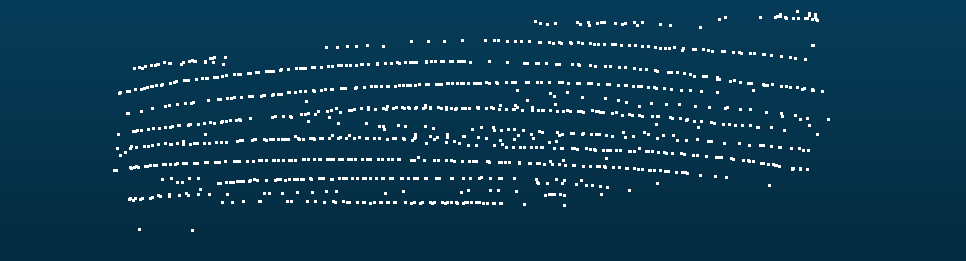} &
\includegraphics[width=0.20\textwidth, valign=m,  keepaspectratio,] {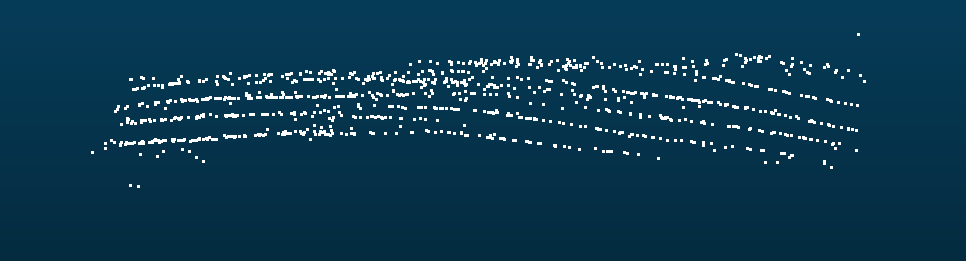} & 
\includegraphics[width=0.20\textwidth,  valign=m, keepaspectratio,] {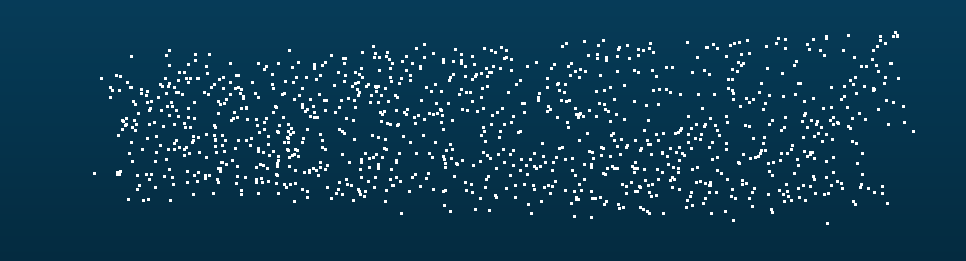} &
\includegraphics[width=0.20\textwidth,  valign=m, keepaspectratio,] {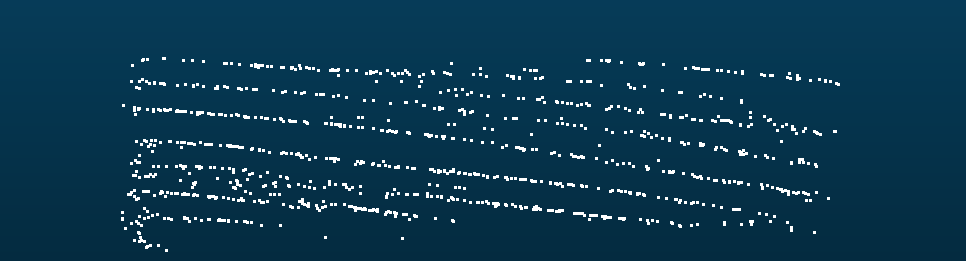} &
\includegraphics[width=0.20\textwidth,  valign=m, keepaspectratio,] {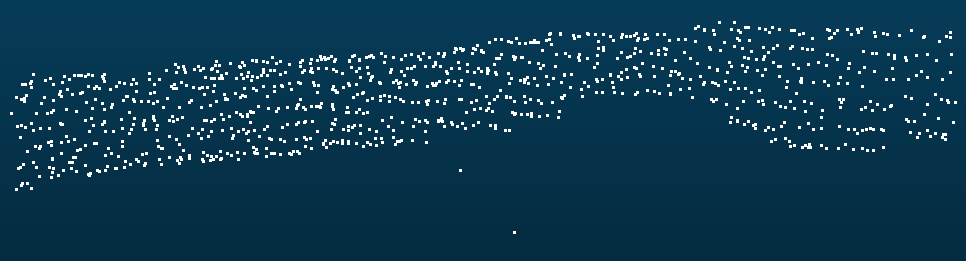} \\

{\rotatebox[origin=t]{90}{bike\vphantom{p}}} 
\includegraphics[width=0.20\textwidth, valign=m, keepaspectratio,] {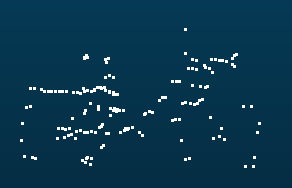} &
\includegraphics[width=0.20\textwidth, valign=m,  keepaspectratio,] {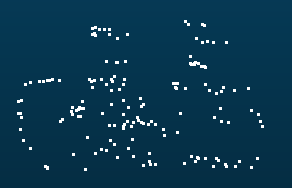} &
\includegraphics[width=0.20\textwidth, valign=m,  keepaspectratio,] {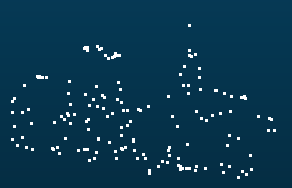} & 
\includegraphics[width=0.20\textwidth, valign=m,  keepaspectratio,] {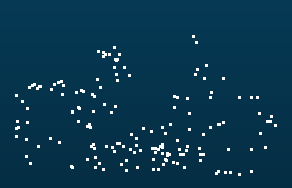} &
\includegraphics[width=0.20\textwidth, valign=m,  keepaspectratio,] {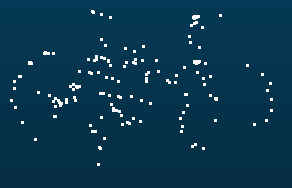} &
\includegraphics[width=0.20\textwidth, valign=m,  keepaspectratio,] {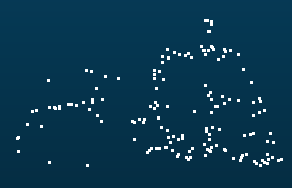} \\

{\rotatebox[origin=t]{90}{bus\vphantom{p}}} 
\includegraphics[width=0.20\textwidth,  valign=m,  keepaspectratio,] {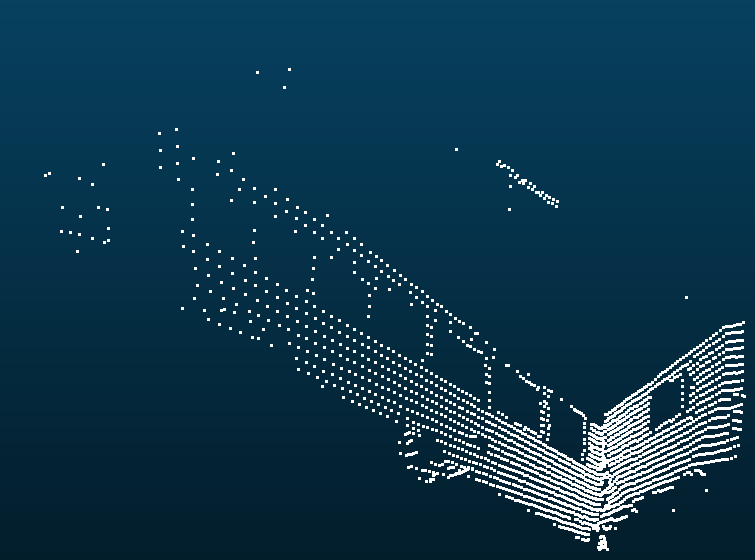} &
\includegraphics[width=0.20\textwidth,  valign=m,   keepaspectratio,] {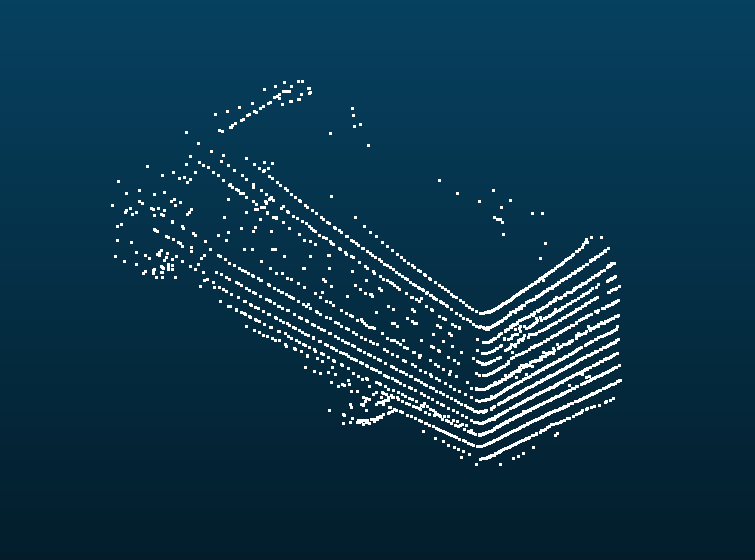} &
\includegraphics[width=0.20\textwidth,  valign=m,  keepaspectratio,] {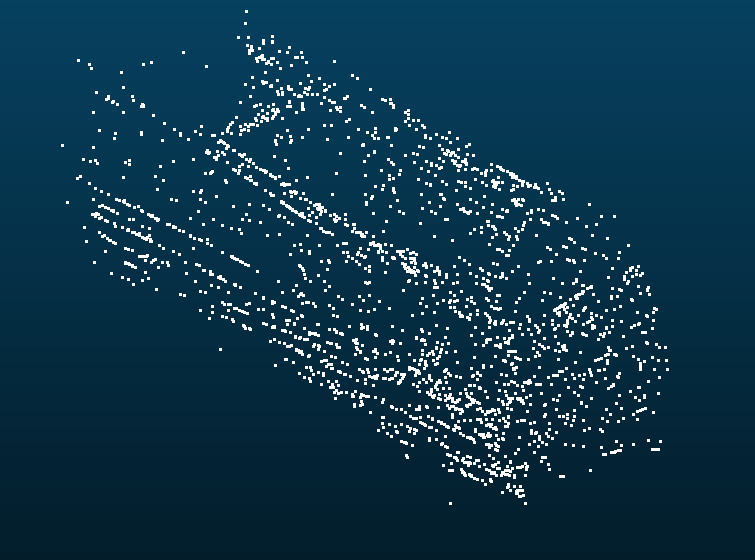} & 
\includegraphics[width=0.20\textwidth,  valign=m,   keepaspectratio,] {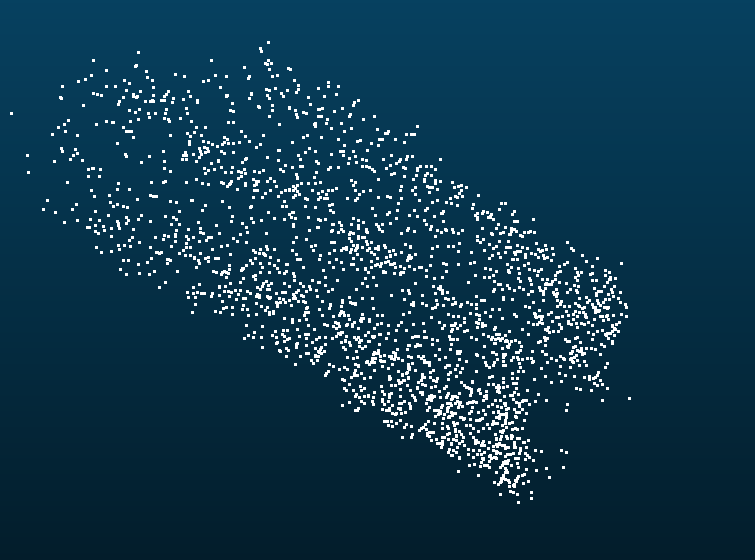} &
\includegraphics[width=0.20\textwidth,  valign=m,  keepaspectratio,] {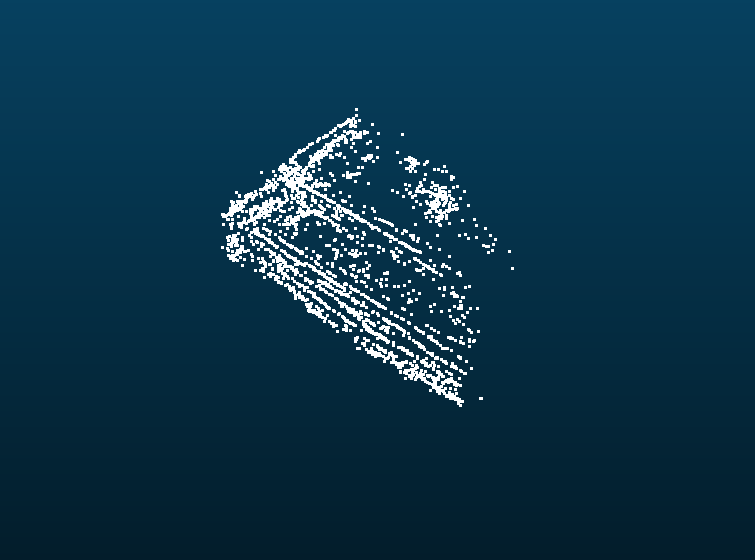} &
\includegraphics[width=0.20\textwidth,  valign=m,   keepaspectratio,] {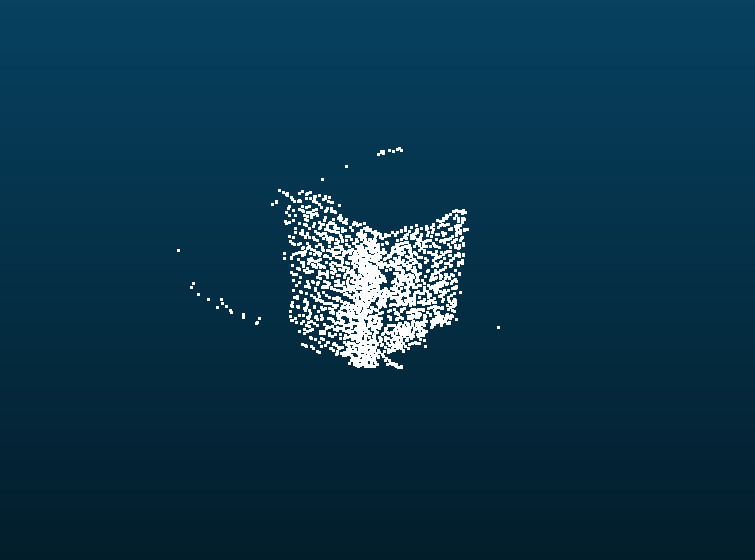} \\

{\rotatebox[origin=t]{90}{car\vphantom{p}}} 
\includegraphics[width=0.20\textwidth,  valign=m,   keepaspectratio,] {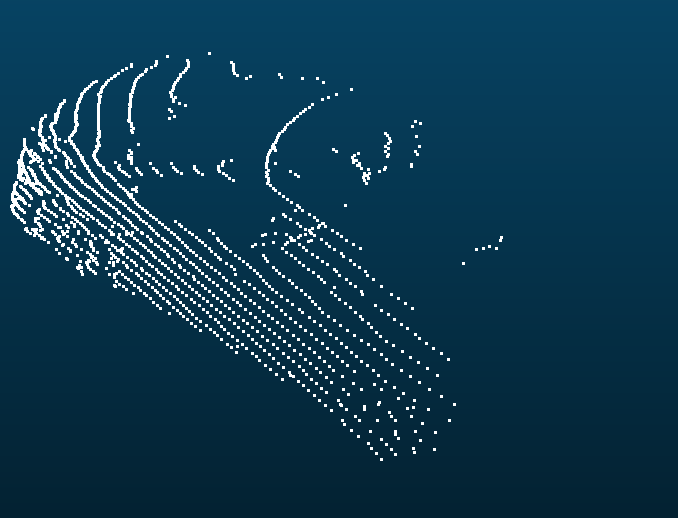} &
\includegraphics[width=0.20\textwidth,  valign=m,   keepaspectratio,] {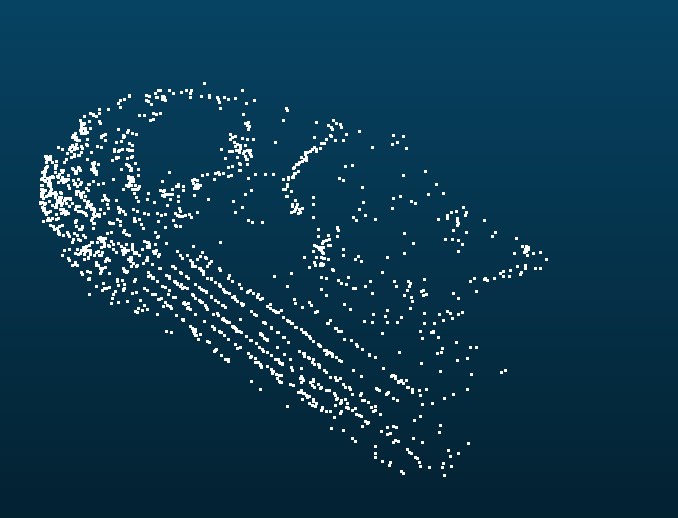} &
\includegraphics[width=0.20\textwidth,   valign=m,  keepaspectratio,] {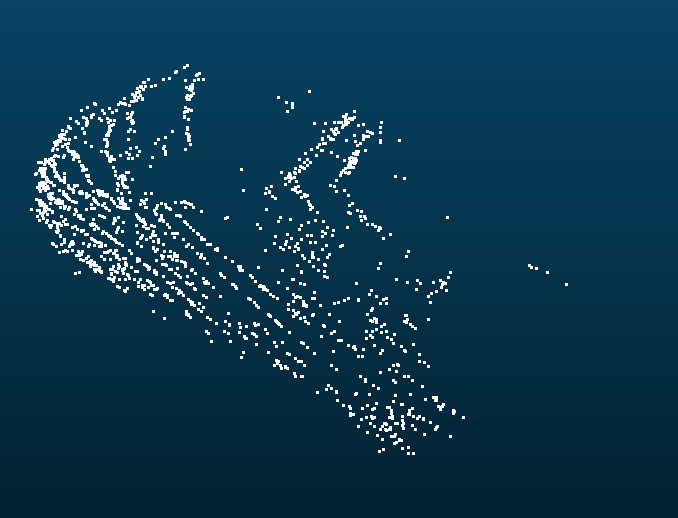} & 
\includegraphics[width=0.20\textwidth,  valign=m,   keepaspectratio,] {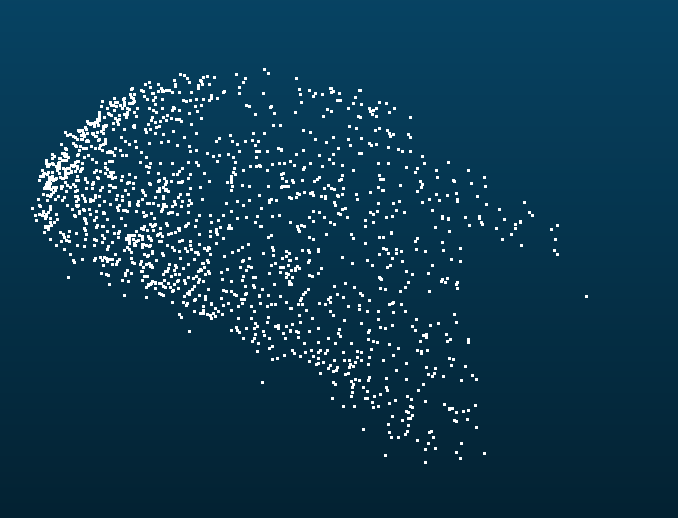} &
\includegraphics[width=0.20\textwidth,  valign=m,   keepaspectratio,] {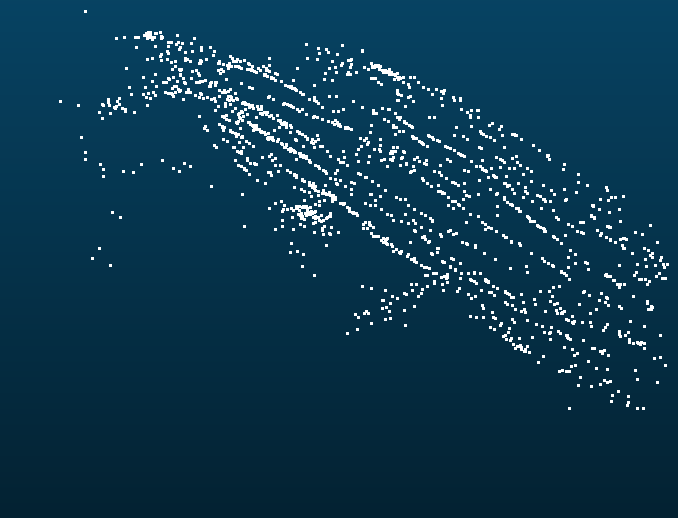} &
\includegraphics[width=0.20\textwidth,   valign=m,  keepaspectratio,] {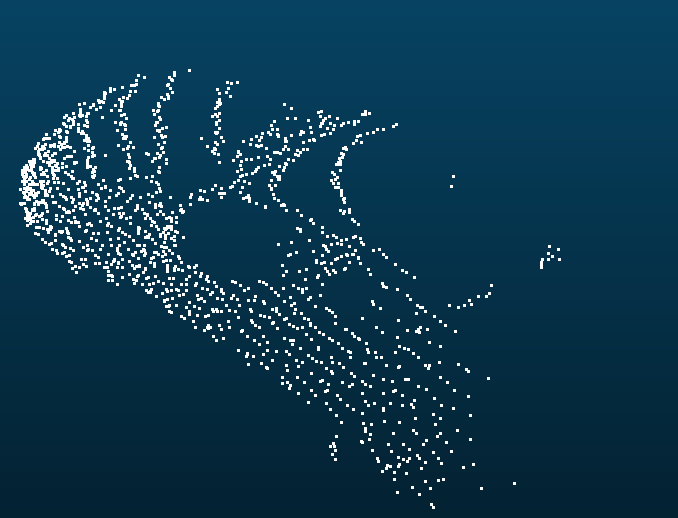} \\

{\rotatebox[origin=t]{90}{constr.\ vehicle\vphantom{p}}} 
\includegraphics[width=0.20\textwidth,  valign=m,   keepaspectratio,] {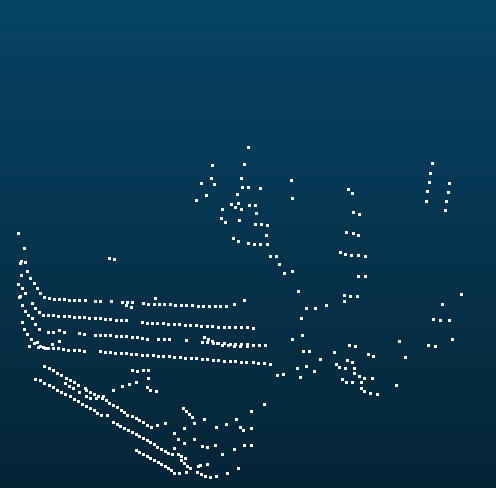} &
\includegraphics[width=0.20\textwidth,  valign=m,   keepaspectratio,] {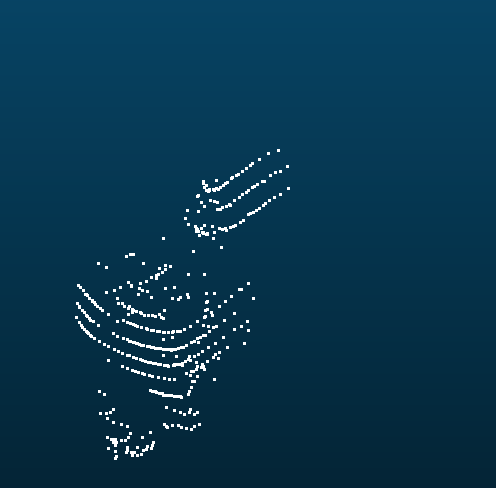} &
\includegraphics[width=0.20\textwidth,  valign=m,   keepaspectratio,] {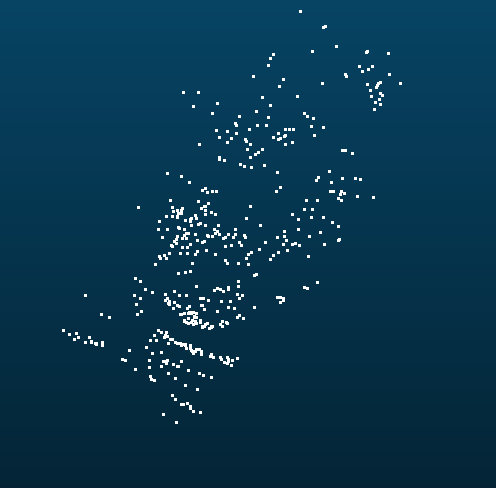} & 
\includegraphics[width=0.20\textwidth,  valign=m,   keepaspectratio,] {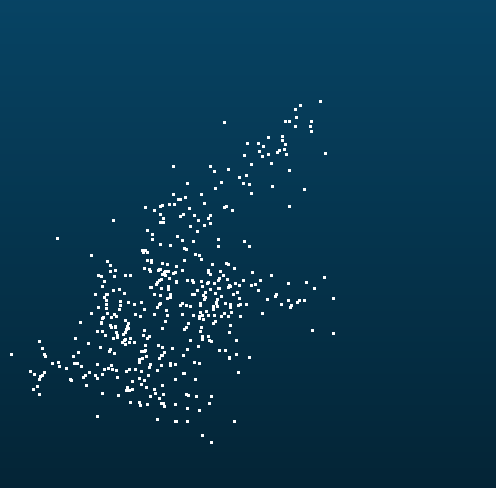} &
\includegraphics[width=0.20\textwidth,  valign=m,   keepaspectratio,] {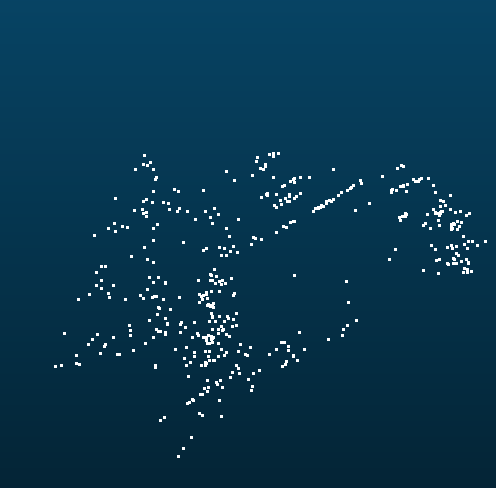} &
\includegraphics[width=0.20\textwidth,  valign=m,   keepaspectratio,] {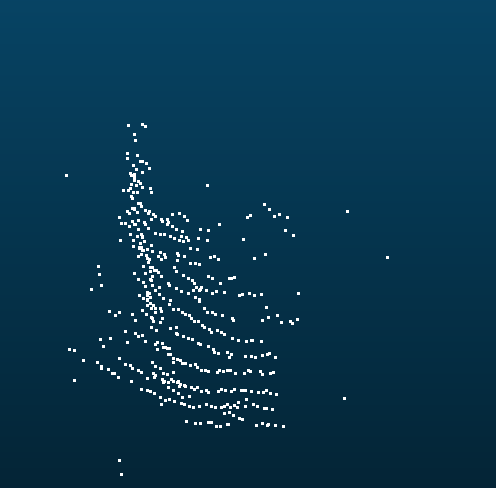} \\

{\rotatebox[origin=t]{90}{motorcycle\vphantom{p}}} 
\includegraphics[width=0.20\textwidth,   valign=m,  keepaspectratio,] {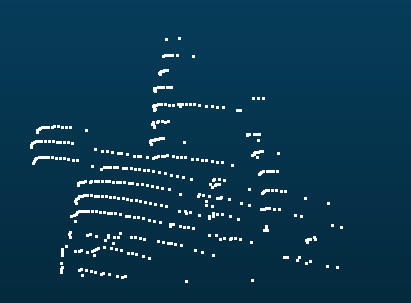} &
\includegraphics[width=0.20\textwidth,  valign=m,   keepaspectratio,] {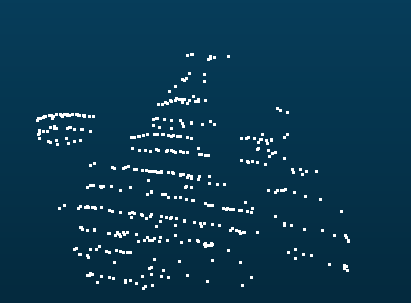} &
\includegraphics[width=0.20\textwidth,  valign=m,   keepaspectratio,] {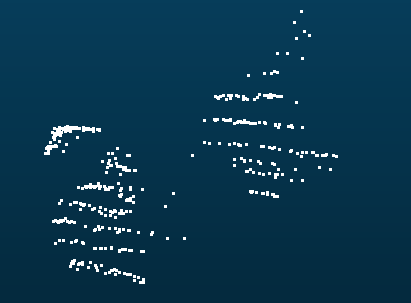} & 
\includegraphics[width=0.20\textwidth,  valign=m,   keepaspectratio,] {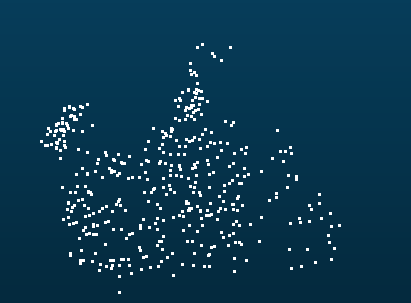} &
\includegraphics[width=0.20\textwidth,  valign=m,   keepaspectratio,] {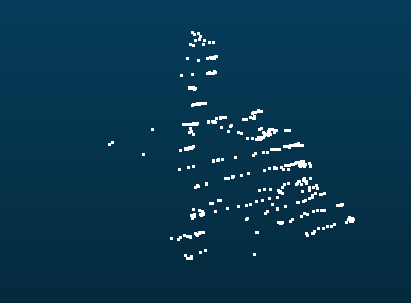} &
\includegraphics[width=0.20\textwidth,  valign=m,   keepaspectratio,] {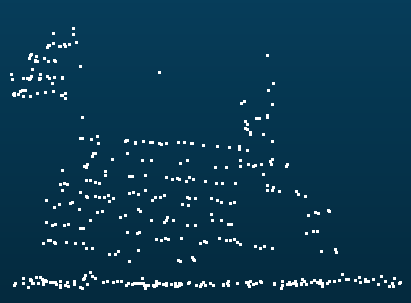} \\

{\rotatebox[origin=t]{90}{pedestrian\vphantom{p}}} 
\includegraphics[width=0.20\textwidth,  valign=m,   keepaspectratio,] {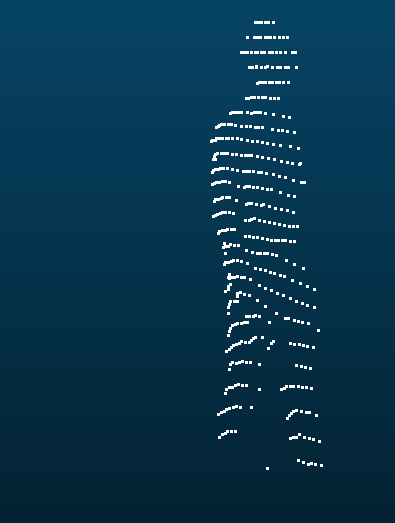} &
\includegraphics[width=0.20\textwidth,   valign=m,  keepaspectratio,] {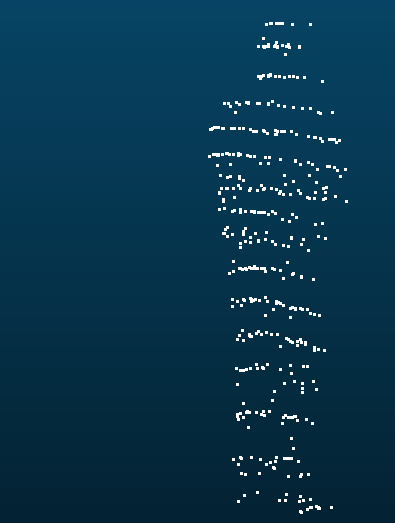} &
\includegraphics[width=0.20\textwidth,  valign=m,   keepaspectratio,] {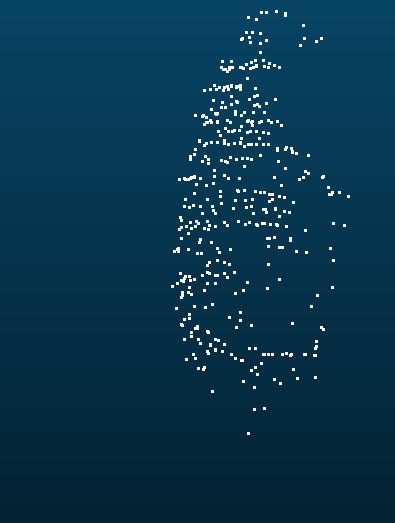} & 
\includegraphics[width=0.20\textwidth,  valign=m,   keepaspectratio,] {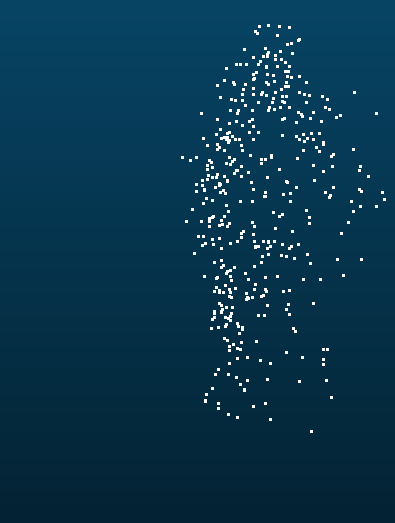} &
\includegraphics[width=0.20\textwidth,  valign=m,   keepaspectratio,] {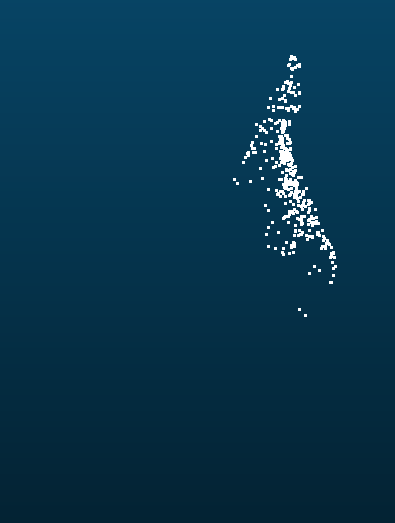} &
\includegraphics[width=0.20\textwidth,  valign=m,   keepaspectratio,] {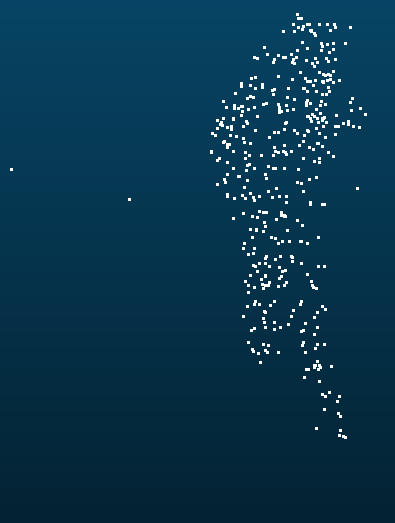} \\

{\rotatebox[origin=t]{90}{\!\!\!traffic cone\vphantom{p}}} 
\includegraphics[width=0.20\textwidth,  valign=m,   keepaspectratio,] {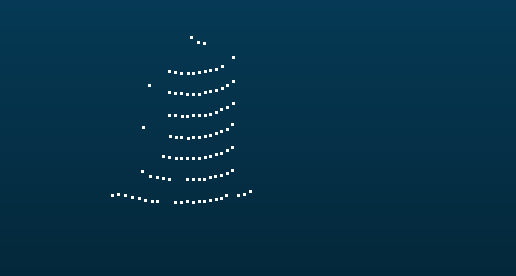} &
\includegraphics[width=0.20\textwidth,  valign=m,   keepaspectratio,] {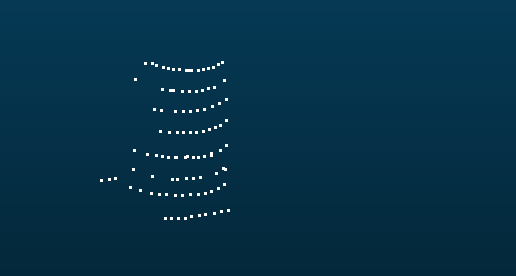} &
\includegraphics[width=0.20\textwidth,  valign=m,   keepaspectratio,] {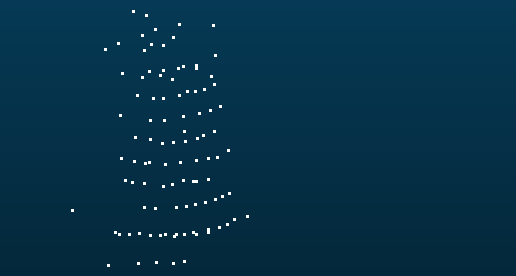} & 
\includegraphics[width=0.20\textwidth, valign=m,    keepaspectratio,] {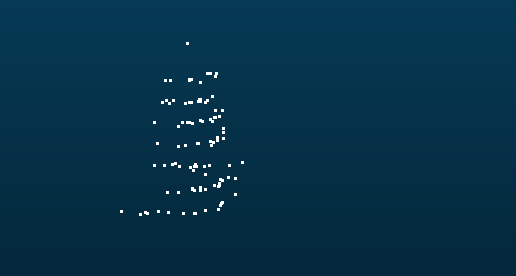} &
\includegraphics[width=0.20\textwidth,  valign=m,   keepaspectratio,] {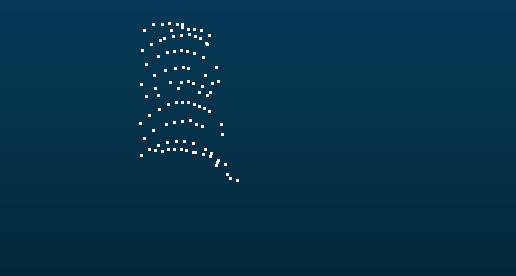} &
\includegraphics[width=0.20\textwidth,  valign=m,   keepaspectratio,] {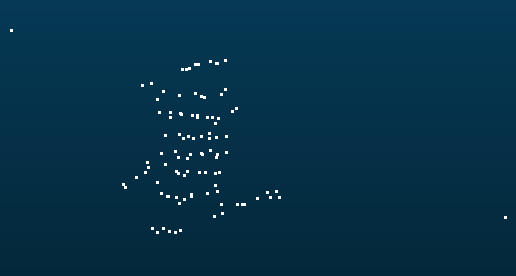} \\

{\rotatebox[origin=t]{90}{trailer\vphantom{p}}} 
\includegraphics[width=0.20\textwidth,  valign=m,   keepaspectratio,] {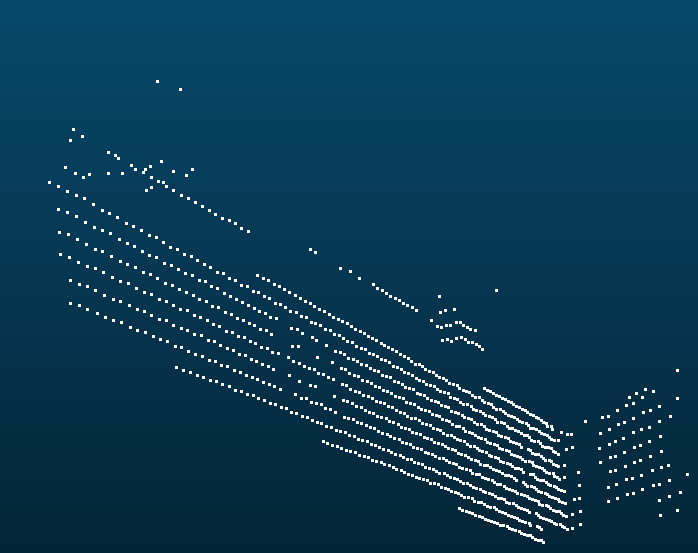} &
\includegraphics[width=0.20\textwidth,   valign=m,  keepaspectratio,] {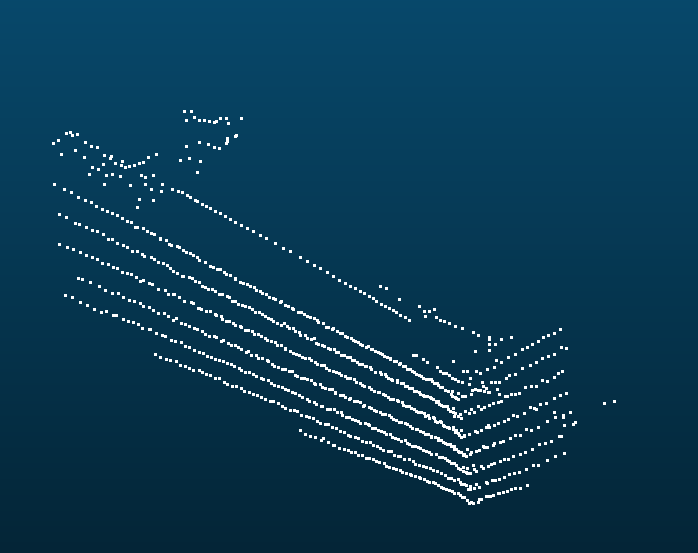} &
\includegraphics[width=0.20\textwidth,   valign=m,  keepaspectratio,] {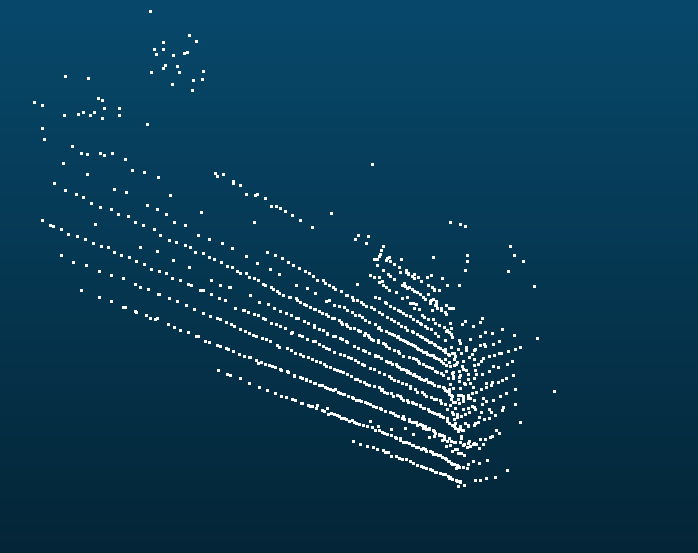} & 
\includegraphics[width=0.20\textwidth,  valign=m,   keepaspectratio,] {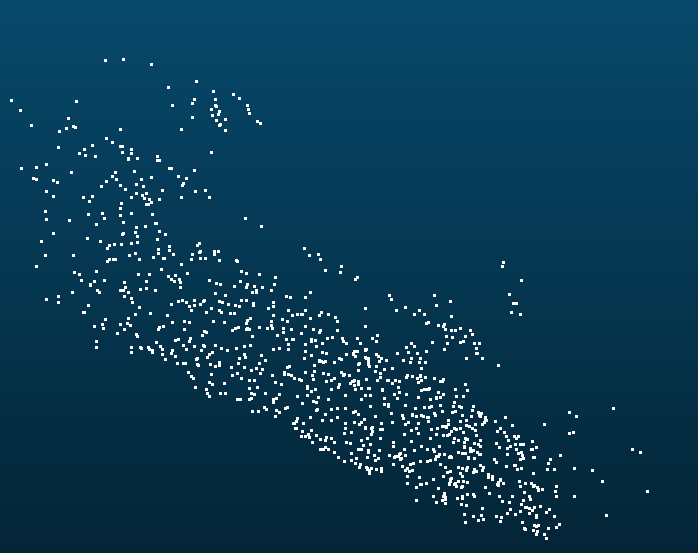} &
\includegraphics[width=0.20\textwidth,  valign=m,   keepaspectratio,] {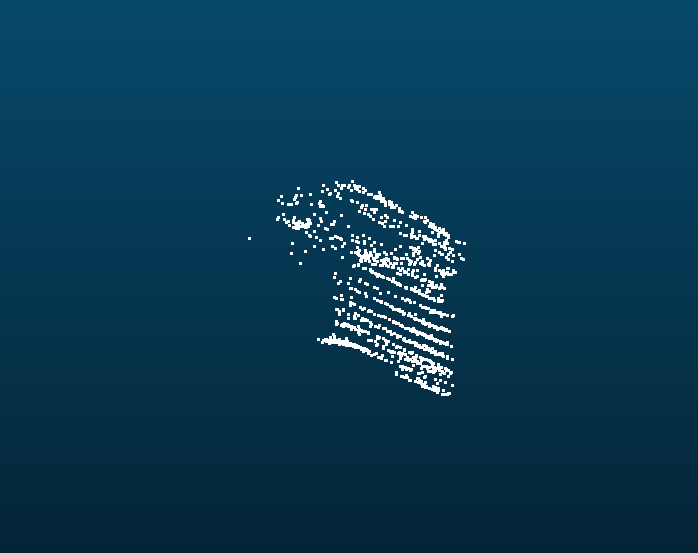} &
\includegraphics[width=0.20\textwidth,   valign=m,  keepaspectratio,] {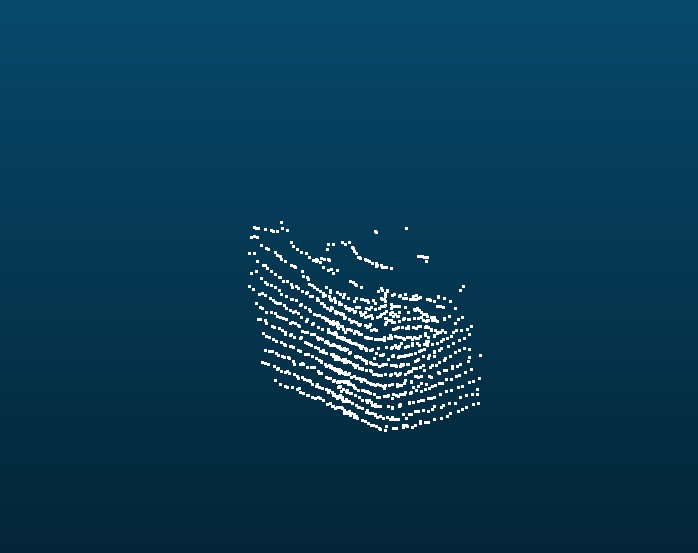} \\

{\rotatebox[origin=t]{90}{truck\vphantom{p}}} 
\includegraphics[width=0.20\textwidth, valign=m,   keepaspectratio,] {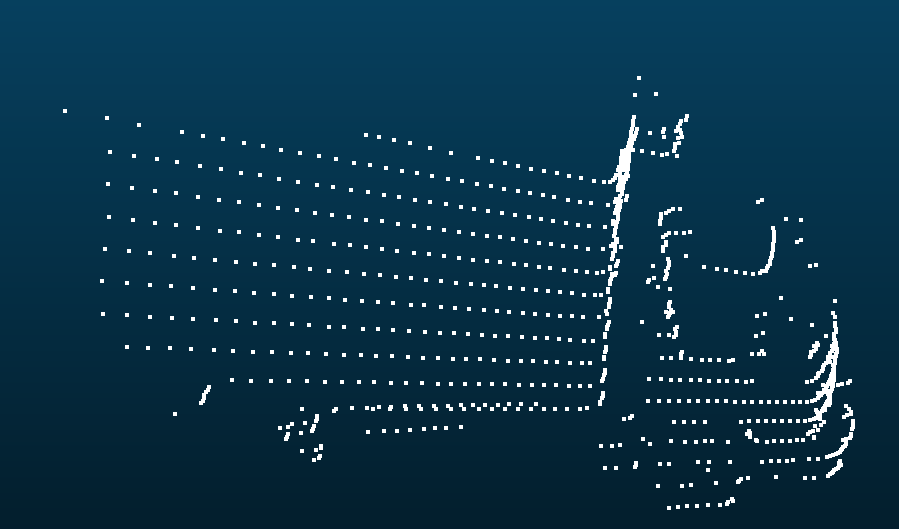} &
\includegraphics[width=0.20\textwidth, valign=m,   keepaspectratio,] {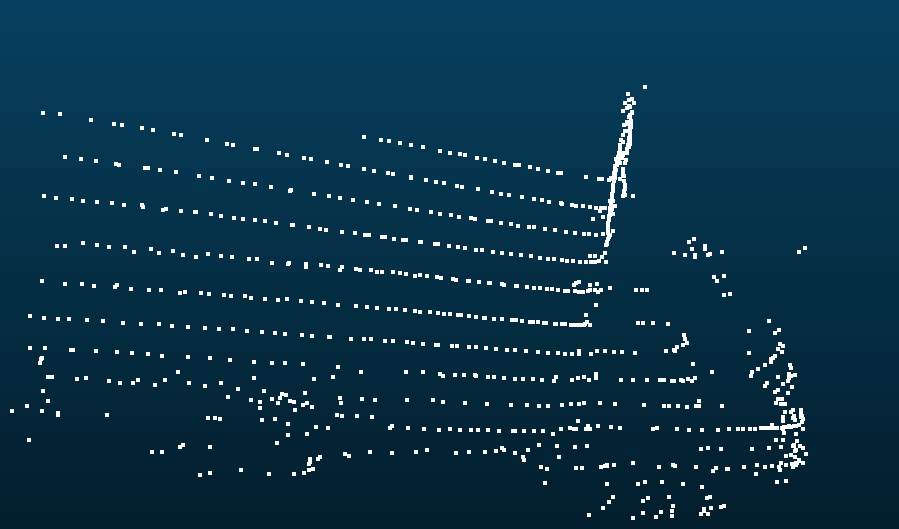} &
\includegraphics[width=0.20\textwidth, valign=m,   keepaspectratio,] {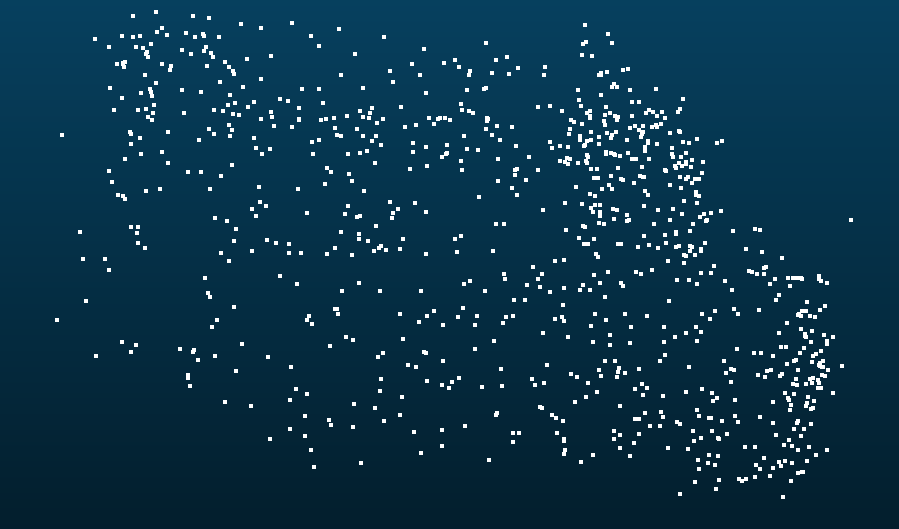} & 
\includegraphics[width=0.20\textwidth,valign=m,    keepaspectratio,] {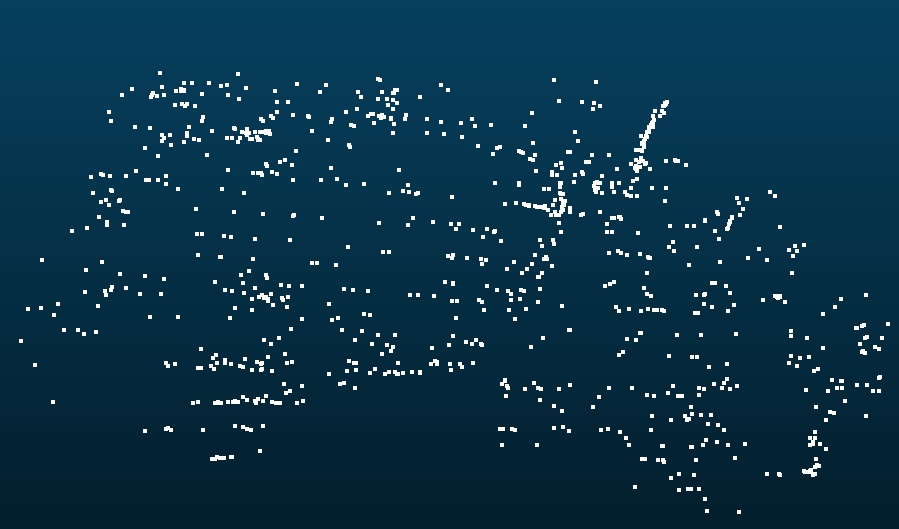} &
\includegraphics[width=0.20\textwidth,  valign=m,  keepaspectratio,] {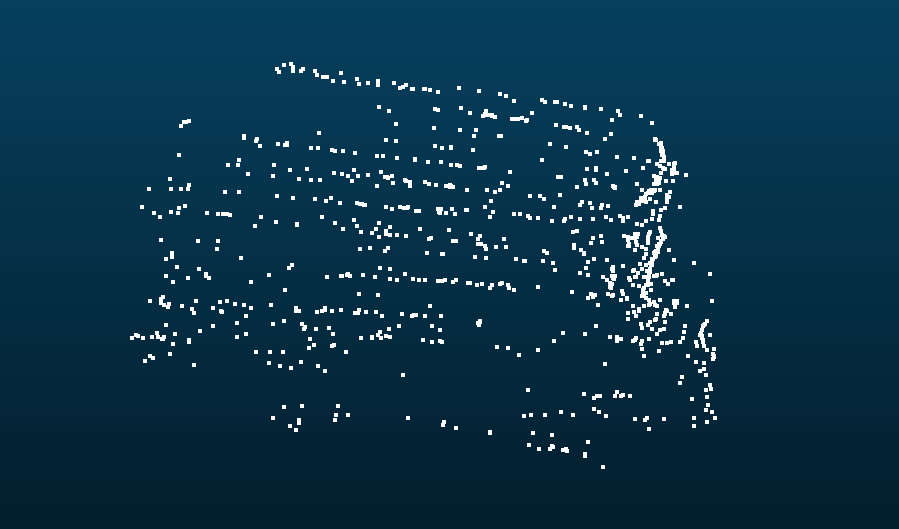} &
\includegraphics[width=0.20\textwidth, valign=m,   keepaspectratio,] {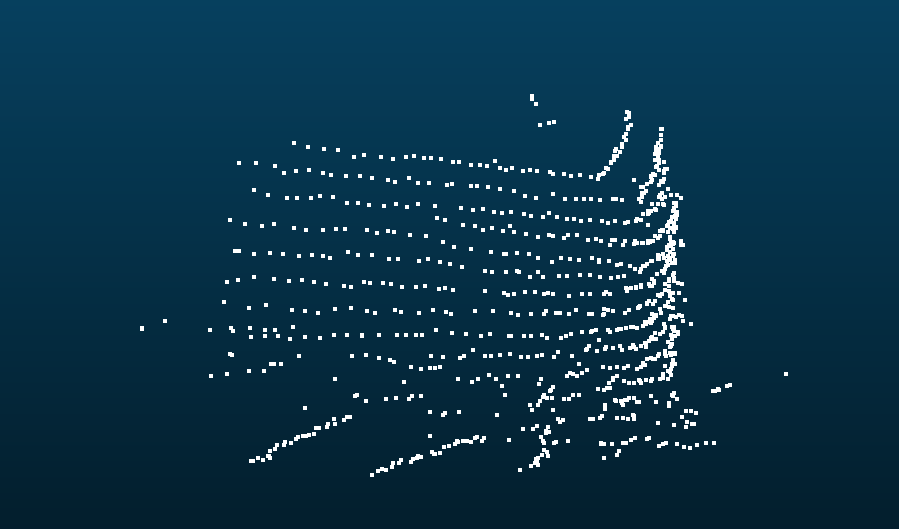} \\

\end{tabular}}
\caption{\textbf{Comparisons of real objects and generated output for all ten classes.} Note that the \ours is able to both capture the LiDAR pattern and generate objects of varying scales and shapes. Other models produce outputs with a degraded LiDAR pattern or do not generate coherent examples of rare classes such as bikes, motorcycles and trucks.}
\label{fig:visual-all-models}
\end{figure*}

\begin{figure*}[t]
\captionsetup[subfigure]{labelformat=empty}
\centering
\setlength\tabcolsep{1.5pt} 
\resizebox{1.0\textwidth}{!}{
\vspace{1cm}
\begin{tabular}{rccccc}

GT~~~~~~~~~~~~~~  & 
recreation (rot.\ 0)&
novel view, rot.\ \nicefrac{1}{5}  &
novel view, rot.\ \nicefrac{2}{5}  &
novel view, rot.\ \nicefrac{3}{5}  &
novel view, rot.\ \nicefrac{4}{5}  \\

{\rotatebox[origin=t]{90}{\!\!\!barrier\vphantom{p}}} 
\includegraphics[width=0.20\textwidth,  valign=m, keepaspectratio,] {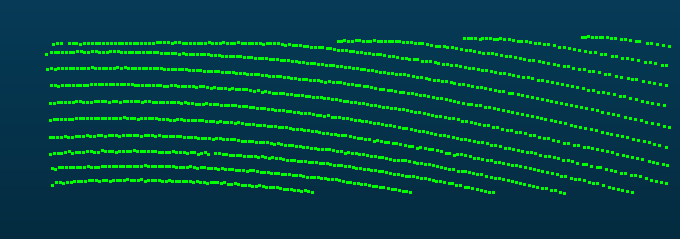} &
\includegraphics[width=0.20\textwidth, valign=m,  keepaspectratio,] {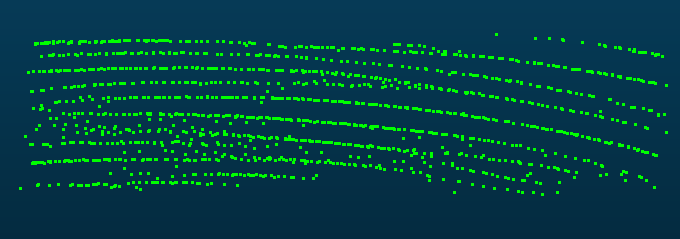} &
\includegraphics[width=0.20\textwidth, valign=m,  keepaspectratio,] {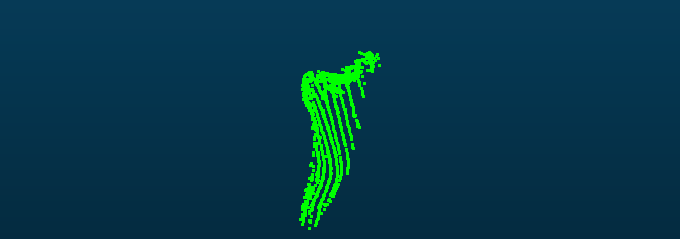} & 
\includegraphics[width=0.20\textwidth,  valign=m, keepaspectratio,] {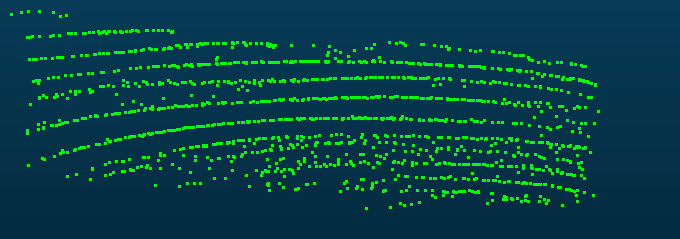} &
\includegraphics[width=0.20\textwidth,  valign=m, keepaspectratio,] {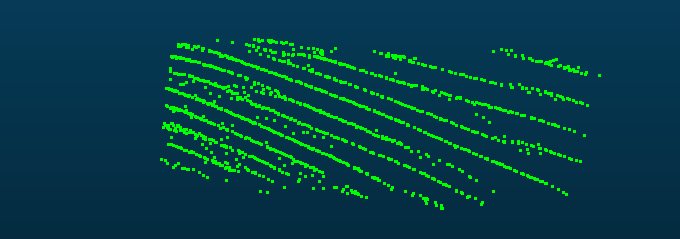} &
\includegraphics[width=0.20\textwidth,  valign=m, keepaspectratio,] {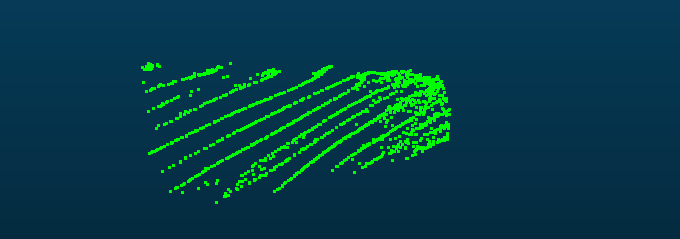} \\

{\rotatebox[origin=t]{90}{bike\vphantom{p}}} 
\includegraphics[width=0.20\textwidth, valign=m, keepaspectratio,] {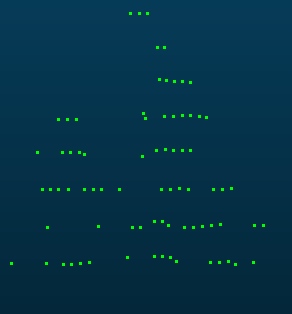} &
\includegraphics[width=0.20\textwidth, valign=m,  keepaspectratio,] {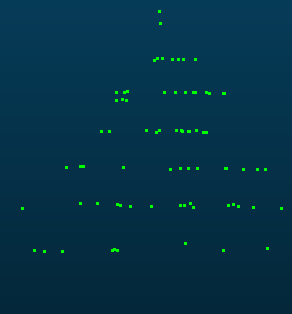} &
\includegraphics[width=0.20\textwidth, valign=m,  keepaspectratio,] {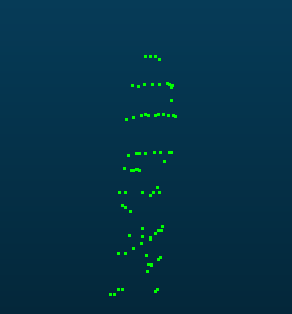} & 
\includegraphics[width=0.20\textwidth, valign=m,  keepaspectratio,] {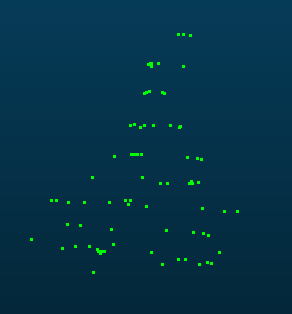} &
\includegraphics[width=0.20\textwidth, valign=m,  keepaspectratio,] {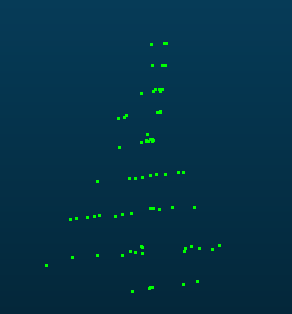} &
\includegraphics[width=0.20\textwidth, valign=m,  keepaspectratio,] {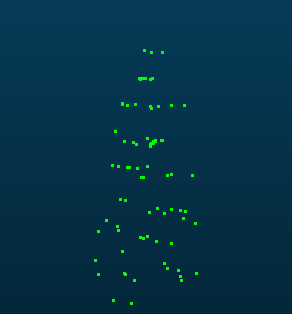} \\

{\rotatebox[origin=t]{90}{bus\vphantom{p}}} 
\includegraphics[width=0.20\textwidth,  valign=m,  keepaspectratio,] {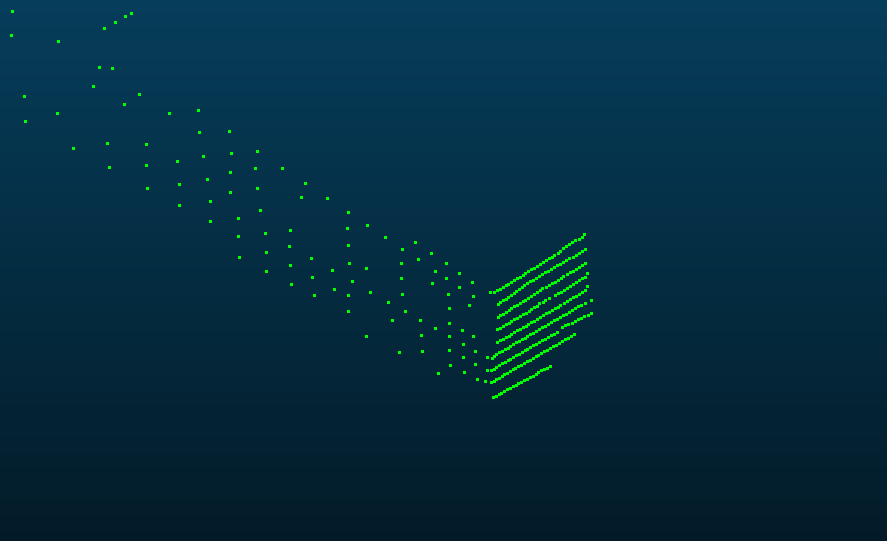} &
\includegraphics[width=0.20\textwidth,  valign=m,   keepaspectratio,] {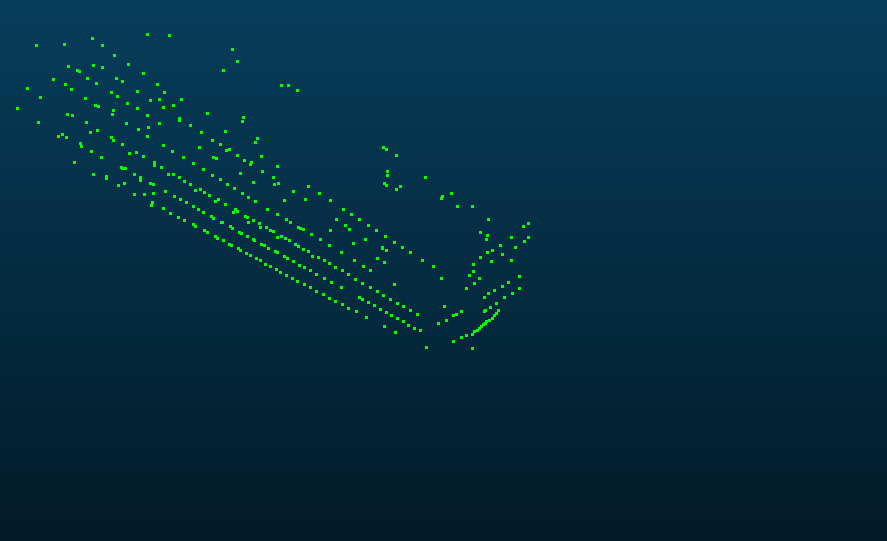} &
\includegraphics[width=0.20\textwidth,  valign=m,  keepaspectratio,] {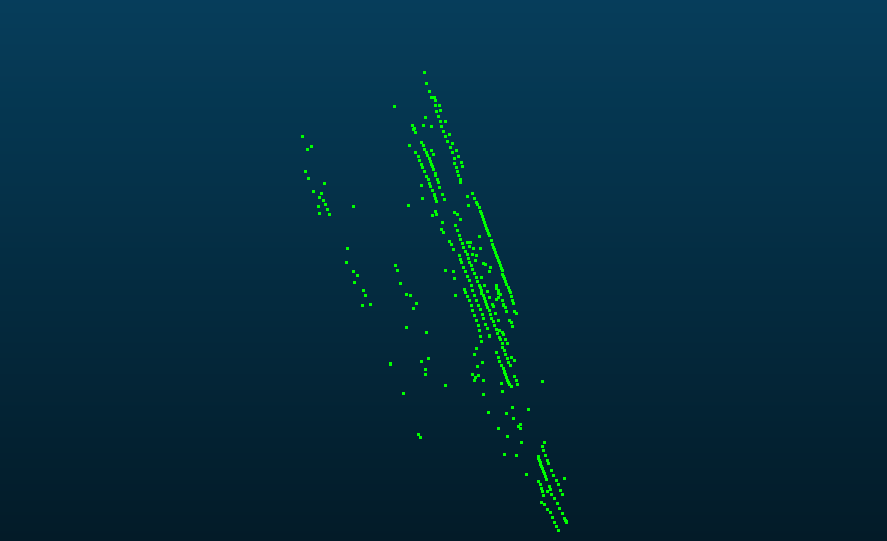} & 
\includegraphics[width=0.20\textwidth,  valign=m,   keepaspectratio,] {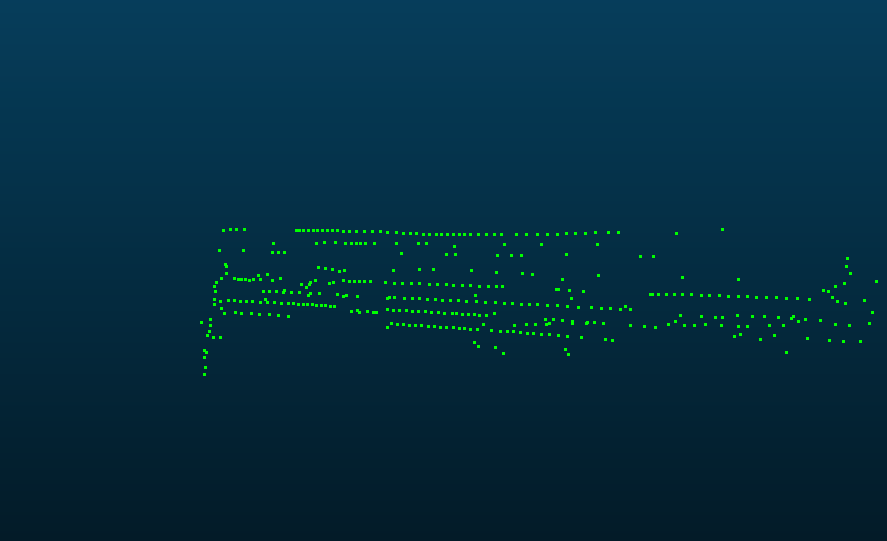} &
\includegraphics[width=0.20\textwidth,  valign=m,  keepaspectratio,] {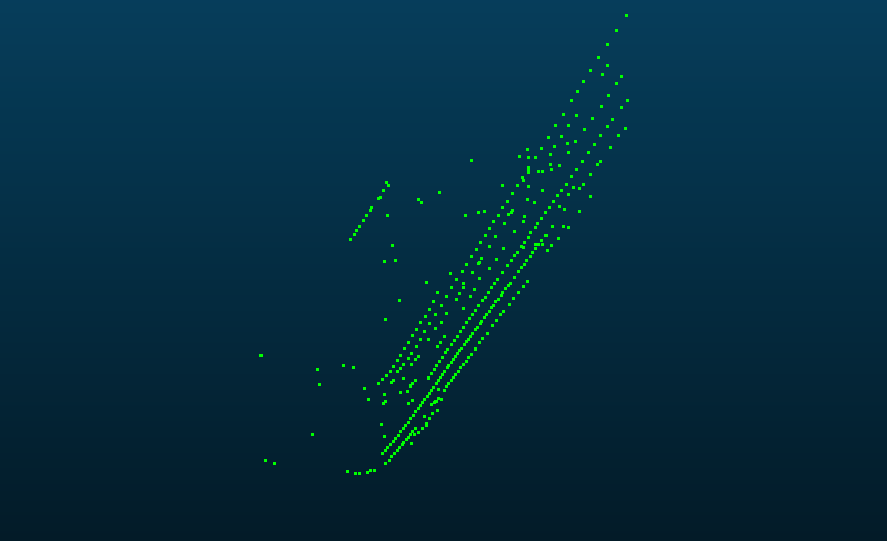} &
\includegraphics[width=0.20\textwidth,  valign=m,   keepaspectratio,] {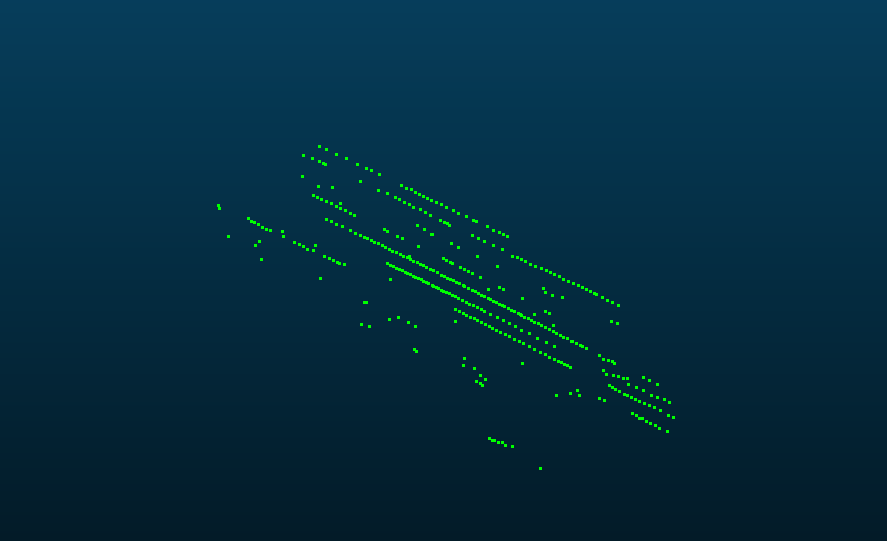} \\

{\rotatebox[origin=t]{90}{car\vphantom{p}}} 
\includegraphics[width=0.20\textwidth,  valign=m,   keepaspectratio,] {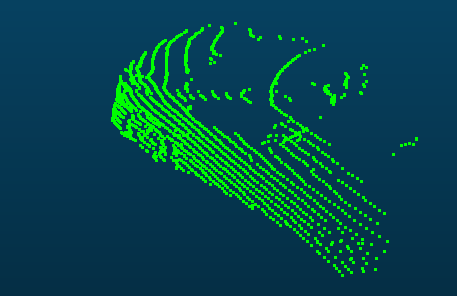} &
\includegraphics[width=0.20\textwidth,  valign=m,   keepaspectratio,] {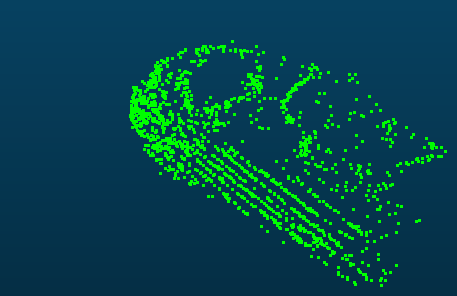} &
\includegraphics[width=0.20\textwidth,   valign=m,  keepaspectratio,] {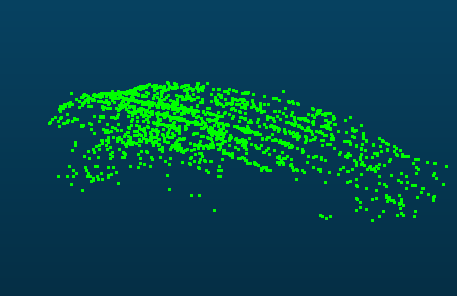} & 
\includegraphics[width=0.20\textwidth,  valign=m,   keepaspectratio,] {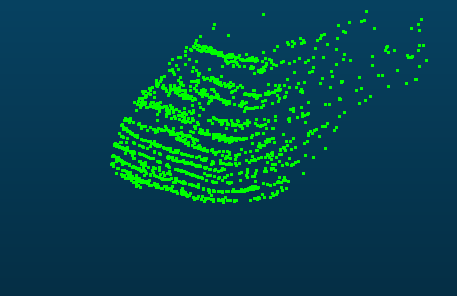} &
\includegraphics[width=0.20\textwidth,  valign=m,   keepaspectratio,] {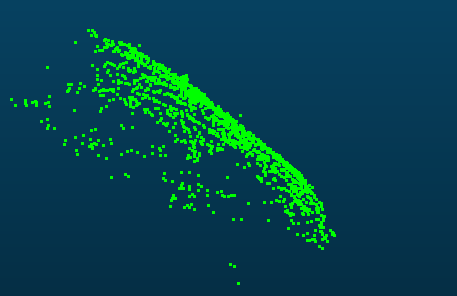} &
\includegraphics[width=0.20\textwidth,   valign=m,  keepaspectratio,] {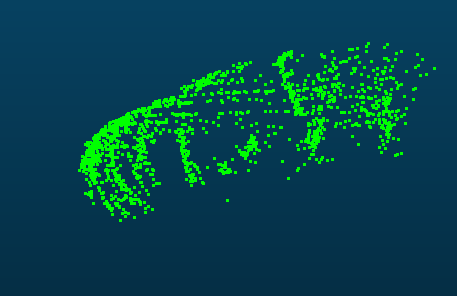} \\

{\rotatebox[origin=t]{90}{\clap{constr.\ veh.}\vphantom{p}}} 
\includegraphics[width=0.20\textwidth,  valign=m,   keepaspectratio,] {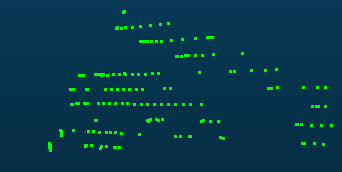} &
\includegraphics[width=0.20\textwidth,  valign=m,   keepaspectratio,] {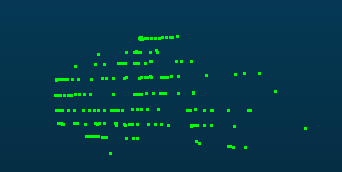} &
\includegraphics[width=0.20\textwidth,  valign=m,   keepaspectratio,] {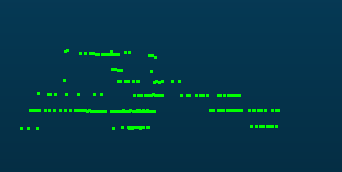} & 
\includegraphics[width=0.20\textwidth,  valign=m,   keepaspectratio,] {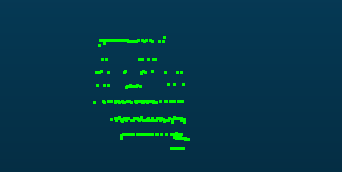} &
\includegraphics[width=0.20\textwidth,  valign=m,   keepaspectratio,] {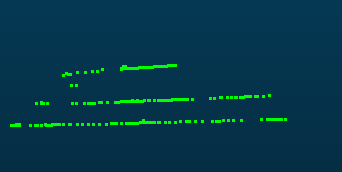} &
\includegraphics[width=0.20\textwidth,  valign=m,   keepaspectratio,] {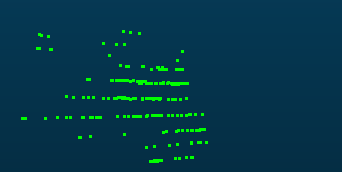} \\

{\rotatebox[origin=t]{90}{motorcycle\vphantom{p}}} 
\includegraphics[width=0.20\textwidth,   valign=m,  keepaspectratio,] {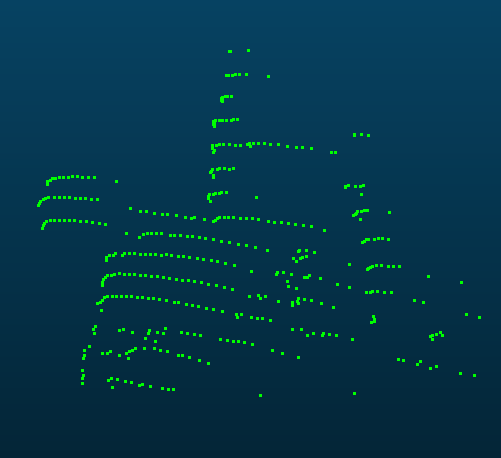} &
\includegraphics[width=0.20\textwidth,  valign=m,   keepaspectratio,] {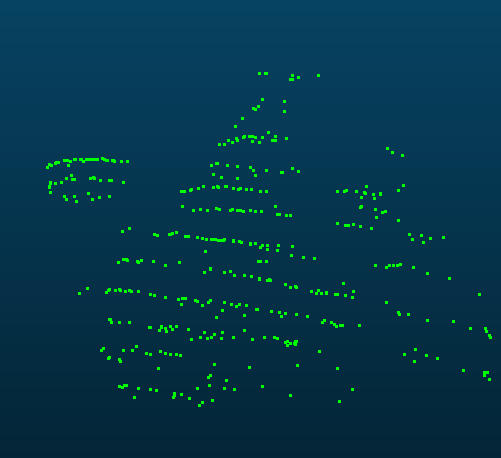} &
\includegraphics[width=0.20\textwidth,  valign=m,   keepaspectratio,] {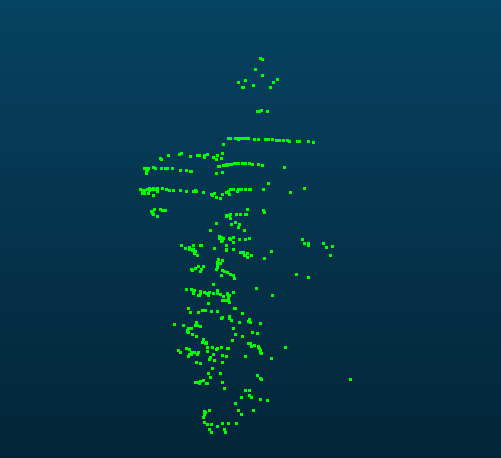} & 
\includegraphics[width=0.20\textwidth,  valign=m,   keepaspectratio,] {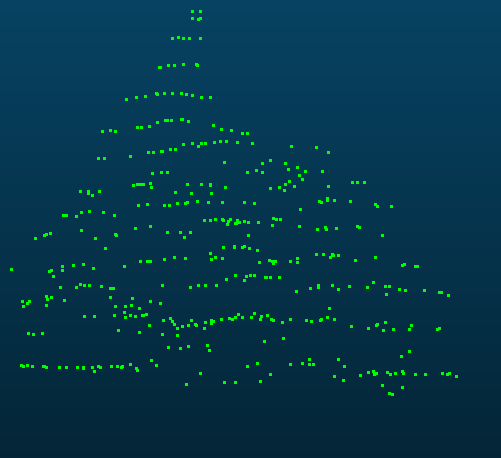} &
\includegraphics[width=0.20\textwidth,  valign=m,   keepaspectratio,] {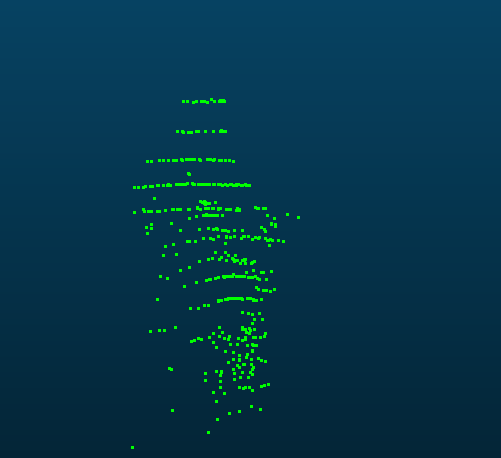} &
\includegraphics[width=0.20\textwidth,  valign=m,   keepaspectratio,] {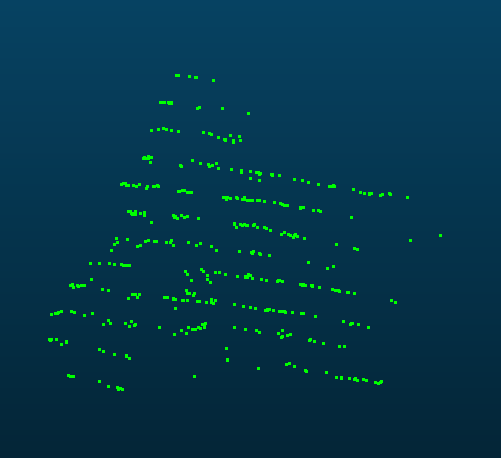} \\

{\rotatebox[origin=t]{90}{pedestrian\vphantom{p}}} 
\includegraphics[width=0.20\textwidth, height=0.3\textwidth, valign=m] {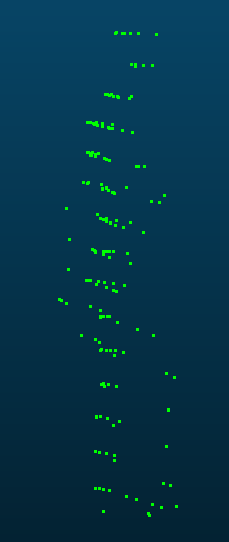} &
\includegraphics[width=0.20\textwidth, height=0.3\textwidth, valign=m] {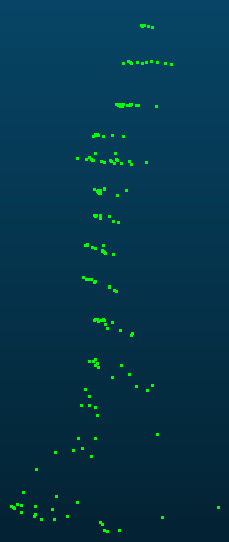} &
\includegraphics[width=0.20\textwidth, height=0.3\textwidth, valign=m] {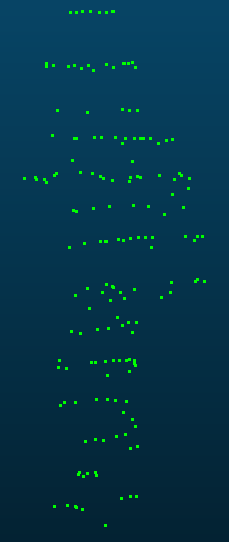} & 
\includegraphics[width=0.20\textwidth, height=0.3\textwidth, valign=m] {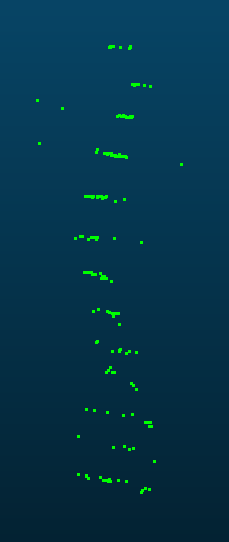} &
\includegraphics[width=0.20\textwidth, height=0.3\textwidth, valign=m] {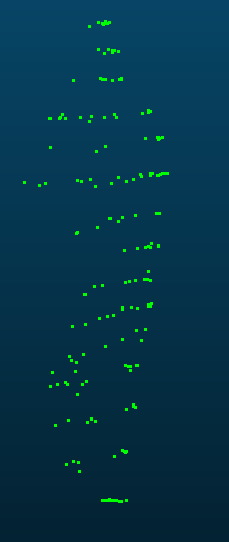} &
\includegraphics[width=0.20\textwidth, height=0.3\textwidth, valign=m] {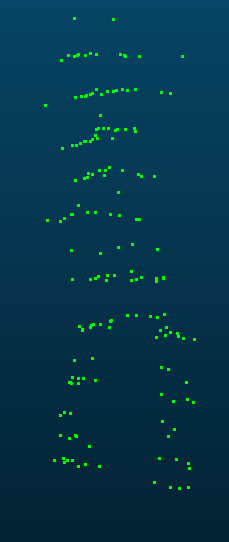} \\

{\rotatebox[origin=t]{90}{\!\!\!traffic cone\vphantom{p}}} 
\includegraphics[width=0.20\textwidth, height=0.2\textwidth, valign=m] {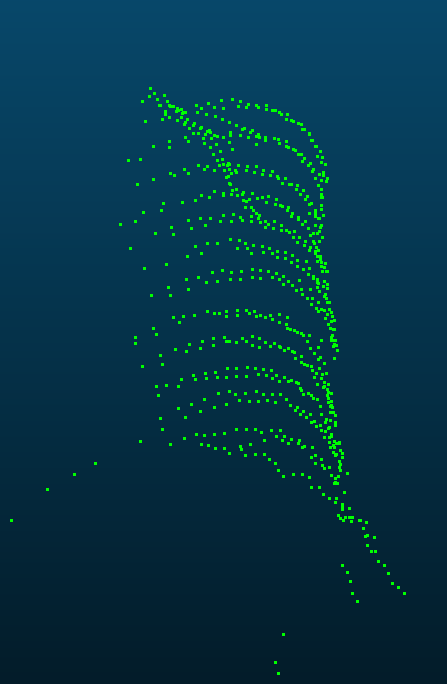} &
\includegraphics[width=0.20\textwidth, height=0.2\textwidth, valign=m] {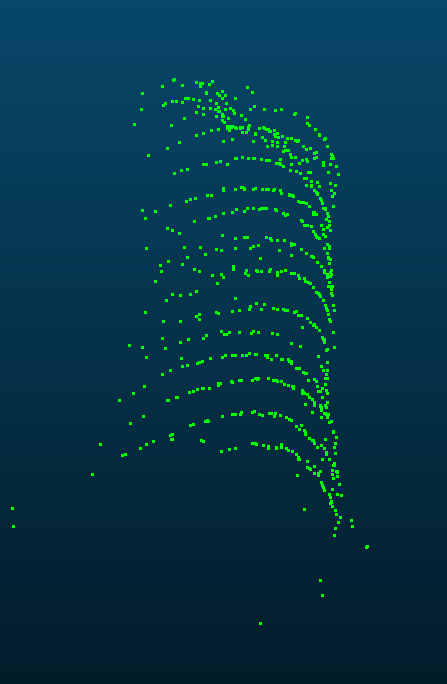} &
\includegraphics[width=0.20\textwidth, height=0.2\textwidth, valign=m] {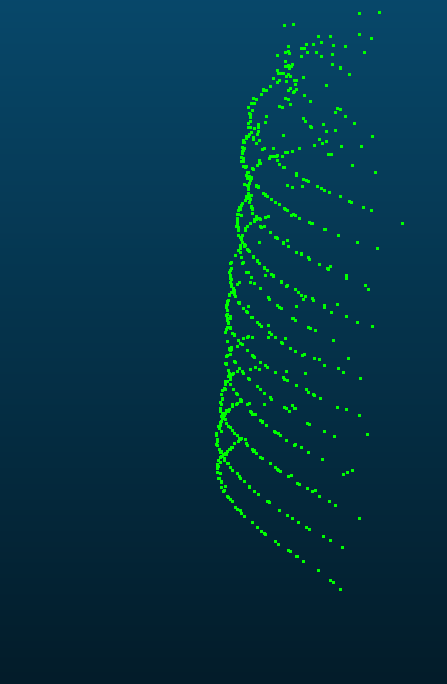} & 
\includegraphics[width=0.20\textwidth, height=0.2\textwidth, valign=m] {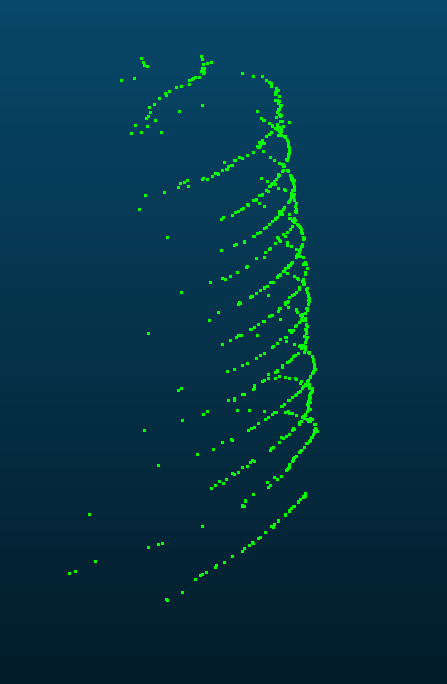} &
\includegraphics[width=0.20\textwidth, height=0.2\textwidth, valign=m] {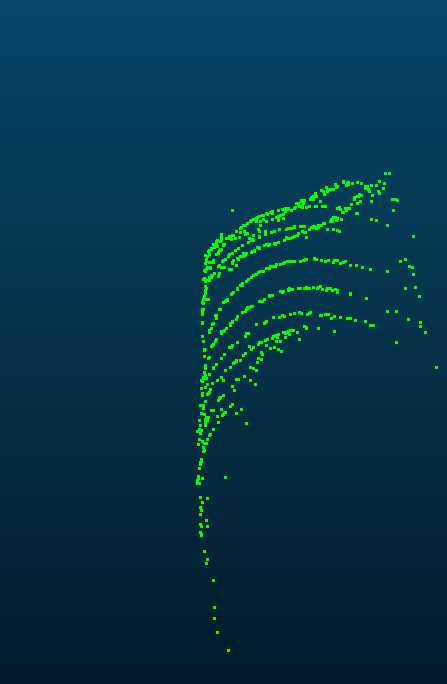} &
\includegraphics[width=0.20\textwidth, height=0.2\textwidth, valign=m] {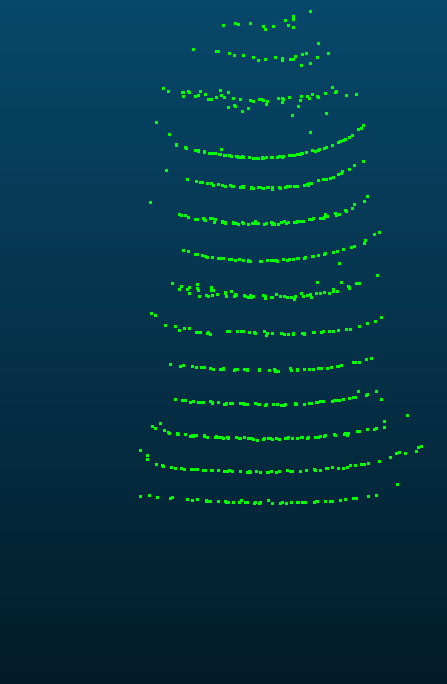} \\

{\rotatebox[origin=t]{90}{trailer\vphantom{p}}} 
\includegraphics[width=0.20\textwidth,  valign=m,   keepaspectratio,] {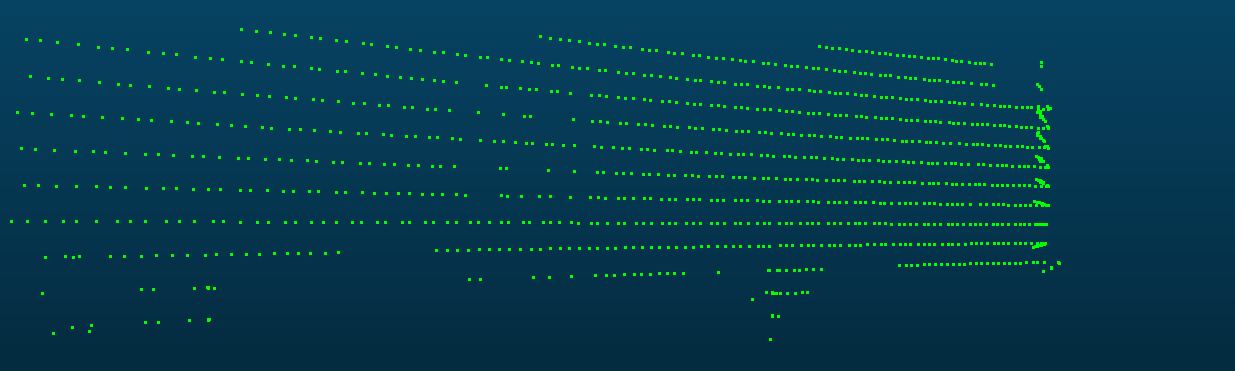} &
\includegraphics[width=0.20\textwidth,   valign=m,  keepaspectratio,] {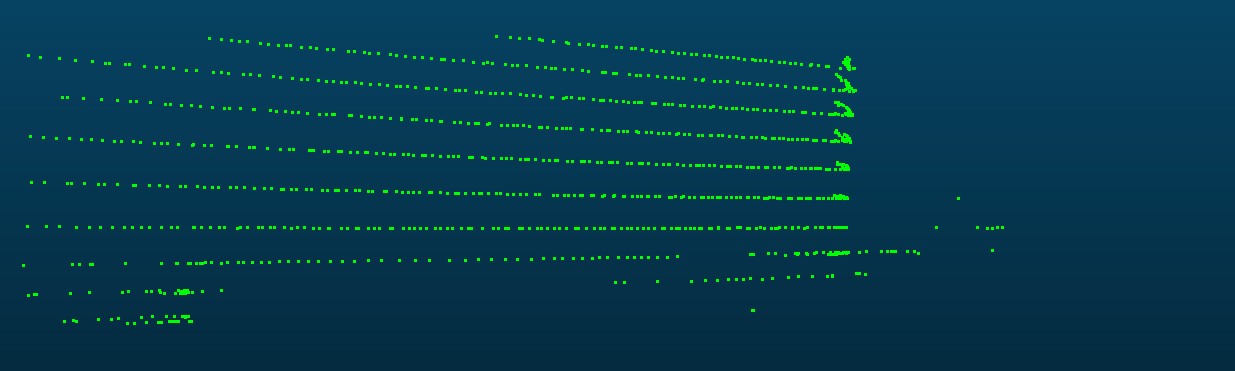} &
\includegraphics[width=0.20\textwidth,   valign=m,  keepaspectratio,] {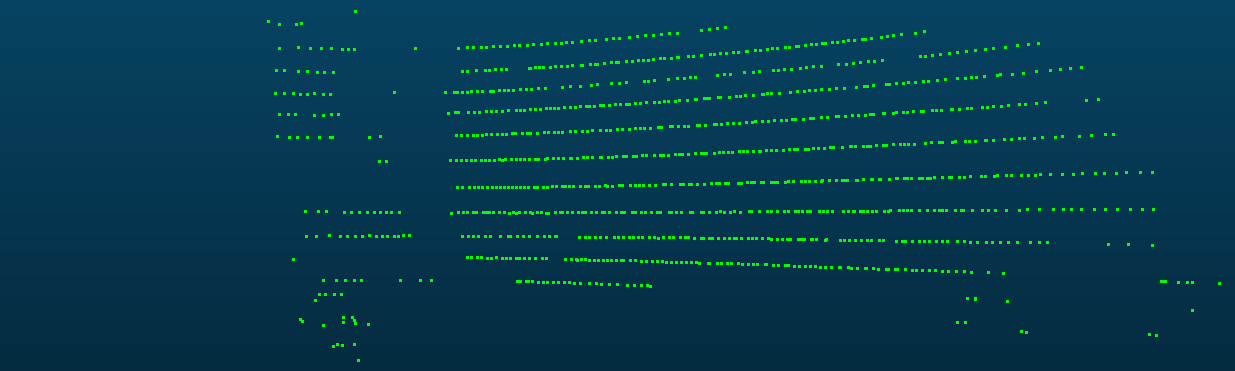} & 
\includegraphics[width=0.20\textwidth,  valign=m,   keepaspectratio,] {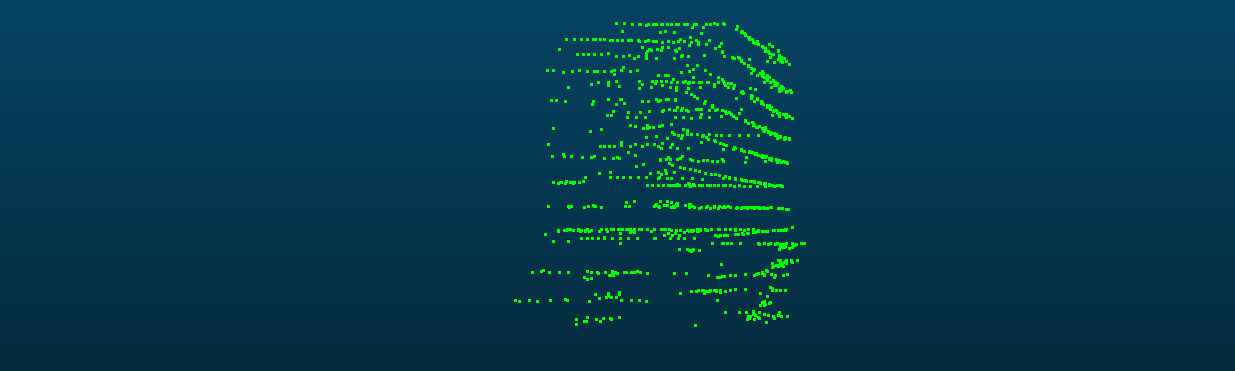} &
\includegraphics[width=0.20\textwidth,  valign=m,   keepaspectratio,] {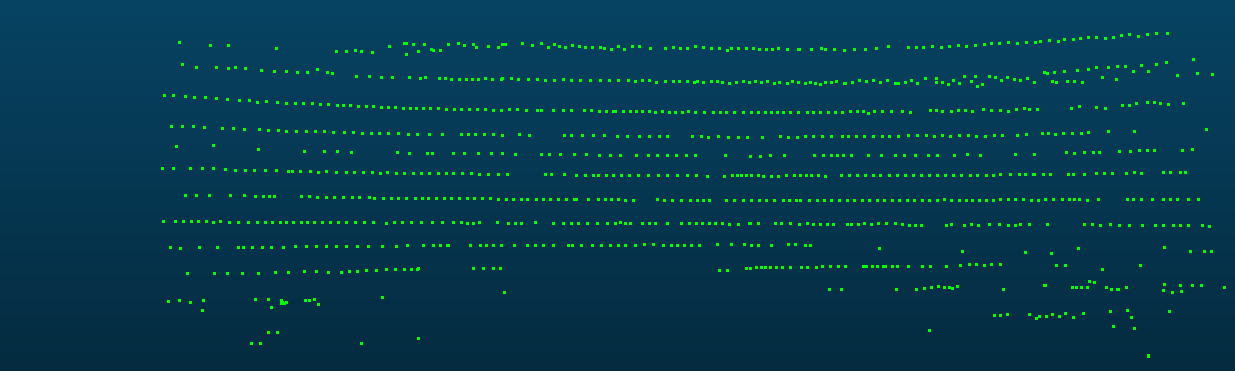} &
\includegraphics[width=0.20\textwidth,   valign=m,  keepaspectratio,] {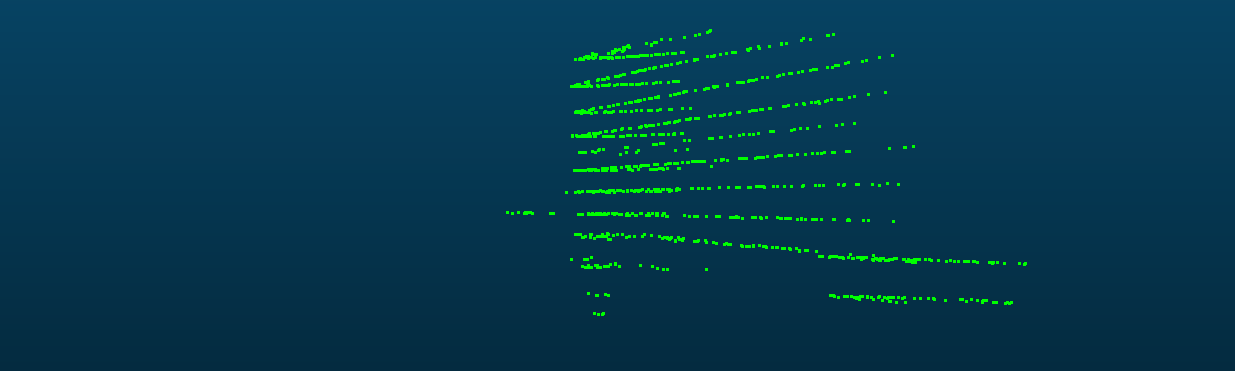} \\

{\rotatebox[origin=t]{90}{truck\vphantom{p}}} 
\includegraphics[width=0.20\textwidth, valign=m,   keepaspectratio,] {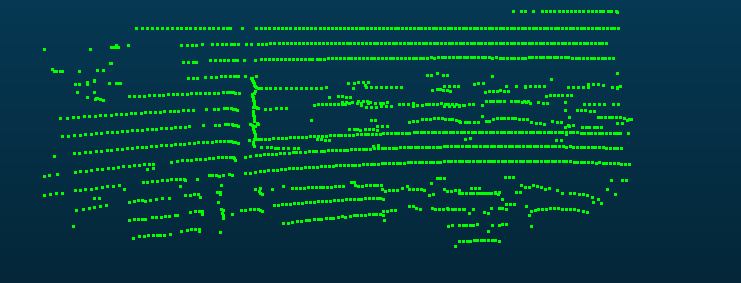} &
\includegraphics[width=0.20\textwidth, valign=m,   keepaspectratio,] {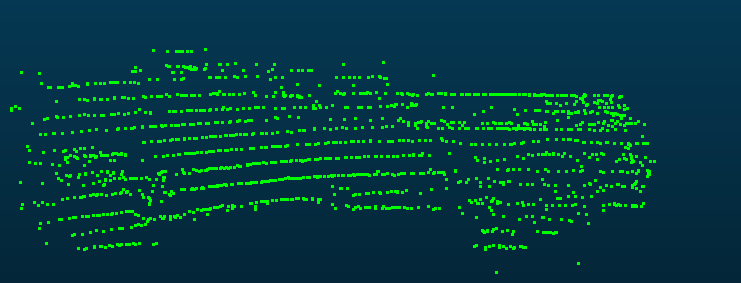} &
\includegraphics[width=0.20\textwidth, valign=m,   keepaspectratio,] {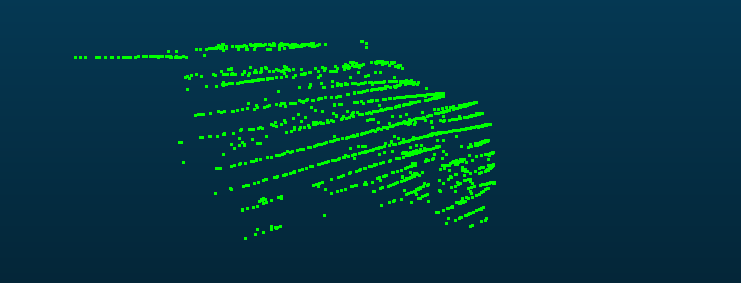} & 
\includegraphics[width=0.20\textwidth,valign=m,    keepaspectratio,] {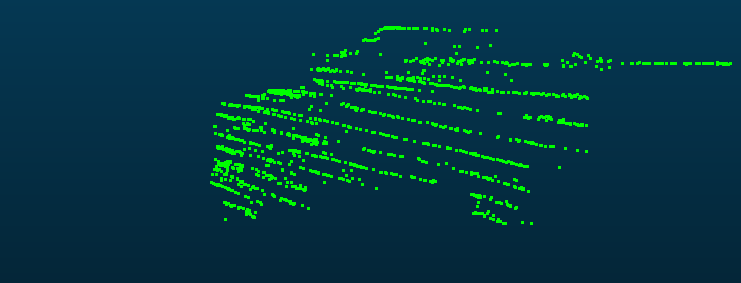} &
\includegraphics[width=0.20\textwidth,  valign=m,  keepaspectratio,] {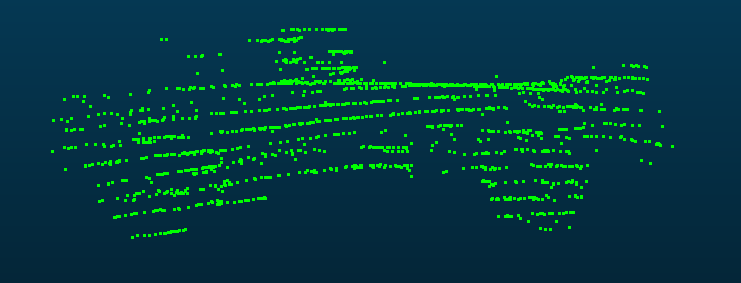} &
\includegraphics[width=0.20\textwidth, valign=m,   keepaspectratio,] {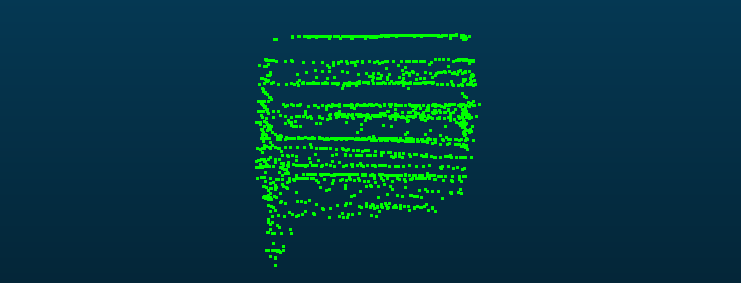} \\

\end{tabular}}
\caption{\textbf{Novel objects produced by \ours.} Recreations are  generated using the conditioning information $\kappa$ of a real object, the rest of the objects are created from novel views by interpolating the viewing angle $\phi$ of the condition.
}
\label{fig:visual-augs}
\vspace{1cm}
\end{figure*}

\begin{table*}[t]
    \centering
    \vspace{8pt}
    \resizebox{0.8\linewidth}{!}{
    \begin{tabular}{l|c c c c c c c c c c c}
        \toprule
         \textbf{Mean} & barrier & bicycle & bus & car & constr.\,veh. & motorcycle & pedestrian & traf.\,cone & trailer & truck \\
        \midrule
          $+\textbf{0.3}$ & $-0.5$ & $-0.8$ & $-1.2$ & $+0.7$ & $-1.7$ & $+0.5$ & 0 & $+4.3$ & $+0.3$ & $+1.1$ \\
        \bottomrule
    \end{tabular}
    }
    \captionof{table}{\textbf{Impact of inplace data augmentation with LOGen samples for semantic segmentation on nuScenes validation set.} During training, each sample is selected with a 50\% probability to be either a real sample or a generated one.  We display the class-wise increase in IoU\%.}
    \label{tab:mix_training}
    \vspace{-5pt}
\end{table*}

\subsection{Downstream evaluation}
\label{sec:downstream}

\paragraph{Object-level augmentation for semantic segmentation on nuScenes.} To further motivate the focus on object level generation, we design a simple experiment highlighting one use case of the generated objects, enhancing diversity of objects in the training dataset. We train the SPVCNN~\cite{tang2020spvcnn} model for semantic segmentation on the nuScenes dataset. During training, with a 50\% probability, we replace real objects with versions generated by \ours. 
Results are in Tab.\,\ref{tab:mix_training}. We highlight the improvements 
for specific rare classes such as traffic cones, trailers, and trucks, and further improvements for cars. Note that, in this downstream task, using a single-class model is consistent with the selection of specific rare classes to augment.

\paragraph{Cross-dataset object augmentation.} 
We use LOGen trained on nuScenes to enhance the semantic segmentation performance on Semantic\-KITTI for classes \textit{pedestrian} and \textit{motorcycle}, which are commonly targeted for augmentation due to the relatively low performance of their segmentation.  
We randomly insert 10 such generated objects for each class into SemanticKITTI scenes during SPVCNN training. This simple augmentation strategy improves classwise IoU\% from 64.8 to 67.3 for \textit{motorcycle} and from 65.0 to 66.9 for \textit{pedestrian}.

We provide a comprehensive supplementary material that includes detailed information on the diffusion background, implementation specifics, official metric definitions, dataset statistics, evaluation of conditioning parameters, all classwise performance results, comparisons between \ours's single-model-per-class and multi-class approaches, qualitative comparisons between \ours and baselines, examples of novel \ours objects and videos showcasing our object generation results.

\subsection{A discussion on scalability}  
Following other works in the object generation literature \cite{mo2023dit}, we mainly chose to train and evaluate a single model per class. This allowed us to use a compact network, and avoid inconsistent batch sizes. Due to the use of zero-padding, mixing objects with high variation in size results in certain objects being padded with many zero elements. We note that training single model per class introduces no drawbacks in terms of runtime or model efficiency when evaluated in iterations per instance. 

We train and evaluate a multi-class model of the same size trained with the same iterations as shown in Table~\ref{tab:multiclass}, where the multi-class model does not perform as well as the single class versions under most perceptual metrics. However the multi-class model does have the best overall 1-NNA and Coverage, indicating avenues for future work.

Further, while the point embeddings outperform the voxelization approaches in terms of generated quality, the lack of any point compression creates memory consumption issues for larger objects.

\section{Conclusion}
\label{sec:conclusion}
This work demonstrates that a well-designed diffusion model is capable of generating realistic LiDAR objects. Novel metrics, baselines and \ours (best performing architecture) establish the first standard in LiDAR object generation. 
Interesting perspectives include further development of downstream applications.
Future work can increase generation speed and memory efficiency of the models, and develop novel methods to use generated objects.

\section*{Acknowledgments}
This work was granted access to the HPC resources of IDRIS under the allocations AD011013503R1 and AD011013503R3 made by GENCI.

\clearpage
\appendix
\label{sec:appendix}
 \section*{Appendix}
\setcounter{figure}{4}
\setcounter{table}{5}

\noindent This appendix is organized as follows.
\begin{itemize}
    \item \cref{sec:backgrounddiffusion} provides some background on diffusion and in particular on transformer-based diffusion. 
    \item \cref{sec:implemdetails} provides more implementation details.
    \item \cref{sec:metrics} offers definitions and formulas for the metrics we used.
    \item \cref{sec:datadistrib} provides various statistics regarding the distribution of objects we evaluate on.
    \item \cref{sec:conditioning} evaluates the impact of interpolating the conditioning information to produce novel objects.
    \item \cref{sec:moreresults} gives class-wise results. 
\end{itemize}
We also provide videos corresponding to generated examples from each class on this webpage \url{https://nerminsamet.github.io/logen}. 

\section{Background on diffusion}
\label{sec:backgrounddiffusion}

Diffusion has become a common approach for generation at large, and this paper relies on standard diffusion techniques to do so, namely DDPM \cite{ho2020DDPM}, Classifier Free Guidance \cite{ho2022classifier} and DiT \cite{peebles2023scalable}. The novelty that we introduce is at the level of LiDAR object parameterization, including point embeddings (Sec.\,3.1), and architecture of the diffusion transformer (Sec.\,3.3). Nevertheless, to make the paper (with its appendix) more self-contained, we include here a some background on these techniques. We also provide some details on \pixartal and \dittd, from which we draw inspiration and base some of our baselines.

\subsection{Diffusion process}
\label{sec:diffusionprocess}

Generative modeling via the diffusion objective, introduced in \cite{sohl2015deep, song2020score}, has been demonstrated to learn high-quality approximations of data-generating distributions, sufficient to claim the state of the art in a variety of generative tasks. 
The diffusion process was originally formulated as a continuous process modeled on stochastic differential equations.
Interest and development in the topic have led to a confluence of different interpretations of the process, including a time discrete process known as Denoising Diffusion Probabilistic Models, or DDPM \cite{ho2020DDPM}.
Our work focuses on the time-discrete formulation. However, as noted in \cite{karras2022elucidating}, this choice of formulation is largely orthogonal to other design choices, and the sampler and specific diffusion process propose different possible directions to take the model design. We chose to rely on DDPM as it was previously used in LiDAR scan generation literature \cite{nunes2024lidiff}.

Generally, the diffusion process entails the corruption of data by iteratively adding Gaussian noise, arriving finally at data distributed according to a standard Gaussian.
A model then learns to reverse this destructive process via iterative denoising steps.
This allows the generation of novel samples from the data distribution by sampling noise and applying the denoising process. 

\subsection{Time-discrete vs time-continuous diffusion}

The time-discrete vs time-continuous options imply the background in which the diffusion process is formulated.
Under the continuous-time formulation, the forward process is typically a stochastic differential equation that transforms data into noise. 
The reverse process can then be computed via Langevin Monte Carlo Markov chain sampling in order to arrive at the denoised sample \cite{song2020score}. 
Under the discrete-time formulation, the forward process is assumed to be a discrete Markov chain. Under this assumption the process can be modeled via a hierarchical variational auto-encoder with specific restrictions on the data. 
The data dimension and latent dimension are exactly equal, the latent structure is pre-defined as a Gaussian distribution, and the parameters of the Gaussian are controlled such that at the final step $T$ of the Markov chain, the latent is equivalent to the standard normal distribution $\mathcal{N}(\mathbf{0}, \mathbf{I})$ \cite{luo2022understanding}.

The above assumptions assist in the simplification of the optimization process. 
Similar to other VAEs, the approximation of the generative distribution $\log(p(x))$ is optimized via the Evidence Lower Bound (ELBO) \cite{luo2022understanding}. 
Under the assumptions on the shape and underlying distribution of the discrete Markov chain above, optimizing the ELBO simplifies to minimizing the Kullback Leibler divergence (KL) between the forward destructive process and the backwards generative process. And given that both distributions are assumed to be Gaussian, optimizing the KL becomes a simple mean squared error loss.

\subsection{Denoising neural network}
\label{sec:denoisingnetwork}

We define the forward destructive process with a distribution $q(x_t|x_{t-1})$ at step $t$ for a dataset sample~$x$. The associated Markov chain $\{x_0 \,...\, x_t \,...\, x_T\}$ is defined such that $q(x_t|x_{t-1})=\mathcal{N}(x_t;\sqrt{1-\beta_t}x_{t-1}, \beta_t\mathbf{I})$ where $0 \,{<}\, ... \,{<}\, \beta_t \,{<}\, ...\,  \beta_T \,{<}\, 1$ defines a scaling schedule for the applied noise. This forward process can be re-written such that each time step depends only on the original data via the introduction of $\alpha_i = 1-\beta_i$ and $\Bar{\alpha}_t=\prod_{i=1}^t\alpha_i$ such that $q(x_t|x_o)=\mathcal{N}(x_t;\sqrt{\Bar{\alpha}_t}x_0, (1-\Bar{\alpha}_t)\mathbf{I}))$. Given Gaussian noise $\epsilon_0\sim\mathcal{N}(\mathbf{0}, \mathbf{I})$, the reparameterization trick can then be recursively applied to derive a closed form of any $x_t$ as explained in \cite{luo2022understanding}:
\begin{equation}
    x_t = \sqrt{\Bar{\alpha}_t}x_0 + \sqrt{1-\Bar{\alpha}_t}\epsilon_0 \>.
    \label{eq:closedformxt}
\end{equation}

Via the assumptions above on the structure of the Hierarchical VAE, the reverse diffusion process then can be defined by a separate discrete Markov chain as
\begin{equation}
    p(x_{t-1}|x_t)=\mathcal{N}(x_{t-1};\mu_{q}(x_t, x_0),\Sigma_q(t))
    \label{eq:reversediffmc}
\end{equation}
where $\bm\mu_q$ is a function of both $x_t$ and $x_0$, and $\Sigma_q(t) = \sigma_q^2(t)\mathbf{I}$ where
\begin{equation}
    \sigma_q^2(t) = (1-\alpha_t)(1-\Bar{\alpha}_{t-1})/(1-\Bar{\alpha}_{t}) \>.    
\end{equation}
However, in the generative case, we do not have access to the ground truth $x_0$ and must construct a corresponding $\bm\mu_{\theta}(x_t, t)$, which predicts $\bm\mu_{q}(x_t, x_0)$ and depends only on $x_t$, via the application of \cref{eq:closedformxt}. This $\bm\mu_{\theta}$ is learned via a Denoising Neural Network.

Computing this $\bm\mu_{\theta}(x_t, t)$ requires learning an estimation $\hat{x}_{\theta}(x_t, t)$ of the original $x_0$. This can be further simplified via an additional application of the reparameterization trick such that, instead of learning $\hat{x}_{\theta}(x_t, t)$, the goal is to learn a function $\hat{\epsilon}_{\theta}(x_t, t)$ to predict the source noise $\epsilon_0$, i.e., the noise that when added to $x_0$ produces $x_t$. Given all of the simplifications on $\bm\mu_{\theta}(x_t, t)$, the final ELBO objective becomes \cite{luo2022understanding}:
\begin{equation}
    \arg \min_{\theta} ||\epsilon_0 - \hat{\epsilon}_{\theta}(x_t, t)||^2_2
    \label{eq:elboobjective}
\end{equation}
This estimated noise $\hat{\epsilon}_{\theta}(x_t, t)$ enables the computation of the mean 
\begin{equation}
    \bm\mu_{\theta}(x_t, t) = \frac{1}{\sqrt{\alpha_t}} x_t - \frac{1-\alpha_t}{\sqrt{1-\Bar{\alpha}_t}\sqrt{\alpha_t}}\hat{\epsilon}_{\theta}(x_t, t) \>,
\end{equation}
which in turn provides the denoising step:
\begin{equation}
    x_{t-1} = \frac{1}{\sqrt{\alpha_t}} \left(x_t - \frac{1-\alpha_t}{\sqrt{1-\Bar{\alpha}_t}}\hat{\epsilon}_{\theta}(x_t, t)\right) + \sigma_t \epsilon_0\>.
\end{equation}

\subsection{Guiding the diffusion process}

Under the continuous-time interpretation, the diffusion process can be described as a direction through the latent space, from a standard Gaussian Distribution towards the data generating distribution \cite{song2020score}. This direction formulation ties directly into the use of conditioning information to guide the diffusion process, changing the direction through the latent spaces towards a specific part of the data generating distribution.

Several methods of guidance have been proposed. This work relies on Classifier Free Guidance \cite{ho2022classifier}. Under the Classifier Free Guidance formulation, two separate diffusion models estimate the noise added to the same example. The first model is unconditioned, and the second uses the conditioning information. By evaluating the difference between the predicted noise by both models, the direction enforced by the conditioning information is found. A utility parameter $\lambda\geq0$ then controls the degree to which the conditioning information is used during inference. 

The result is that, during sampling, the estimated noise is predicted using: 
\begin{equation}
    \hat{\epsilon}_{\theta}(x_t, t) + \lambda(\hat{\epsilon}_{\theta}(x_t, t, y) - \hat{\epsilon}_{\theta}(x_t, t)) \>.
    \label{eq:noisebalancewithguidance}
\end{equation}
Ultimately it is not necessary to train two different diffusion models. During training, the conditioning information is replaced by a null token in a manner which imitates random dropout. A single diffusion model is thus capable of performing Classifier Free Guidance \cite{luo2022understanding}.

\subsection{Diffusion transformer}

\paragraph{DiT.}

\begin{figure}
    \centering
    \includegraphics[width=1\linewidth]{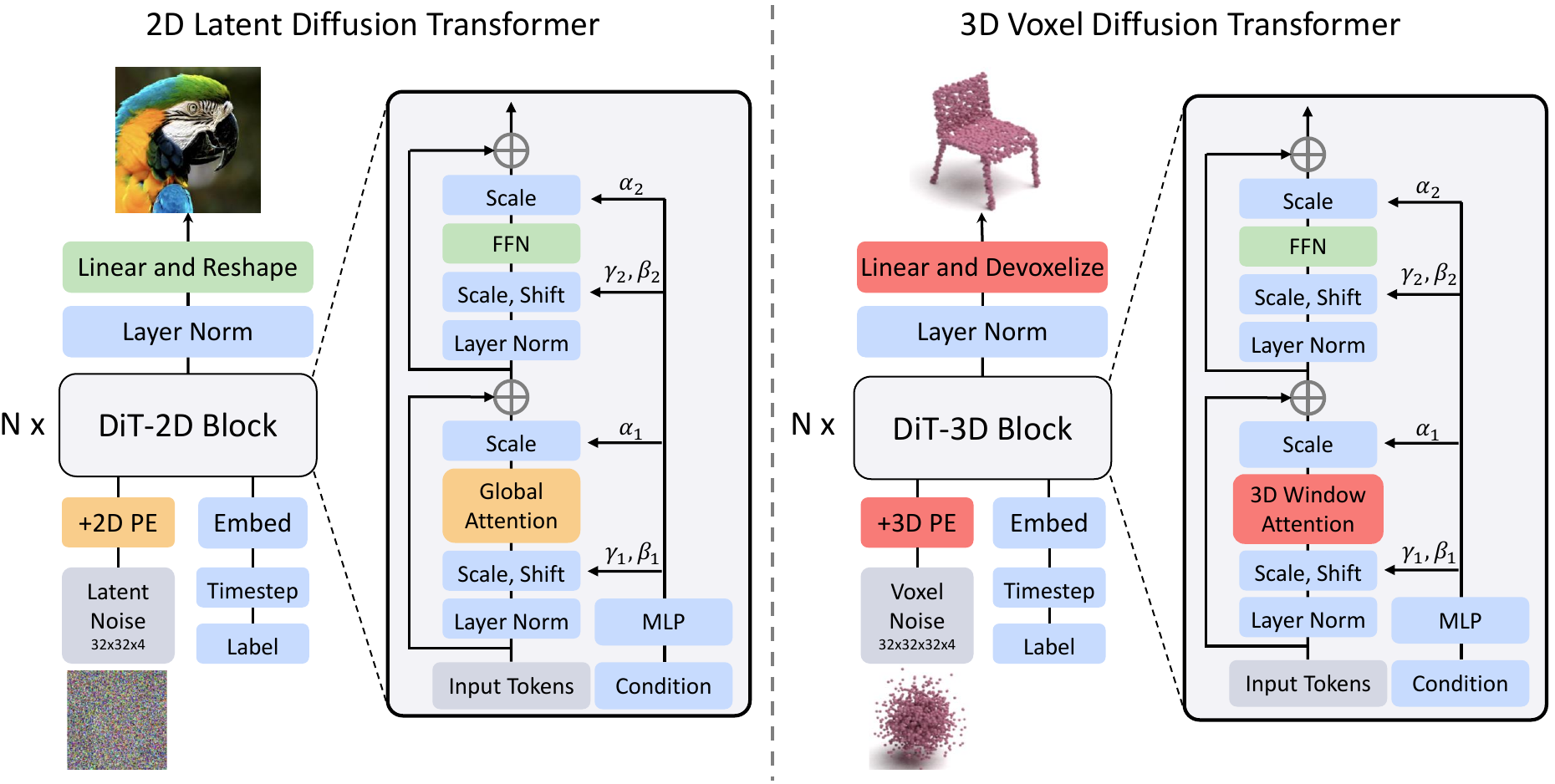}
    \caption{Comparison of two Diffusion Transformer architectures. On the left is the classic 2D DiT. Patch embeddings are fed into classical self-attention layers. Note the Adaptive Layer Norm layers to apply conditioning information. On the right is the DiT-3D, which introduces a voxelization step to embed 3D data. (The image is taken from \cite{mo2023dit}.)}
    \label{fig:dit-arch}
\end{figure}

In \cite{peebles2023scalable}, Peebles et al.\ introduced a scalable denoising architecture based on the transformer \cite{vaswani2017attention}. The authors demonstrate that the attention mechanism can be adapted to predict added noise to 2D images, and then trained via the diffusion objective to produce a state of the art 2D image generation model. This work contrasts with previous diffusion methodology by focusing on architectural choices in regards to both the quality of the generated results and the scalability of the model. Prior to the introduction of the Diffusion Transformer (DiT), most works applied the diffusion objective via a denoising U-Net, i.e., a classical convolutional U-Net trained to predict added noise to an image. The novel architecture of the DiT is shown in \cref{fig:dit-arch} (left) of this appendix. DiT uses a standard self-attention layer followed by a feed-forward architecture, but uses adaptive layer-norm layers to propagate conditioning information, such as the time step or class label. The conditioning information is used to the scale and shift parameters of the normalization layers via an MLP. Applying conditioning information in adaptive normalization layers is a trick introduced first by StyleGAN \cite{karras2019style}, and remains a popular method for applying conditioning information.

\paragraph[\pixart-alpha]{\pixartal.}

\begin{figure}
    \centering
    \includegraphics[width=0.5\linewidth]{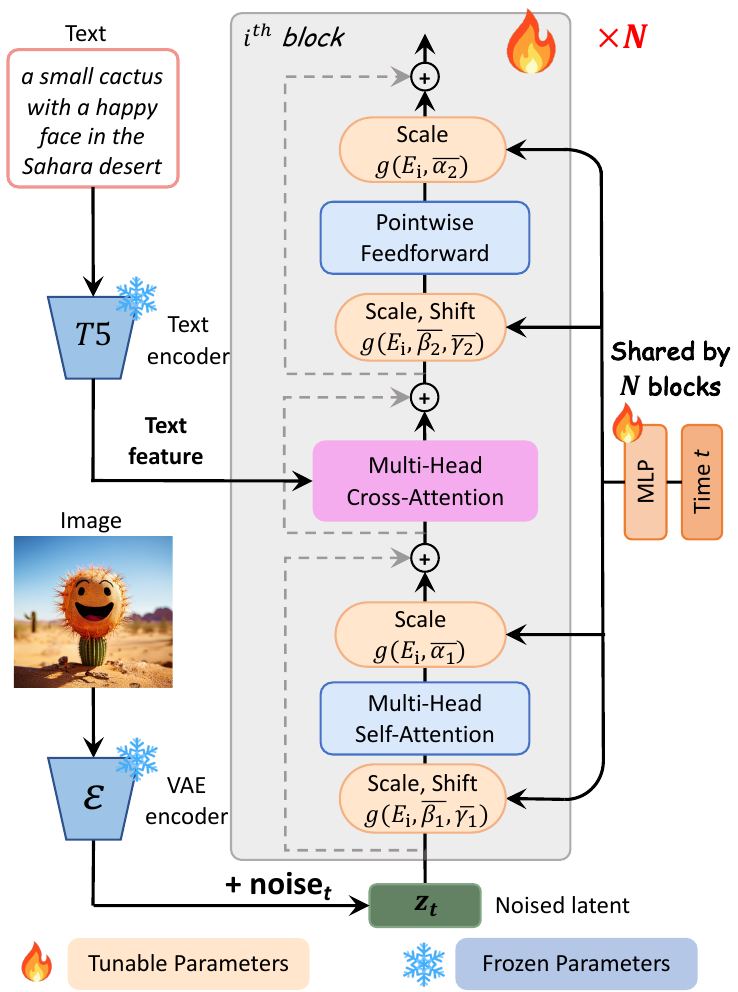}
    \caption{The \pixartal architecture, from \cite{chen2023pixart}. Note the shared shift and scale parameters and the introduction of cross attention layers to propagate conditioning via text tokens. (The image is taken from \cite{chen2023pixart}.)}
    \label{fig:pix-arch}
\end{figure}

A state-of-the-art variant of the Diffusion Transformer, modified to use text prompts as conditioning, was introduced in \cite{chen2023pixart}. This model, called \pixartal, introduces cross-attention layers between the input features and the embedded text prompts to guide the diffusion process. The authors also introduce an optimization of the adaptive layer norm architecture, where a single MLP is used to learn the scale and shift parameters for the entire model. \cref{fig:pix-arch} of this appendix shows the architecture of this model. \pixartal is significant to LiDAR object generation because it provides guidelines for how to apply more complex conditioning guidance. The original DiT paper uses only class labels as conditioning.

\paragraph{DiT-3D.}

The Diffusion Transformer has been adapted for a 3D shape generation task, where it achieves state-of-the-art results \cite{mo2023dit}. This \dittd introduce a voxelization step, which converts the unstructured 3D point cloud data into features of a 32{$\,\times\,$}32{$\,\times\,$}32 voxel grid. These voxels are then grouped into patches and embedded. This makes it possible to use more or less the same overall architecture as the Diffusion Transformer, using the same adaptive Layer Norm layers to apply conditioning information (cf.\ \cref{fig:dit-arch} (right) of this appendix). This allows the leveraging of models pre-trained on large 2D datasets for 3D generative tasks.

\paragraph{3D diffusion transformers.}

The significant aspects of the Diffusion Transformer architecture, in regards to the LiDAR object generation task, can be summarized as follows: 
\begin{itemize}
    \item \emph{Easily scalable:} DiT-based architectures build upon the wealth of literature regarding the scaling of parameter size of transformer based architectures. This is especially important in LiDAR object generation where the variability in point cloud sizes between objects at different locations makes memory consumption per batch highly irregular. Thus it is useful to be able to train small models with the justified expectation of improved performance with increased size.
    \item \emph{Positional embeddings by default:} In the LiDAR object generation task, absolute positioning information is inherent to the conditioning of the model. Transformer architectures include explicit layers to encode absolute position information into the training process, and thus this absolute positional knowledge is provided by default to the model.
\end{itemize}

\paragraph{Architecture of \ours, \dittdl and \pixartl}

Our baselines \dittdl and \pixartl, whose architecture is shown Figure 2 (of main paper) together with \ours, are directly adapted from their original counterparts to handle: (i) LiDAR data, i.e., point embedding, and (ii) LiDAR object conditioning. LOGen is then a modification of PixArt-L, as described in main paper.

\section{Implementation details}
\label{sec:implemdetails}


\paragraph{Sparse convolutional architecture.} For experiments using a sparse convolutional architecture, we follow the same model structure and size as LiDiff \cite{nunes2024lidiff}, using MinkUNet, the sparse convolutional U-Net introduced in \cite{choy20194d}. The additional conditioning information is concatenated onto the time step information. The MinkUNet models contain roughly 36M parameters.

\paragraph{Transformer architectures.} Unless otherwise stated, the transformer architectures used were extra small (XS) variants of the DiT. These models contain 12 transformer blocks, each with 3 heads, and an embedding dimension of 192. Where voxelization was used, a patch size of 4 was used. This means that four voxels were combined into a single patch. The XS transformer variants contain roughly 7.5M parameters.

To apply the additional conditioning information when testing vanilla DiT-3D variants, the new conditions were embedded and then concatenated to the time step. The resulting vector was then flattened and used as input to the MLP to learn the adaptive layer-norm parameters. For the cross-attention variants, the condition was used as input only to the cross-attention layers and the adaptive layer-norm was learned only using the time step. For all variants, the adaptive layer-norm MLP was initialized to zero following \cite{peebles2023scalable}.

\paragraph{Diffusion parameters.} The DDPM scheduler was used with 1000 time steps during training. During inference 500 time steps were used. The $\beta$ schedule was linearly spread on the range of $[3.5\ 10^{-5}, 0.007]$, as in LiDiff. The parameter $\lambda$ controlling the strength of the guidance was set to~1.0 (best value from a grid search, while LiDiff uses $\lambda = 6.0$). This value of $\lambda$ cancels out the unconditional score term in \cref{eq:noisebalancewithguidance}. However, this equation is only used during the sampling procedure (which we do at inference or validation). During training, we periodically use null conditions for 10\% of the conditions in a batch (as LiDiff does). So the classifier free guidance does have an impact on the training procedure.

\paragraph{Training.} The Adam optimizer was used with a constant learning rate of $0.0001$.
We trained small classes (i.e., barrier, bike, car, motorcycle, pedestrian, traffic cone) for 500,000 iterations with a batch size of~16. The remaining large object classes (i.e., bus, construction vehicle, trailer and truck) are trained for 5,000,000 iterations with a batch size~4. All trainings were conducted on 4 Nvidia A100 GPUs.

\section{Metrics}
\label{sec:metrics}

We provide below a formal definition of the metrics used in the paper.

\paragraph{Chamfer distance (CD), Earth mover distance (EMD\rlap{\textbf{).}}}

We evaluate object-level metrics using standard CD and EMD distances to measure the similarity between two point clouds.
\begin{equation}
    \textrm{CD}(X,Y) = \sum_{x\in X}\min_{y\in Y}||x-y||^2_2 + \sum_{y\in Y}\min_{x\in X}||x-y||^2_2 \>,
\end{equation}

\begin{equation}
    \text{EMD}(X,Y)= \min_{\gamma:X\rightarrow Y}\sum_{x\in X} ||x-\gamma(x)||_2 \>, 
\end{equation}
where $\gamma$ is a bijection between point clouds $X$ and $Y$. (To follow previous work in this domain \cite{yang2019pointflow, zhou2021pvd, vahdat2022lion}, we use the square distance for the CD computation.)

\paragraph{Coverage (COV).}

Given a set of real object samples $S_r$ and a set of generated objects~$S_g$, COV \cite{achlioptas2018learningrepres} measures the proportion of real pointsets in $S_r$ matched to at least one generated pointset in $S_g$. Although COV does not assess the quality of the generated pointsets, it serves as a metric to quantify its diversity. Some low-quality sets of point clouds can however obtain high COV scores.
\begin{equation}
    \textrm{COV}(S_g,S_r)=\frac{|\{\textrm{arg min}_{Y\in S_r} D(X,Y)|X\in S_g\}|}{|S_r|} \>.
\end{equation}
In our experiments, we have used both CD and EMD to compute $D(\cdot,\cdot)$. For~$S_r$, we use all objects of a given class, and for~$S_g$, we generate as one sample per sample in $S_r$ with corresponding conditioning.

\paragraph{1-nearest neighbor accuracy (1-NNA).}

1-NNA \cite{lopezpaz2018revisiting, yang2019pointflow} evaluates both the diversity and quality aspects in a single metric. Like COV, it relies on two sets $S_r$ and $S_g$ of, respectively, real and generated pointsets. Real and synthetic objects are merged and 1-NNA calculates the proportion of nearest neighbors for each sample that come from the same source. The 1-NNA score ranges from 100\%, when the distributions are distinct, to 50\%, when they are indistinguishable.
\begin{multline}
     \hspace{-3.6mm}\textrm{1-NNA}(S_g,S_r)= \\ \frac{\sum_{X\in S_g}\mathbb{I}[N_X\in S_g]+\sum_{Y\in S_r}\mathbb{I}[N_Y\in S_r]}{|S_g|+|S_r|} \>,
\end{multline}
where $\mathbb{I}[\cdot]$ denotes the indicator function and $N_X$ represents the nearest neighbor of point cloud $X$ in the set $(S_r \cup S_g) \setminus \{X\}$. 
In our experiments, we have used both CD and EMD as distance metrics to calculate the nearest neighbors, and $S_r$ and $S_g$ are defined as for COV.

\paragraph{Fr\'echet Point Cloud Distance (FPD).}

FPD \cite{shu2019treegan} is an adaptation of the classic Fréchet Inception Distance (FID) \cite{heusel2017fid}, where 2D Inception features are replaced with 3D PointNet features \cite{qi2017pointnet}. Like FID, FPD computes the 2-Wasserstein distance between the real and generated Gaussian distributions in the feature space\rlap:
\begin{multline} \label{eq:fpd}
   \hspace{-3.6mm}\text{FPD}(S_r, S_g) = \\ \norm{\mathbf{m}_{S_r} - \mathbf{m}_{S_g}}_2^2 + \text{Tr}(\Sigma_{S_r}\,{+}\,\Sigma_{S_g}\,{-}\, 2(\Sigma_{S_r}\Sigma_{S_g})^{\frac{1}{2}} ),
\end{multline}
where $\mathbf{m}_{S_r}$ and $\Sigma_{S_r}$ represent  the mean vector and covariance matrix of the points calculated from real point clouds in $S_r$, while  $\mathbf{m}_{S_g}$ and $\Sigma_{S_g}$ are the mean vector and covariance matrix calculated from generated point clouds in $S_g$. $\text{Tr}(A)$ is the sum of the elements along the main diagonal of matrix $A$.

The original FPD is based on the PointNet classifier trained on ModelNet-40 \cite{qi2017pointnet} with dense and uniform point sampling. To account for the sparse and non-uniform characteristics of LiDAR-scanned objects in nuScenes, we instead use a separate PointNet classifier that has been trained on the real LiDAR data from nuScenes.
Like with the original FPD, we extract a 1803-dimensional feature vector (5 fewer dimensions than the original 1808, due to the presence of 5 fewer classes) from the output of the dense layers, to calculate the mean and covariance in~\cref{eq:fpd}.

\paragraph{Kernel Point Cloud Distance (KPD).}

Likewise, we adapt the Kernel Inception Distance (KID)~\cite{binkowski2018kid},
replacing 2D Inception features with our 1803-dimensional 3D PointNet features. Like KID, KPD  calculates the maximum mean discrepancy (MMD) between the real and generated instances in the feature space:
\begin{equation} \label{eq:kpd}
\text{KPD}(S_r, S_g) = \text{MMD}(S_r, S_g)^2 \>.
\end{equation}

\paragraph{Accuracy of PointNet Classifier (APC).}

We also measure the accuracy of the PointNet classifier, i.e., trained on real objects, for recognizing synthetic objects. It is to be compared to the accuracy of the classifier for recognizing real objects.

\section{Distribution of evaluated objects}
\label{sec:datadistrib}

In LiDAR datasets, the number of objects displays large variations among classes, not to mention the number of points in these objects depending on the classes, distance to the sensor and orientation. As an illustration, the class\-wise frequency of objects in nuScenes is pictured on \cref{fig:distrib}. More statistics are in \cite{caesar2020nuscenes} and Table~\ref{tab:datadistrib}.

In nuScenes data, some objects can have as few as 1 point. Although some of our experiments (not reported here) went as low as keeping objects with at least 5 points, evaluating the quality of generation with so few points makes little sense. Therefore, we set a minimum threshold to 20 points for an object to be used for training and for evaluation, which covers more than 95\% of all object points. It can also be noted that the official instance segmentation evaluation of nuScenes only considers instances with at least 15 points~\cite{fong2022panoptic}.

\section{Evaluation of conditioning}
\label{sec:conditioning}

\paragraph{Evaluation of conditioning on sensor viewing angle} To assess \ours's ability to generate realistic objects regardless of viewing angle, we modify the synthetic replacements by rotating their angular condition. This reduces the need to find suitable locations for new objects. We consider five evenly spaced angles around a full turn, i.e., $\nicefrac{i}{5} \times 360$\textdegree\ where $i \in \{0..4\}$, as shown in Figure 1.

Tab.\,\ref{tab:conditioning_ablation_rot} shows a moderate variation of the performance when rotating objects. It is partly due to collisions between objects that occur after rotation (the insertion is performed without such a test). Further, the distribution of object viewing angles in the dataset is not uniform. Less viewed angles can result in noisier generations.

\begin{figure}[t]
\centering
\scalebox{0.9}{%
    \resizebox{1.\columnwidth}{!}{
\begin{tikzpicture}
    \begin{axis}[
        ybar,
        ylabel={Frequency},
        symbolic x coords={barrier, bike, bus, car, constr.\ veh., m.cycle, pedestrian, traffic cone, trailer, truck},
        xtick=data,
        ymin = -0.001,
        bar width=0.5cm,
        ytick={0,0.2,0.4,0.6},  
        x tick style={draw=none},
        grid=none,
        ymajorgrids = true,
        height=8cm,
        width=15cm,
        xticklabel style={font=\normalsize, rotate=45, anchor=north east}, 
        every axis x label/.append style={yshift=1.5em} 
    ]
    \addplot[fill=Purple!70, ybar, draw=none, bar width=0.9cm, thick]  coordinates {
        (traffic cone, 0.02951794673)  [you]
        (bike, 0.005605097438) 
        (m.cycle, 0.01008247607)
        (constr.\ veh., 0.01972577452)
        (trailer, 0.04234714386)
        (car, 0.4889237915)
        (bus, 0.02988640931)
        (pedestrian, 0.0816647065)
        (truck, 0.1234610174)
        (barrier, 0.1687856367)
    };
    \end{axis}
\end{tikzpicture}
}
    }
    \vspace*{-3mm}
    \caption{\textbf{Frequency of objects per class} in nuScenes, considering objects with at least 20 points.}
    \label{fig:distrib}
\label{fig:imbalance}
\end{figure}

\begin{table}[t]
    \centering

    \newcommand\turn[1]{{\small+}$\nicefrac{#1}{5}\,$ 360\textdegree}
    \renewcommand\turn[1]{rot.~$\nicefrac{#1}{5}$}
        \vspace{0pt} 
        \centering
        \begin{tabular}{l c c c}
            \toprule
            \bf Test & \bf Box & \multicolumn{2}{c}{\bf mIoU\,(\%)\,$\uparrow$~} \\
            \bf objects & \bf orientation & 3 ch. & 4 ch.  \\
            \midrule
            real & orig. & 74.5 & 76.3   \\
            \midrule   
            \ours  & orig.  & 63.7  & 69.1 \\
            \ours & \turn 1 & 53.7  & 64.6  \\
            \ours & \turn 2 & 54.8  & 66.0  \\
            \ours & \turn 3 & 53.8  & 63.8  \\
            \ours & \turn 4 & 53.3  & 64.1 \\
            \bottomrule
        \end{tabular}
        \captionof{table}{\textbf{Evaluation of conditioning parameters: sensor viewing angle.} As in Tab.\,\ref{tab:mix_training} of Sec.\,\ref{sec:downstream}, 50\% of real objects are replaced by generated objects, and we evaluate semantics segmentation (mIoU\%). Here we generate new objects with a given rotation relatively to the original viewing angle.}
        \label{tab:conditioning_ablation_rot}
\end{table}
\begin{table}[t]
    \centering
        \centering
        \setlength{\tabcolsep}{4pt}
        \begin{tabular}{c @{~}| c c c c | c c}
            \toprule
            \textbf{Distance}  & \multicolumn{2}{c}{\bf FPD\,$\downarrow$}  & \multicolumn{2}{c}{\bf KPD\,$\downarrow$} & \multicolumn{2}{c}{\bf APC\,(\%)\,$\uparrow$} \\
            \textbf{to sensor} & 3~ch. & 4~ch. & 3~ch. & 4~ch. & 3~ch. & 4~ch.   \\
            \midrule
            orig.\ $d$  &  1.34   & 2.18 &   0.10 & 0.12 & 40.9 & 48.1\\
            $d\times 2$ & 4.79 & 12.70 & 0.61 & 1.71 & 34.0 & 38.4 \\
            $d \mathbin{/} 2$ &  3.08  & 8.13 & 0.33 & 0.85 & 38.3 & 49.8 \\
            \bottomrule
        \end{tabular}
        \captionof{table}{\textbf{Evaluation of conditioning parameters: distance to the sensor.} We vary the distance when generating objects.}
        \label{tab:conditioning_ablation_dist}
\end{table}

\begin{table*}[t]
    \centering\setlength{\tabcolsep}{4pt}
    \newcommand{\OD}{\hphantom{0}}
    \newcommand{\TD}{\hphantom{00}}
    \newcommand{\OC}{\hphantom{0,}}
    \newcommand{\TC}{\hphantom{00,}}
    \newcommand{\RC}{\hphantom{000,}}
        \begin{tabular}{l||cc||cccc|cc}
    & \multicolumn{2}{c||}{obj.\ size $|s|\,{<}$\,20 pts} & \multicolumn{6}{c}{objects size $|s|\,{\geq}$\,20 points}\\
    \midrule
          & \bf total & \bf total & \bf min & \bf median & \bf mean & \bf max & \bf total & \bf total \\
    \bf Class & \bf no.\,pts & \bf no. obj\rlap. & \bf no.\,pts & \bf no.\,pts & \bf no.\,pts & \bf no.\,pts & \bf no.\,pts & \bf no.\,obj\rlap. \\
    \toprule
barrier & \OC 493,786
& \OD 78,087 & 20 & 62 & 163 & \OD 2,667 & \OD 7,422,517 & \OD 45,350 \\
bicycle  & \TC 41,078
& \TD 8,021 & 20 & 36 & \OD 57 & \TC  838 & \RC 86,997 & \TD   1,506 \\
bus & \TC 46,248
& \TD 6,381& 20 & 96 &    536 &    11,587 & \OD 4,309,567 & \TD  8,030 \\
car   & 1,263,050
& 231,531 & 20 & 81 &    264 & \OD 8,695 & 34,783,810 &    131,366 \\
constr.\ vehicle & \TC 46,194
& \TD 6,932& 20 & 69 &    209 & \OD 8,641 & \OD 1,108,460 & \TD  5,300 \\
motorcycle & \TC 42,679
& \TD  7,620& 20 & 59 &    135 & \OD 1,952 & \TC 368,415 & \TD  2,709 \\
pedestrian & \OC 753,031
& 165,449& 20 & 37 & \OD 64 & \OD 1,192 & \OD 1,408,328 & \OD 21,942 \\
traffic cone & \OC 229,178
& \OD 62,951& 20 & 39 & \OD 51 & \TC  726 & \TC 408,092 & \TD   7,931 \\
trailer & \TC 66,359
& \TD 9,310& 20 & 85 &    380 & \OD 10,825 & \OD 4,333,342 & \OD 11,378 \\
truck & \OC 251,543
& \OD  39,061& 20 & 81 &    428 & \OD 10,719 & 14,197,897 & \OD 33,172 \\
\midrule
Overall & 3,233,146 
& 615,343 & 20 & 65 &    254 &    11,587 & 68,427,425 & 268,684
\end{tabular}
    \caption{\textbf{Distribution of object data in nuScenes.} We provide detailed statistics on objects with at least 20 points, which we use for our generation (training and testing). Objects with at least 20 points represent 95.5\% of all object points.}
    \label{tab:datadistrib}
\end{table*}

\begin{table*}[!t]
\centering
    \newcommand\turn[1]{{\small+}$\nicefrac{#1}{5}\,$ 360\textdegree}
    \renewcommand\turn[1]{rot.~$\nicefrac{#1}{5}$}
\begin{tabular}{c l@{~~}c | c c c c c c c c c c | c }
\toprule
& Test objects  &  Box orient. & bar. & bic. & bus & car & c.veh  & mot.  & ped. & t.con. & tra. & tru. & mean\\
\midrule

\parbox[t]{2mm}{\multirow{6}{*}{\rotatebox[origin=c]{90}{3 ch.}}} 
& real   & orig. &  80.3 & 48.8 & 90.7 & 94.2 & 40.6 & 86.0 & 085.5 & 73.4 & 60.7 & 85.2 & 74.5 \\
\cmidrule(lr){2-14}
& \ours    & orig. & 78.2 & 50.9 & 60.1 & 79.1 & 28.1 & 82.2 & 84.2 & 73.5 & 37.5 & 63.2 & 63.7 \\
& \ours    & \turn 1 & 71.1 & 45.0 & 41.1 & 68.5 & 19.9 & 72.6 & 80.4 & 67.2 & 22.7 & 48.6 & 53.7 \\
& \ours  & \turn 2 & 70.1 & 45.3 & 40.9 & 73.7 & 18.3 & 73.4 & 82.0 & 66.9 & 27.1 & 50.5 & 54.8 \\
& \ours  & \turn 3 & 71.2 & 42.4 & 39.7 & 65.3 & 16.5 & 73.9 & 81.7 & 68.8 & 26.6 & 52.2 & 53.8 \\
& \ours  & \turn 4 & 71.1 & 41.9 & 43.7 & 65.2 & 19.8 & 73.6 & 81.4 & 67.9 & 20.9 & 47.1 & 53.3 \\

\midrule
\parbox[t]{2mm}{\multirow{6}{*}{\rotatebox[origin=c]{90}{4 ch.}}} 
& real   & orig. &  81.6 & 50.4 & 90.9 & 94.6 & 44.5 & 88.1 & 85.6 & 77.4 & 64.6 & 85.6 & 76.3 \\
\cmidrule(lr){2-14}
& \ours    & orig. & 78.0 & 48.9 & 71.7 & 91.3 & 37.1 & 82.0 & 84.8 & 79.1 & 45.1 & 72.9 & 69.1 \\
& \ours    & \turn 1 & 71.8 & 46.6 & 64.0 & 87.2 & 32.8 & 76.5 & 84.0 & 75.5 & 42.7 & 64.6 & 64.6 \\
& \ours  & \turn 2 & 72.5 & 46.1 & 64.3 & 89.2 & 36.1 & 81.3 & 84.2 & 75.6 & 41.6 & 69.5 & 66.0 \\
& \ours  & \turn 3 & 73.8 & 46.9 & 58.3 & 87.9 & 28.6 & 75.5 & 84.3 & 76.1 & 43.4 & 63.3 & 63.8 \\
& \ours  & \turn 4 & 72.0 & 44.5 & 66.7 & 84.9 & 32.6 & 78.9 & 83.8 & 74.5 & 39.8 & 63.3 & 64.1 \\

\bottomrule
    \end{tabular}
    \hspace*{6mm}
    \caption{\textbf{Evaluation of conditioning: sensor viewing angle.} We present per-class IoU\% performances. }
\label{tab:evaluation_view}
\end{table*}

\paragraph{Evaluation of conditioning on distance to sensor.} To assess the quality of generated objects at varying sensor distances, we created two variants at distances $2\times d$ and $ d \mathbin{/} 2$ where $d$ is the original object distance. We scale the number of points from the initial size according to a power law \cite{clauset2009power} as the distance changes. Since ground truths at these distances are unavailable, we evaluate quality on feature-based metrics (Table~\ref{tab:conditioning_ablation_dist}). Additionally, integrating these objects into scenes for scene-level evaluation is nontrivial due to the distance variation.

\section{Detailed results}
\label{sec:moreresults}

\paragraph{Class-wise numbers on nuScenes.}

Statistics on the number of points and objects in nuScenes are in Table~\ref{tab:datadistrib}.

For our experiments on nuScenes, we provide here results per class: barrier (bar.), bicycle (bic.), bus, car, construction vehicle (c.veh), motorbike (mot.), pedestrian (ped.), traffic cone (t.con.), trailer (tra.), truck (tru.).

Complementary class-wise results regarding the evaluation of conditioning on sensor viewing angle are shown in  Table~\ref{tab:evaluation_view}. Results regarding the conditioning on distance to sensor are in Table~\ref{tab:evaluation_distance_fpd}, Table~\ref{tab:evaluation_distance_kpd}, Table~\ref{tab:evaluation_distance_apc}. 

For class-wise results on point-based metrics, refer to Table~\ref{tab:point_based_cd_emd}, Table~\ref{tab:point_based_1nn1}, Table~\ref{tab:point_based_cov}, and Table~\ref{tab:point_based_intensities}. For feature-based metrics, see Table~\ref{tab:feature_metrics_fpd}, Table~\ref{tab:feature_metrics_kpd}, and Table~\ref{tab:feature_metrics_apc}.

\begin{table}[t]
    \centering
    \vspace{-1mm}
    \begin{tabular}{l | c c c c c}
        \toprule
        Class & CD\,$\downarrow$ & EMD\,$\downarrow$ & JSD\,$\downarrow$ & FPD\,$\downarrow$ & KPD $\downarrow$ \\
        \midrule
        bicycle   & 0.28 & 0.30 & 0.24  & 3.00  & 0.03 \\
        car &  0.29 & 0.39 & 0.17 &1.00  & 0.21 \\
        person  &  0.14  & 0.11  & 0.18 &   1.29 & 0.11 \\
        \midrule
        \textbf{mean} & 0.24  & 0.27  & 0.19  & 1.76 & 0.12
        \\
        \bottomrule
    \end{tabular}
\caption{\textbf{Class-wise results of \ours on KITTI-360.}}
\label{tab:kitti360_all}
    \vspace{-2.3mm}
\end{table}

\paragraph{Class-wise numbers on KITTI-360.}

In Table~\ref{tab:kitti360_all}, we also provide class-wise results of KITTI-360, along with metrics CD, EMD, JSD, FPD and KPD.

\begin{table*}[!t]
\centering
\begin{tabular}{l@{~~}c | c c c c c c c c c c | c }
\toprule
  & \bf Distance to sensor & bar. & bic. & bus & car & c.veh  & mot.  & ped. & t.con. & tra. & tru. & mean\\
\midrule

\parbox[t]{2mm}{\multirow{4}{*}{\rotatebox[origin=c]{90}{3 ch.}}}
&  $d $ &  0.37 & 1.78 & 1.44 & 0.60 & 1.49 & 0.79 & 0.14 & 1.52 & 4.16 & 1.02 & 1.34 \\
\cmidrule(lr){2-13}
&  $d \mathbin{/} 2$ &  0.44 & 4.95 &  2.75 &  2.72  &  3.22 &  1.72  &  1.13 &  4.16 & 5.80  & 3.94  &  3.08 \\

& \small{$d\times 1.5$} & 1.19 & 1.13 &  1.46 & 0.85  &  2.19  &  1.45  &  3.64 &  0.77 &  6.19  & 1.23 &  2.01  \\
& \small{$d\times 2$}&  2.80 &  4.12 &  2.98  &  1.80  &  3.63  &  3.03  & 13.53 &  5.15 &  8.71  & 3.63  &  4.79  \\

\midrule
\parbox[t]{2mm}{\multirow{4}{*}{\rotatebox[origin=c]{90}{4 ch.}}} 
&  $d $ &  0.64 & 3.78 & 1.89 & 0.56 & 3.04 & 3.03 & 0.52 & 0.99 & 4.93 & 2.31 & 2.18 \\
\cmidrule(lr){2-13}
&  $d \mathbin{/} 2$ & 2.17 & 9.92 &  11.88  & 5.40 &  9.80  &  7.01  &  4.27 &  4.73 & 16.68  & 9.44 &  8.13 \\

& \small{$d\times 1.5$} &  1.57 &  4.48 & 6.20  &  4.64 &  7.35  & 6.70 &  8.08 &  5.65 &  5.48  & 6.04 &  5.62  \\

& \small{$d\times 2$}&  3.38 &  11.67 &  12.23  &  12.82  &  13.03  &  10.64  &  22.78 &  17.89 &  10.28  & 12.29  &  12.70  \\

\bottomrule
    \end{tabular}
    \hspace*{6mm}
    \caption{\textbf{Evaluation of conditioning on distance to sensor: class-wise FPD.}}
 \label{tab:evaluation_distance_fpd}
\end{table*}

\begin{table*}[!t]
\centering
\begin{tabular}{l@{~~}c | c c c c c c c c c c | c }
\toprule
  & \bf Distance to sensor & bar. & bic. & bus & car & c.veh  & mot.  & ped. & t.con. & tra. & tru. & mean\\
\midrule

\parbox[t]{2mm}{\multirow{4}{*}{\rotatebox[origin=c]{90}{3 ch.}}}
&  $d $ &  0.00 & 0.01 & 0.23 & 0.04 & 0.00 & 0.00 & -0.01 & 0.22 & 0.31 & 0.22 & 0.10 \\
\cmidrule(lr){2-13}
&  $d \mathbin{/} 2$ &  0.08 & 0.54 &  0.13 & 0.32  &  0.20  &  0.14 & 0.23 &  0.81 &  0.48  & 0.39  &  0.33  \\
   & \small{$d\times 1.5$} &  0.13 &  0.05 &  0.06  &  0.02 &  0.10 &  0.05  &  0.67 &  0.04 & 0.58  & 0.06  &  0.18  \\
   
  & \small{$d\times 2$}&  0.43 &  0.41 &  0.16  &  0.14  &  0.34  & 0.00 &  2.31 & 0.98 &  1.01  & 0.10  & 0.61  \\

\midrule
\parbox[t]{2mm}{\multirow{4}{*}{\rotatebox[origin=c]{90}{4 ch.}}} 
&  $d $ &  0.03 & 0.14 & 0.11 & 0.04 & 0.09 & 0.02 & 0.06 & 0.04 & 0.48 & 0.17 & 0.12 \\
\cmidrule(lr){2-13}
&  $d \mathbin{/} 2$ &  0.08 & 1.00 &  1.09 &  0.74  &  0.88  &  0.36  &  0.73 & 0.57 &  2.17  & 0.91  &  0.85 \\
   
   & \small{$d\times 1.5$} &  0.10 &  0.57 &  0.63  &  0.59  &  0.65  &  0.59  &  1.33 &  0.83 &  0.37  & 0.74 & 0.64 \\
   
  & \small{$d\times 2$}&  0.42 &  1.44 & 1.31  &  1.92  &  1.48  &  1.25 &  3.66 &  3.10 &  0.83  & 1.69 &  1.71  \\

\bottomrule
    \end{tabular}
    \hspace*{6mm}
    \caption{\textbf{Evaluation of conditioning on distance to sensor: class-wise KPD.} 
    }
\label{tab:evaluation_distance_kpd}
\smallskip
\begin{minipage}{\textwidth}{\small Note: KPD computation (like KID) uses an unbiased U-statistic estimator of MMD\textsuperscript{2} from finite samples. While its expected value equals the true (nonnegative) MMD\textsuperscript{2}, individual estimates can dip below zero due to sampling noise. This explains why some values in the table are ``slightly'' negative.}
\end{minipage}
\end{table*}

\begin{table*}[!t]
\centering
\begin{tabular}{l@{~~}c | c c c c c c c c c c | c }
\toprule
  & \bf Distance to sensor & bar. & bic. & bus & car & c.veh  & mot.  & ped. & t.con. & tra. & tru. & mean\\
\midrule

\parbox[t]{2mm}{\multirow{4}{*}{\rotatebox[origin=c]{90}{3 ch.}}}
&  $d$ & 65.7 & 6.3 & 7.8 & 49.4 & 30.8 & 19.8 & 86.3 & 94.1 & 32.0 & 16.1 & 40.9 \\
\cmidrule(lr){2-13}
&  $d \mathbin{/} 2$ &  67.1 & 4.9 & 11.7  &  46.3  & 40.1  &  22.7  &  90.1 &  84.5 &  27.7  & 17.0  &  38.3  \\

& \small{$d\times 1.5$} &  63.6 &  5.4 &  5.4 &  51.1  &  22.7  & 18.9  & 75.5 &  87.9 & 33.2  & 14.5  & 37.8  \\

& \small{$d\times 2$}&  60.9 &  3.2 &  3.6  &  53.2  &  14.3  &  19.1  & 66.9 &  73.7 & 32.1  & 12.5  &  34.0  \\

\midrule
\parbox[t]{2mm}{\multirow{4}{*}{\rotatebox[origin=c]{90}{4 ch.}}} 
&  $d$ &  72.9 & 8.6 & 9.6 & 58.5 & 49.0 & 24.5 & 97.7 & 94.3 & 49.9& 16.4 & 48.1  \\

\cmidrule(lr){2-13}
&  $d \mathbin{/} 2$ & 72.8 & 8.3 &  17.0  &  63.0  &  49.5  &  23.0  &  97.7 &  91.3 & 56.0  & 19.4  & 49.8  \\

& \small{$d\times 1.5$} &  72.2 &  8.3 & 5.1 &  44.8  &  39.9  &  21.8 & 90.1  & 82.3 & 49.0  & 14.5 &  42.8  \\

& \small{$d\times 2$}&  71.1 &  6.0 &  4.1  & 35.1  & 32.0  &  22.7  & 81.2 &  68.7 &  49.1  & 13.8  &  38.4  \\

\bottomrule
    \end{tabular}
    \hspace*{6mm}
    \caption{\textbf{Evaluation of conditioning on distance to sensor: class-wise APC (\%).}}
\label{tab:evaluation_distance_apc}
\end{table*}

\begin{table*}[!t]
\centering
\begin{tabular}{l l@{~}c | c c c c c c c c c c | c }
\toprule
& Model  & \bf Emb. & bar. & bic. & bus & car & c.veh  & mot.  & ped. & t.con. & tra. & tru. & mean\\
\midrule

\parbox[t]{2mm}{\multirow{5}{*}{\rotatebox[origin=c]{90}{CD}}} & \minkunet &  & 0.16 & 0.22  & 0.19 & 0.14 & 0.20 & 0.19 & 0.26 & 0.32 & 0.23 & 0.17  & 0.208  \\
& \dittdl   & \VE &  0.09 & 0.13  & 0.19 & 0.17 & 0.11 & 0.11 & 0.11 & 0.09  & 0.20 & 0.15  & 0.136  \\
& \dittdl   & \PE & 0.08 & 0.13  & 0.19 & 0.13 & 0.17 & 0.11 & 0.13 & 0.10  & 0.21 & 0.17  & 0.142  \\
& \pixartl  & \PE & 0.12 & 0.16  & 0.31 & 0.25 & 0.21 & 0.15 & 0.13 & 0.13  & 0.26 & 0.24  & 0.196  \\
& \ours     & \PE & 0.08 & 0.12  & 0.15 & 0.13 & 0.15 & 0.11 & 0.11 & 0.09  & 0.20 & 0.16  & 0.130 \\

\midrule
\parbox[t]{2mm}{\multirow{5}{*}{\rotatebox[origin=c]{90}{EMD}}} & 
\minkunet &  & 0.16 & 0.22  & 0.20 & 0.13 & 0.20 & 0.19 & 0.27 & 0.35  & 0.24 & 0.16  & 0.212 \\
& \dittdl   & \VE &  0.07 & 0.10  & 0.20 & 0.11 & 0.16 & 0.08 & 0.08 & 0.06 & 0.20 & 0.15  & 0.121  \\
& \dittdl   & \PE & 0.06 & 0.09  & 0.21 & 0.11 & 0.15 & 0.08 & 0.10 & 0.07  & 0.21 & 0.17  & 0.125  \\
& \pixartl  & \PE & 0.10 & 0.13  & 0.39 & 0.3 & 0.22 & 0.14 & 0.10 & 0.10  & 0.29 & 0.27 & 0.204  \\
& \ours     & \PE & 0.06 & 0.08  & 0.15 & 0.11 & 0.14 & 0.08 & 0.08 & 0.06  & 0.19 & 0.16  & 0.111 \\

\bottomrule
    \end{tabular}
    \hspace*{6mm}
    \caption{\textbf{Class-wise distance metrics CD and EMD.}}
    \label{tab:point_based_cd_emd}
\end{table*}

\begin{table*}[!t]
\centering
\begin{tabular}{l l@{~}c | c c c c c c c c c c | c }
\toprule
& Model  & \bf Emb. & bar. & bic. & bus & car & c.veh  & mot.  & ped. & t.con. & tra. & tru. & mean\\
\midrule

\parbox[t]{2mm}{\multirow{5}{*}{\rotatebox[origin=c]{90}{CD}}} & \minkunet &  & 77.1 & 75.8  & 83.7 & 72.3 & 81.7 & 82.7 & 79.3 & 80.2 & 84.6 & 77.4  & 79.5  \\
& \dittdl   & \VE & 73.2 & 71.1  & 79.2 & 77.4 & 85.4 & 71.6 & 70.8 & 65.3 & 83.1 & 80.2  & 75.8 \\
& \dittdl   & \PE & 68.6 & 72.8  & 76.5 & 71.2 & 82.2 & 71.6 & 69.4 & 64.8 & 80.0 & 76.8  & 73.4   \\
& \pixartl  & \PE & 75.5 & 78.0  & 76.8 & 72.6 & 84.7 & 73.2 & 73.1 & 71.9 & 82.6 & 76.4  & 76.5   \\
& \ours     & \PE & 67.4 & 75.8  & 76.1 & 71.6& 87.7 & 73.3& 66.3 & 64.9 & 83.2 & 73.7  & 74.1  \\

\midrule
\parbox[t]{2mm}{\multirow{5}{*}{\rotatebox[origin=c]{90}{EMD}}} & 
\minkunet &  & 71.0 & 70.5  & 81.7 & 67.4 & 74.3 & 74.7 & 74.0 & 77.8 & 78.0 & 74.2  & 74.4  \\
& \dittdl   & \VE &  64.8 & 67.7  & 79.6 & 73.0 & 75.4 & 72.6 & 62.7 & 59.6 & 77.3 & 75.6 & 70.9 \\
& \dittdl   & \PE & 62.4 & 70.8  & 79.6 & 67.8 & 73.8 & 73.7 & 66.8 & 63.9 & 74.8 & 72.7  & 70.7   \\
& \pixartl  & \PE & 68.6 & 72.4  & 82.2 & 73.8 & 76.1 & 69.3 & 61.0 & 67.2& 79.0 & 74.8 & 72.5  \\
& \ours     & \PE & 64.0 & 69.8  & 74.3 & 68.4 & 82.9 & 74.7 & 61.1 & 59.7 & 81.0 & 69.6  & 70.6  \\

\bottomrule
    \end{tabular}
    \hspace*{6mm}
    \caption{\textbf{Class-wise 1-NNA (\%) metric} using either CD or EMD as distance.}
    \label{tab:point_based_1nn1}
\end{table*}

\begin{table*}[!t]
\centering
\begin{tabular}{l l@{~}c | c c c c c c c c c c | c }
\toprule
& Model  & \bf Emb. & bar. & bic. & bus & car & c.veh  & mot.  & ped. & t.con. & tra. & tru. & mean\\
\midrule

\parbox[t]{2mm}{\multirow{5}{*}{\rotatebox[origin=c]{90}{CD}}} & \minkunet &  & 34.8 &  38.2 & 33.6 & 38.8 & 33.7 & 31.5 & 25.5 & 30.7 & 36.2 & 33.4  & 33.7  \\
& \dittdl   & \VE &  34.7 & 40.0  & 27.3 & 33.4 & 32.1 & 39.9 & 32.5 & 43.7 & 30.0 & 29.0  & 34.3   \\
& \dittdl   & \PE & 35.6 & 38.5  & 29.2 & 36.8 & 32.1 & 39.5 & 37.0 & 42.6 & 32.5 & 30.9  & 35.5   \\
& \pixartl  & \PE & 33.2 & 40.4  & 21.3 & 31.7 & 31.5 & 37.3 & 33.8 & 41.6 & 33.0 & 30.9  & 33.5   \\
& \ours     & \PE & 38.5 & 38.5  & 38.0 & 37.9 & 30.4 & 36.4 & 35.9 & 41.0 & 27.4 & 35.0 & 36.0  \\

\midrule
\parbox[t]{2mm}{\multirow{5}{*}{\rotatebox[origin=c]{90}{EMD}}} & 
\minkunet &  & 44.4 & 41.1  & 44.1 & 47.2 & 45.3 & 42.1 & 34.5 & 33.8 & 40.0 & 44.3  & 41.7  \\
& \dittdl   & \VE &  44.7 & 44.5 & 33.9 & 44.5 & 42.2 & 49.0 & 42.7 & 42.3 & 35.6 & 40.2 & 42.0  \\
& \dittdl   & \PE & 45.2 & 40.0  & 33.0 & 43.9 & 41.7 & 46.5 & 47.8 & 44.4 & 39.1 & 38.3  & 42.1  \\
& \pixartl  & \PE & 44.9 & 44.5  & 28.1 & 40.4 & 41.9 & 49.2 & 46.2 & 44.1 & 36.3 & 42.0  & 41.8  \\
& \ours     & \PE & 44.5  & 42.3  & 40.4 & 44.2 & 42.1 & 46.7 & 46.0 & 41.3 & 32.8 & 41.7  & 42.3  \\

\bottomrule
    \end{tabular}
    \hspace*{6mm}
\caption{\textbf{Class-wise COV (\%) metric} using either CD or EMD as distance.}
    \label{tab:point_based_cov}
\end{table*}

\begin{table*}[!t]
\centering
\begin{tabular}{l l@{~}c | c c c c c c c c c c | c }
\toprule
& Model  & \bf Emb. & bar. & bic. & bus & car & c.veh  & mot.  & ped. & t.con. & tra. & tru. & mean\\
\midrule

\parbox[t]{2mm}{\multirow{3}{*}{\rotatebox[origin=c]{90}{1-NNA}}}
& \dittdl   & \PE & 75.0  & 61.9 & 76.0 & 69.7 & 71.7 & 68.3 & 82.6 & 88.3 & 77.2  & 73.1 & 74.4 \\
& \pixartl  & \PE & 73.9  & 56.9 & 80.4 & 72.0 & 73.2 & 66.6 & 74.9 & 62.7 & 76.2  & 79.2  &  71.6 \\
& \ours     & \PE & 70.8  & 60.4 & 69.0 & 68.7 & 70.1 & 64.8 & 75.0 & 62.4 & 71.7  & 69.5 & 68.2 \\

\midrule
\parbox[t]{2mm}{\multirow{3}{*}{\rotatebox[origin=c]{90}{COV}}}  
& \dittdl   & \PE & 36.1  & 38.9 & 28.8 & 38.8 & 35.0 & 37.9 & 23.8 & 15.8 & 26.7 & 32.4 &  31.4 \\
& \pixartl  & \PE & 35.0  & 38.5 & 27.7 & 36.6 & 32.4 & 37.5 & 31.2 & 35.5 & 28.3  & 26.6 & 32.9 \\
& \ours     & \PE & 39.5 & 39.7  & 37.4 & 39.2& 40.7 & 35.7 & 32.6 & 32.7 & 35.5 & 37.1  & 37.0 \\ 

\bottomrule
    \end{tabular}
    \hspace*{6mm}
    \vspace*{-1mm}
    \caption{\textbf{Point-based metrics.} We present class-wise results of COV (\%) and 1-NNA (\%) on intensity realism.}
    \label{tab:point_based_intensities}
\end{table*}
\begin{table*}
\centering
\begin{tabular}{l l@{~}c | c c c c c c c c c c | c }
\toprule
& Model  & \bf Emb. & bar. & bic. & bus & car & c.veh  & mot.  & ped. & t.con. & tra. & tru. & mean\\
\midrule

\parbox[t]{2mm}{\multirow{5}{*}{\rotatebox[origin=c]{90}{3 ch.}}} & \minkunet &  & 23.15 & 7.63 & 8.14 & 0.32 & 1.68 & 8.62 & 72.06 & 115.69 & 13.44 & 3.23& 25.4  \\
& \dittdl   & \VE &  3.31 & 2.86 & 6.08 & 0.70 & 2.96 & 0.87 & 0.65 & 0.69 & 5.78 & 4.09 & 2.80 \\
& \dittdl   & \PE & 3.27 & 3.19 & 4.26 & 0.41 & 1.32 & 0.85 & 3.17 & 1.76 & 4.95 & 2.96 & 2.62 \\
& \pixartl  & \PE & 0.30 & 4.11 & 5.80 & 1.17 & 2.42 & 0.93 & 1.45 & 1.13 & 7.77 & 4.37 & 2.95 \\
& \ours     & \PE & 0.37 & 1.78 & 1.44 & 0.60 & 1.49 & 0.79 & 0.14 & 1.52 & 4.16 & 1.02 & 1.34 \\

\midrule
\parbox[t]{2mm}{\multirow{3}{*}{\rotatebox[origin=c]{90}{4 ch.}}} 
& \dittdl   & \PE & 4.73 & 7.36 & 4.78 & 1.52 & 4.0 & 4.74 & 20.09 & 8.44 & 9.04 & 5.16 & 6.99 \\
& \pixartl  & \PE & 2.67 & 4.62 & 10.26 & 2.19 & 3.52 & 2.97 & 3.18 & 1.03 & 11.64 & 9.46 & 5.16 \\
& \ours     & \PE & 0.64 & 3.78 & 1.89 & 0.56 & 3.04 & 3.03 & 0.52 & 0.99 & 4.93 & 2.31 & 2.18 \\

\bottomrule
    \end{tabular}
    \hspace*{6mm}
    \vspace*{-1mm}
    \caption{\textbf{Feature-based metrics.} We present class-wise results of FPD.}
 \label{tab:feature_metrics_fpd}
\end{table*}

\begin{table*}
\centering
\begin{tabular}{l l@{~}c | c c c c c c c c c c | c }
\toprule
& Model  & \bf Emb. & bar. & bic. & bus & car & c.veh  & mot.  & ped. & t.con. & tra. & tru. & mean\\
\midrule

\parbox[t]{2mm}{\multirow{5}{*}{\rotatebox[origin=c]{90}{3 ch.}}} & \minkunet &  &  3.68 & 0.73 & 0.98 & 0.00 & 0.04 & 1.16 & 8.34 & 14.84 & 1.65 & 0.43 & 3.18  \\
& \dittdl   & \VE &  0.30 & 0.12 & 0.61 & 0.07 & 0.18 & 0.02 & 0.08 & 0.09 & 0.54 & 0.44 & 0.25 \\
& \dittdl   & \PE & 0.40 & 0.17 & 0.37 & 0.05 & 0.04 & -0.01 & 0.49 & 0.19 & 0.49 & 0.28 & 0.25 \\
& \pixartl  & \PE & 0.01 & 0.22 & 0.53 & 0.12 & 0.11 & 0.00 & 0.16 & 0.11 & 0.70 & 0.48 & 0.24 \\
& \ours     & \PE & 0.00 & 0.01 & 0.23 & 0.04 & 0.00 & 0.00 & -0.01 & 0.22 & 0.31 & 0.22 & 0.10 \\

\midrule
\parbox[t]{2mm}{\multirow{3}{*}{\rotatebox[origin=c]{90}{4 ch.}}} 
& \dittdl   & \PE & 0.49 & 0.33 & 0.42 & 0.10 & 0.24 & 0.14 & 2.80 & 1.50 & 0.93 & 0.53 & 0.75 \\
& \pixartl  & \PE &  0.18 & 0.05 & 0.81 & 0.15 & 0.07 & 0.02 & 0.39 & 0.07 & 1.28 & 1.04 & 0.41 \\
& \ours     & \PE & 0.03 & 0.14 & 0.11 & 0.04 & 0.09 & 0.02 & 0.06 & 0.04 & 0.48 & 0.17 & 0.12 \\

\bottomrule
    \end{tabular}
    \hspace*{6mm}
    \vspace*{-1mm}
    \caption{\textbf{Feature-based metrics.} We present class-wise results of KPD. (See also caption of Tab.\,\ref{tab:evaluation_distance_kpd}.)}
  \label{tab:feature_metrics_kpd}
\end{table*}

\begin{table*}
\centering
\begin{tabular}{l l@{~}c | c c c c c c c c c c | c }
\toprule
& Model  & \bf Emb. & bar. & bic. & bus & car & c.veh  & mot.  & ped. & t.con. & tra. & tru. & mean\\
\midrule

\parbox[t]{2mm}{\multirow{6}{*}{\rotatebox[origin=c]{90}{3 ch.}}} & \minkunet &  & 34.4 & 2.2 & 3.1 & 45.1 & 24.0 & 10.8 & 44.9 & 34.4 & 14.0 & 12.9 & 22.6  \\
& \dittdl   & \VE &  62.8 & 5.2 & 3.6 & 46.5 & 34.7 & 19.8 & 89.5 & 93.3 & 26.1 & 16.2 & 39.8 \\
& \dittdl   & \PE & 63.4 & 7.1 & 4.7 & 42.9 & 26.1 & 19.8 & 86.5 & 83.8 & 23.0 & 14.2 & 37.2 \\
& \pixartl  & \PE & 65.4 & 6.7 & 3.7 & 39.1 & 28.9 & 22.7 & 81.9 & 88.3 & 27.0 & 14.3 & 37.8 \\
& \ours     & \PE & 65.7 & 6.3 & 7.8 & 49.4 & 30.8 & 19.8 & 86.3 & 94.1 & 32.0 & 16.1 & 40.9 \\
\cmidrule(lr){2-14}
& \multicolumn{2}{l|}{Real dataset objects} & 66.3 & 5.6 & 8.0 & 45.6 & 27.5 & 17.8 & 83.7 & 88.5 & 30.3& 16.6 & 39.0 \\

\midrule
\parbox[t]{2mm}{\multirow{4}{*}{\rotatebox[origin=c]{90}{4 ch.}}} 
& \dittdl   & \PE & 71.1 & 8.6 & 8.9 & 45.5 & 33.6 & 20.5 & 80.8 & 93.9& 39.6 & 13.7 & 41.7 \\
& \pixartl  & \PE & 71.8 & 8.9 & 6.5 & 46.2 & 44.4 & 24.9 & 93.7 & 91.5 & 50.1 & 15.9 & 45.5 \\
& \ours     & \PE & 72.9 & 8.6 & 9.6 & 58.5 & 49.0 & 24.5 & 97.7 & 94.3 & 49.9& 16.4 & 48.1 \\
\cmidrule(lr){2-14}
& \multicolumn{2}{l|}{Real dataset objects}  & 73.3 & 8.23 & 10. & 55.5 & 42.2 & 22.0 & 97.3 & 91.7 & 46.5 & 16.4 & 46.4 \\
\bottomrule
    \end{tabular}
    \hspace*{6mm}
    \vspace*{-1mm}
    \caption{\textbf{Feature-based metrics.} We present class-wise results of APC.}
  \label{tab:feature_metrics_apc}
\end{table*}

\clearpage
\clearpage
{
    \small
    \bibliographystyle{ieeenat_fullname}
    \bibliography{main}
}

\end{document}